\documentclass[letterpaper, 11pt]{article}
\usepackage[margin=1in]{geometry}

\RequirePackage[OT1]{fontenc}
\RequirePackage{amsthm,amsmath,amsfonts,amssymb,mathtools}
\usepackage{bm, bbm}
\usepackage{color,xcolor}
\usepackage[colorlinks, citecolor=blue, linkcolor=red]{hyperref}
\usepackage[capitalise,nameinlink]{cleveref}
\usepackage{natbib}
\usepackage{subfigure}
\usepackage{multirow}
\usepackage{tikz}
\usepackage{amsmath}
\usepackage{ulem}

\newtheorem{theorem}{Theorem}[section]

\theoremstyle{definition}
\newtheorem{definition}[theorem]{Definition}

\theoremstyle{remark}
\newtheorem{remark}[theorem]{Remark}

\numberwithin{equation}{section}

\usepackage[linesnumbered,ruled,vlined]{algorithm2e}
\usepackage{color}
\usepackage{graphicx}
\usepackage[section]{placeins}
\usepackage[title]{appendix}

\usepackage{subcaption}
\usepackage{multicol}

\usepackage[pagewise]{lineno} %\linenumbers

\usepackage{hyperref}
\usepackage{verbatim}

\title{Dynamic PET Image Reconstruction via Non-negative INR Factorization}

\author{Chaozhi Zhang\thanks{School of Mathematical Sciences, Shanghai Jiao Tong University, 200240 Shanghai, China.
(\href{zhangcz4991@sjtu.edu.cn}{zhangcz4991@sjtu.edu.cn}).}
\and Wenxiang Ding\thanks{Institute of Natural Sciences, Shanghai Jiao Tong University, 200240 Shanghai, China.
(\href{dingwenxiang@sjtu.edu.cn}{dingwenxiang@sjtu.edu.cn}).}
\and Roy Y. He\thanks{Department of Mathematics, City University of Hong Kong, Kowloon Tong, Hong Kong.
(\href{royhe2@cityu.edu.hk}{royhe2@cityu.edu.hk}).}
\and Xiaoqun Zhang\thanks{Institute of Natural Sciences, School of Mathematical Sciences, MOE-LSC \& Shanghai National Center for Applied Mathematics (SJTU Center), Shanghai Jiao Tong University, 200240 Shanghai, China.
(\href{xqzhang@sjtu.edu.cn}{xqzhang@sjtu.edu.cn},\href{dingqiaoqiao@sjtu.edu.cn}{dingqiaoqiao@sjtu.edu.cn}).}
\and Qiaoqiao Ding\footnotemark[4]}

\begin{document}

\maketitle

\begin{abstract}
The reconstruction of dynamic positron emission tomography (PET) images from noisy projection data is a significant but challenging problem. In this paper, we introduce an unsupervised learning approach, Non-negative Implicit Neural Representation Factorization, based on low rank matrix factorization of unknown images and employing neural networks to represent both coefficients and bases. Mathematically, we demonstrate that if a sequence of dynamic PET images satisfies a generalized non-negative low-rank property, it can be decomposed into a set of non-negative continuous functions varying in the temporal-spatial domain. This bridges the well-established non-negative matrix factorization with continuous functions and we propose using implicit neural representations to connect matrix with continuous functions. The neural network parameters are obtained by minimizing the KL divergence, with additional sparsity regularization on coefficients and bases. Extensive experiments on dynamic PET reconstruction with Poisson noise demonstrate the effectiveness of the proposed method compared to other methods, while giving continuous representations for object's detailed geometric features and regional concentration variation.
\end{abstract}

\section{Introduction}

Positron Emission Tomography is a nuclear imaging technique that employs positron-emitting tracers to generate high-resolution 3D metabolic activity maps. Widely used in clinical oncology, neurology, and cardiology, it enables precise detection and monitoring of diseases such as cancer, cardiovascular disorders, and neurodegenerative conditions~\cite{cherry2012physics, gambhir2002molecular}. Unlike conventional static PET, dynamic PET utilizes sequential data acquisition to produce a series
of time-varying images, which can estimate parametric images and time-activity curves (TACs). This temporal resolution enhances the characterization of metabolic processes, offering improved diagnostic accuracy, refined assessment of tumor microenvironments, and quantitative evaluation of therapeutic responses~\cite{rahmim2019dynamic}.

Reconstructing dynamic PET images from their projections is an ill-posed inverse problem, where the goal is to retrieve the dynamic radioisotope distribution with high temporal resolution from the sinogram. There are several challenges in PET image reconstruction. First, sinograms typically contain high-level noise and the radioactive substances decay rapidly, which leads to poor quality of reconstructed image. Second, motion and deformation may exist during the data acquisition. Lastly, dynamic image reconstruction is always computationally expensive and time-consuming. Therefore, developing an effective and efficient algorithm to address the incompleteness and ill-posedness of the reconstruction problem is crucial for achieving high-quality results.

Dynamic PET image reconstruction can be performed by applying static PET reconstruction methods to each frame individually. Traditional reconstruction methods mainly consist of filtered back projection (FBP)~\cite{ramachandran1971three, shepp1974fourier} and iterative methods~\cite{shepp1982maximum, hudson1994accelerated, browne1996row, anthoine2011some, liu2013poissonian}. Deep learning methods utilize neural networks for PET image reconstruction, such as automated transform by manifold approximation~\cite{zhu2018image}, convolutional neural network combined with alternating direction method of multipliers (ADMM)~\cite{gong2018iterative}, generative adversarial network~\cite{xie2020generative} and deep image prior (DIP)~\cite{gong2018pet}. However, reconstructing dynamic PET images frame by frame is inefficient and leads to suboptimal results, especially when the signal-to-noise ratio (SNR) of the sinograms is low. Deep learning methods based on CNNs~\cite{wang2020fbp, li2022deep}, which treat dynamic PET images as a 3D/4D data to input, have been proposed for dynamic PET reconstruction. However, these methods ignore the connection among frames on the time scale. To enhance the robustness, it is necessary to consider structural priors that integrate both time and space. A common approach is to model the tracer distribution function as a linear combination of a set of basis functions~\cite{jin2007dynamic}, including exponential functions~\cite{snyder1984parameter}, B-spline functions~\cite{nichols2002spatiotemporal, verhaeghe2004ml} and functions estimated using Karhunen-Loeve transform~\cite{wernick1999fast}. This naturally leads to a low-rank prior for the underlying images and the reconstruction can be mathematically modeled as a problem of non-negative matrix factorization (NMF). Many NMF models incorporate regularization to achieve satisfying results, and they often lead to non-convex optimization problems which require manually tuning multiple algorithmic parameters~\cite{kawai2017robust, cui2019simultaneous}. In~\cite{yokota2019dynamic}, NMF is incorporated with DIP~\cite{ulyanov2018deep} for dynamic PET image reconstruction.

In this work, we explore the use of implicit neural representations (INRs) within the NMF model for dynamic PET image reconstruction. INRs take continuous coordinates as inputs and model complex data such as images, 3D shapes, and audio signals. This capability has led to a wide range of applications including super resolution~\cite{chen2021learning}, 3D object reconstruction~\cite{mildenhall2021nerf}, and image compression~\cite{strumpler2022implicit}. In the field of medical imaging, INRs have been applied in various contexts, such as implicit spline representations~\cite{barrowclough2021binary} and Retinal INR~\cite{gu2023retinal} for segmentation; medical image registration via neural fields~\cite{sun2024medical} and implicit deformable image registration~\cite{wolterink2022implicit} for registration; and Tree-structured Implicit Neural Compression~\cite{yang2023tinc} for compression. In the context of reconstruction tasks, implicit neural representation learning with prior embedding~\cite{shen2022nerp} uses INR to learn the mapping from coordinates to the image domain for reconstructing sparsely sampled medical images. The method in~\cite{reed2021dynamic} employs INR to represent the linear attenuation coefficient of the 3D volume for reconstructing dynamic, time-varying scenes with computed tomography.

Unlike previous works, we propose propose a novel approach that integrate INRs and NMF for dynamic PET image reconstruction. INRs are utilized to model the spatial and temporal bases within the NMF framework. Since INRs are resolution independent, they offer greater memory efficiency at high resolutions compared to CNN-based models like \cite{yokota2019dynamic}. Additionally, our proposed method is unsupervised so that it can overcome the difficulties of collection of image pairs which requires two-times scan and registration for alignment of every image/projection pair.

The main contributions of this paper are as follows:
\begin{enumerate}
    \item We introduce INRs to model dynamic PET images while leveraging the low-rank properties characterized by the NMF model. By imposing regularization on spatial and temporal bases, we propose the \texttt{NINRF} method for dynamic PET image reconstruction, which is memory efficient and resolution independent.
    \item We proposed to use an unsupervised model, which works well in a data-limited environment.
    \item We perform comprehensive experiments to validate our model. By focusing on simulated and clinical images, we systematically examine the accuracy and stability of our model’s reconstructions concerning specific tissues that are particularly relevant for applications. Additionally, the kinetic parameter derived from the reconstructed images also performs well in quantitative analysis.
\end{enumerate}

This paper is organized as follows. In \cref{dyPET rec with pn}, we introduce the problem formulation for dynamic PET image reconstruction with Poisson noise. \cref{Model} describes the NMF model and analyzes its continuous representation in relation to the INR model. In \cref{Model2}, we present the proposed non-negative INR factorization method along with implementation specifics. Experimental results are presented in \cref{experiment}, followed by the conclusion in \cref{conclude}.

\section{Dynamic PET Image Reconstruction}\label{dyPET rec with pn}

The reconstruction of dynamic PET images is to obtain a series of spatial distributions of radioactivity from a sequence of sinograms. Mathematically, this can be mathematically formulated as a linear inverse problem. Given a sequence of projection data, denoted as $\bm{z}_{1},\dots,\bm{z}_{T}$, we aim to reconstruct the corresponding dynamic PET image sequence $\bm{u}_{1},\dots,\bm{u}_{T}$. At each time $t$, the observation model can be described as
\begin{equation} \label{sinogram1}
    \bm{z}_t=\bm{P}\bm{u}_t+\bm{s}_t,~~~t=1,2 \cdots T.
\end{equation}
where $\bm{P}$ is the forward projection matrix representing discretized Radon transform. Here, $\bm{u}_{t}\in\mathbb{R}^{M}$ represents the image with $M=h\times w$ pixels, where $h$ and $w$ are the image height and width, respectively. The corresponding sinogram $\bm{z}_{t}\in\mathbb{R}^{N}$ has dimensions $N=n_a\times n_l$, where $n_a$ is the number of projection angles and $n_l$ is the detector size. $\bm{s}_t$ represents the expectation of scattered and random events. For ease of notations, the dynamic image is represented as $\bm{U}= [\bm{u}_{1},\dots,\bm{u}_{T}]\in\mathbb{R}^{M\times T}$, the expectation of scatters and randoms is represented as $\bm{S}= [\bm{s}_{1},\dots,\bm{s}_{T}]\in\mathbb{R}^{N\times T}$, and the corresponding sinogram is represented as $\bm{Z}=[\bm{z}_{1},\dots,\bm{z}_{T}]\in\mathbb{R}^{N\times T}$.
Thus, we have 
$$[\bm{z}_{1},\dots,\bm{z}_{T}]=\bm{PU+\bm{S}}=[\bm{P}\bm{u}_1+\bm{s}_{1},\dots,\bm{P}\bm{u}_T+\bm{s}_{T}].$$
In practice, the observed projection data is often contaminated by Poisson noise.
The noisy measurement is still represented as $\bm{z}_{t}\in\mathbb{R}^{N}$ and the observation model is reformulated as
\begin{equation} \label{sinogram2}
    \bm{Z}\sim\text{Poisson}(\bm{P}\bm{U}+\bm{S}).
\end{equation}

In PET reconstruction, FBP method~\cite{ramachandran1971three, shepp1974fourier} and maximum likelihood
estimation method solved by EM algorithm~\cite{shepp1982maximum,lange1984reconstruction} are two fundamental methods to reconstruct PET image frame by frame. Considering the Poisson noise, maximum a posteriori (MAP) estimation~\cite{levitan1987maximum} is adopted by minimizing the Kullback-Leibler (KL) divergence combined with image prior,  
\begin{equation}\label{MAPloss}
    \min_{\bm{U}\geq 0}D_{\text{KL}}(\bm{Z}\Vert\bm{P}\bm{U}+\bm{S})+\lambda R(\bm{U}),
\end{equation}
where $D_{\text{KL}}(\bm{Z}\Vert\bm{P}\bm{U}+\bm{S})=\sum_{t=1}^T{D_{\text{KL}}(\bm{z}_t\Vert\bm{P}\bm{u}_t+\bm{s}_t})$ and $D_{\text{KL}}(\bm{p}\Vert\bm{q}):=\sum_i q_i-p_i\log q_i$. Here, $R(\bm{U})$ represents a prior regularization term, and $\lambda$ is the regularization parameter that balances the influence of data term and prior. In dynamic PET reconstruction, various regularizers that consider the spatial and temporal property are used.  

Notice that the reconstructed values in $\bm{U}$ are restricted to a finite set of grid points. In the following, we present our model utilizing INRs to enhance the approximation accuracy and extend the estimations to continuous domains.

\section{Non-negative INR Factorization}\label{Model}
 
We denote the dynamic image as $\bm{U}=[\bm{u}_{1},\dots,\bm{u}_{T}]\allowbreak\in\mathbb{R}^{M \times T}$, where each column $\bm{u}_{t}$ is a vectorized image at time $t$. In this paper, we use the NMF of the matrix to enforce the low-rankness. Alternatively, the dynamic PET image can be expressed as a tensor $\mathcal{U}\in\mathbb{R}^{h \times w \times T}$, capturing three dimensions: height, width, and time. We will give the existence property of NMF of the tensor.

\subsection{Non-negative Matrix Factorization and Implicit Neural Representations}

Dynamic PET imaging often relies on the compartment model to characterize tracer kinetics across different tissue or organ regions~\cite{phelps1979tomographic}. Intuitively, each compartment represents a physiological region or state whose tracer concentration varies over time according to a TAC. Such a physiological mechanism typically induces a low-rank structure in the spatiotemporal data matrix, since the number of meaningful compartments $K$ is usually much smaller than the number of time frames $T$. This low-rank assumption has been widely used in dynamic PET reconstruction and analysis (e.g.,~\cite{ding2015dynamic, ding2017dynamic, yokota2019dynamic}).

In a discrete representation, the dynamic PET image denoted as $\bm{U}=[\bm{u}_{1},\dots,\bm{u}_{T}]\in\mathbb{R}_{+}^{M \times T}$ can be factorized into two non-negative matrices:
\begin{equation}\label{NMF_disc}
    \bm{U} = \bm{A}\bm{B},
\end{equation}
where $\bm{A}=[\bm{a}_1,\dots,\bm{a}_K]\in\mathbb{R}_{+}^{M\times K}$ contains, for each spatial location, non-negative mixture coefficients over $K$ compartments, and $\bm{B}=\allowbreak[\bm{b}_1,\dots,\bm{b}_K]^{\top}\in\mathbb{R}_{+}^{K \times T}$ represents the corresponding TACs.

To leverage continuous modeling and exploit the advantages of INRs, we formulate the above discrete factorization to the continuous spatiotemporal setting. Let $u(\bm{x},t)$ denote the concentration distribution of the radioisotope at spatial location $\bm{x}=(x_1,x_2)$ and time $t$. The concentration $u(\bm{x},t)$ can be approximated as:
\begin{equation}\label{NMF_disc2}
    u(\bm{x},t) = \sum_{k=1}^K\bm{a}_k(\bm{x})\bm{b}_k(t),
\end{equation}
where $\bm{a}_k(\bm{x})$ is the spatial mixture weight for $k$-th compartment at location $\bm{x}$, and $\bm{b}_k(t)$ is the corresponding TAC at time $t$.

INRs encode signals and data using neural networks, where the information is represented as continuous functions parameterized by neural network weights. In this work, we employ distinct INR networks to model the mappings between coordinates and the coefficients $H_{f_{\bm{\Theta}_1}}: \mathbb{R}^2 \to \mathbb{R}_+^{K}$, as well as the mappings between time $t$ and the corresponding TACs $H_{g_{\bm{\Theta}_2}}: \mathbb{R} \to \mathbb{R}_+^{K}$, where $\bm{\Theta}_1$ and $\bm{\Theta}_2$ are the parameters of the corresponding INRs. This model relies on the decomposition of $u(\bm{x},t)$ such that the space and temporal variables are separable. In the following, we prove that it holds if the function $u$ satisfies a generalized low-rank property.

\subsection{Low Rank Property of Tensor Function}\label{sec analysis}
Recall that for a tensor $\mathcal{U}\in\mathbb{R}^{n_1\times n_2\times n_3}$ with $n_1,n_2,n_3\in\mathbb{N}$, the unfolding matrix along mode $i\in\{1,2,3\}$ is denoted as $\bm{U}^{(i)}\in\mathbb{R}^{n_i\times\prod_{j\ne i}n_j}$, which is obtained by reshaping $\mathcal{U}$ and exchanging coordinates. The Tucker rank of a tensor $\mathcal{U}$ is a vector defined as $\text{rank}_T(\mathcal{U})=[\text{rank}(\bm{U}^{(1)}), \text{rank}(\bm{U}^{(2)}), \text{rank}(\bm{U}^{(3)})]$. We denote $\mathcal{U}_{(i,j,k)}$ as the $i,j,k$-th component of $\mathcal{U}$.

These concepts can be generalized to functions defined over continuous domains. Let $G(\cdot):D_h \times D_w \times D_t \to \mathbb{R}$ be a bounded real-valued function, where $D_h,D_w,D_T \subset \mathbb{R}$ represent the continuous domains of the individual dimensions. Following~\cite{luo2023low}, we interpret $G$ as a \textit{tensor function} and introduce some definitions.

\begin{definition}[Definition 2. in \cite{luo2023low}]
    For a tensor function $G(\cdot):D_h \times D_w \times D_T \to \mathbb{R}$, the sampled tensor set $S[G]$ is defined as
    \begin{equation}
        S[G] := \{\mathcal{U}|\mathcal{U}_{(i,j,k)}=G({\bm{x}_{1}}_{(i)},{\bm{x}_{2}}_{(j)},\bm{t}_{(k)}), {\bm{x}_{1}}\in D_h^{n_1}, {\bm{x}_{2}}\in D_w^{n_2}, \bm{t}\in D_t^{n_3}, n_1, n_2, n_3 \in \mathbb{N}_+ \}
    \end{equation}
    where $\bm{x}_{1}$, $\bm{x}_{2}$, and $\bm{t}$ denote the coordinate vector variables, ${\bm{x}_{1}}_{(i)}$, ${\bm{x}_{2}}_{(j)}$ and $\bm{t}_{(k)})$ are the $i$-th, $j$-th and $k$-th element of $\bm{x}_{1}$, $\bm{x}_{2}$, and $\bm{t}$, and $n_1$, $n_2$, and $n_3$ are positive integer variables that determine the sizes of the sampled tensor $\mathcal{U}$.
\end{definition}

\begin{definition}[Definition 3. in \cite{luo2023low}]
    Given a tensor function $G(\cdot):D_h \times D_w \times D_T \to \mathbb{R}$, we define a measure of its complexity, denoted by $\text{F-rank}[G]$, as the supremum of Tucker rank in the sampled tensor set $S[G]$
    \begin{equation}
        \text{F-rank}[G]:= (r_1,r_2,r_3), \text{ where }  r_i=\sup_{\mathcal{U}\in S[G]} \text{rank}(\bm{U}^{(i)}).
    \end{equation}
\end{definition}

Based on Theorem 2 of~\cite{luo2023low}, we introduce the following Theorem.
\begin{theorem}\label{thm0}
    Let $G(\cdot): D_h \times D_w \times D_T \to \mathbb{R}$ be a bounded tensor function. If the third component of $\text{F-rank}[G]$ is $K$, then there exist two bounded functions $\bm{G}_1(\cdot):D_h \times D_w\to \mathbb{R}^{K}$ and $\bm{G}_2(\cdot):D_T \to \mathbb{R}^{K}$ such that for any $(v_1,v_2,v_3)\in D_h \times D_w \times D_T$, $G(v_1,v_2,v_3)=\bm{G}_1(v_1,v_2)^{\top}\bm{G}_2(v_3)$.
\end{theorem}
The proof is provided in \cref{proof_thm0}. 

\subsection{Main Result: Non-negative INR Factorization}
The elements in dynamic PET image, spatial bases and temporal bases are supposed to be all non-negative. We introduce a new rank definition specifically tailored for bounded tensor functions with non-negative values.

\begin{definition}
    Given a non-negative tensor function $G(\cdot):D_h \times D_w \times D_T \to \mathbb{R}_{+}$, we define its non-negative rank, denoted by $\text{F-rank}_{+}[G]$, as the supremum of non-negative Tucker rank in the sampled tensor set $S[G]$
    \begin{equation}
        \text{F-rank}_{+}[G] := (r_1,r_2,r_3), \text{ where }  r_i=\sup_{\mathcal{U}\in S[h]} \text{rank}_{+}(\bm{U}^{(i)}).
    \end{equation}
\end{definition}
We state our main results for non-negative tensor function factorization as follows.
\begin{theorem}\label{thm1}
    Let $G(\cdot):D_h \times D_w \times D_T \to \mathbb{R}_{+}$ be a non-negative bounded tensor function. If the following two assumptions hold:
    \begin{enumerate}
        \item The third components of both $\text{F-rank}[G]$ and $\text{F-rank}_{+}[G]$ are equal to $K$,
        \item There exists a tensor $\mathcal{T}=G(\mathbf{v}_1,\mathbf{v}_2,\mathbf{v}_3)\in\mathbb{R}^{n_1 \times n_2 \times n_3}$ that belongs to $S[G]$ and satisfies $n_1n_2=K$. $\mathbf{T}^{(3)}$ is the unfolding matrix of $\mathcal{T}$ along mode 3 and $\text{rank}(\mathbf{T}^{(3)})=\text{rank}_{+}(\mathbf{T}^{(3)})=K$. We define the sets $\mathcal{S}_0=\{G(x,y,\mathbf{v}_3)|x\in D_h, y\in D_w\}$ and  $\mathcal{S}_1=\{\mathcal{T}_{(i,j,:)}|i=1,2,\dots,n_1, j=1,2,\dots,n_2\}$. It holds that $\text{Cone}(\mathcal{S}_0)\subset\text{Cone}(\mathcal{S}_1)$, where $\text{Cone}(\mathcal{S})$ denotes the conic hull of set $\mathcal{S}$.
    \end{enumerate}
    then there exist two non-negative bounded functions $\bm{G}_1(\cdot):D_h \times D_w\to \mathbb{R}^{K}_+$ and $\bm{G}_2(\cdot):D_T \to \mathbb{R}^{K}_+$ such that for any $(v_1,v_2,v_3)\in D_h \times D_w \times D_T$, $G(v_1,v_2,v_3)=\bm{G}_1(v_1,v_2)^{\top}\bm{G}_2(v_3)$.
\end{theorem}
The proof is delegated to \cref{proof_thm1}. \cref{thm1} shows that if the continuous representation of dynamic PET image $u$ has a low-rank property in the above sense, then a variable separating decomposition exists. Furthermore, $u$ can be expressed as a product of two non-negative-valued INRs, one defined over spatial dimension and the other over temporal dimension. More explicitly, we obtain a non-negative INR factorization (\texttt{NINRF}) as follows
\begin{equation}\label{eq_NINRF}
    u(\bm{x},t) = H_{f_{\bm{\Theta}_1}}(\bm{x})^{\top}H_{g_{\bm{\Theta}_2}}(t).
\end{equation}
Here $H_{f_{\bm{\Theta}_1}}:D_h\times D_w\to\mathbb{R}_+^K$ and $H_{g_{\bm{\Theta}_2}}:D_T\to\mathbb{R}_+^K$ are two non-negative INRs parameterized by $\bm{\Theta}_1$ and $\bm{\Theta}_2$, respectively. This establishes the theoretical foundation for the proposed method, which will be introduced in the following section.

\section{Proposed Method}\label{Model2}
In this section, we provide a detailed introduction to the proposed self-supervised method for dynamic PET reconstruction from noisy measurements.
\subsection{Dynamic PET Reconstruction via \texttt{NINRF}}\label{Model2-1}
Let $\Omega=[0,1]\times[0,1]$ be a continuous image domain, and $u:\Omega\times[0,1]\to\mathbb{R}_+$ be the PET image to be reconstructed. Assuming that $u$ satisfies the low-rank property outlined in \cref{thm1}, then for some $K$, we can find non-negative INRs $H_f(\cdot,\bm{\Theta}_1):\Omega\to\mathbb{R}_+^K$ and $H_g(\cdot,\bm{\Theta}_2): [0,1]\to\mathbb{R}_+^K$ such that $H_f(\bm{x},\bm{\Theta}_1)^\top H_g(t,\bm{\Theta}_2)$ approximates $u(\bm{x},t)$ for any $(\bm{x},t)\in \Omega\times [0,1]$. In practice, we employ distinct INRs to each direction of $H_f$ and $H_g$, respectively. Thus, we have 
\begin{align}
  H_f(\cdot,\bm{\Theta}_1)=[f(\cdot,{\bm{\theta}_{11}}),\dots,f(\cdot,{\bm{\theta}_{1K}})]^{\top}\nonumber\\
  H_g(\cdot,\bm{\Theta}_2)=[g(\cdot,{\bm{\theta}_{21}}),\dots,g(\cdot,{\bm{\theta}_{2K}})]^{\top},\nonumber
\end{align}
where $f(\cdot,{\bm{\theta}_{1k}}):\Omega\to\mathbb{R}_+$ and $g(\cdot,{\bm{\theta}_{2k}}):[0,1]\to\mathbb{R}_+$ are the $k$-th components in $H_f$ and $H_g$, and $\bm{\theta}_{1k}$ and $\bm{\theta}_{2k}$ are the corresponding network parameters, respectively. Then, we denote the complete parameter sets as $\bm{\Theta}_1=\{\bm{\theta}_{11},\dots,\bm{\theta}_{1K}\}$ and $\bm{\Theta}_2=\{\bm{\theta}_{21},\dots,\bm{\theta}_{2K}\}$. 

For any $h,w,T\in\mathbb{N}$, we define the grid points $\Omega_{h,w}=\{(i/h, j/w)| i=0,\dots,h-1, j = 0,\dots,w-1\}$ and $\mathcal{D}_T = \{0,1/T,\dots,(T-1)/T\}$. We organize the INR values sampled from the grids as the following matrices
\begin{equation}
\bm{A}_{\bm{\Theta}_1} :=\begin{bmatrix}
f\big((0,0),\bm{\theta}_{11}\big) & \cdots & f\big((0,0),\bm{\theta}_{1K}\big) \\
f\big((0,\frac{1}{w}),\bm{\theta}_{11}\big) & \cdots & f\big((0,\frac{1}{w}),\bm{\theta}_{1K}\big) \\
\cdots & \cdots & \cdots \\
f\big((\frac{1}{h},0),\bm{\theta}_{11}\big) & \cdots & f\big((\frac{1}{h},0),\bm{\theta}_{1K}\big) \\
\cdots & \cdots & \cdots \\
f\big((\frac{h-1}{h},\frac{w-1}{w}),\bm{\theta}_{11}\big) & \cdots & f\big((\frac{h-1}{h},\frac{w-1}{w}),\bm{\theta}_{1K}\big)
\end{bmatrix}\;,\nonumber
\end{equation}
\begin{equation}\label{MAT}
~\bm{B}_{\bm{\Theta}_2}:=\begin{bmatrix}
g(0,\bm{\theta}_{21}) & g(\frac{1}{T},\bm{\theta}_{21}) & \cdots & g(\frac{T-1}{T},\bm{\theta}_{21}) \\
g(0,\bm{\theta}_{22}) & g(\frac{1}{T},\bm{\theta}_{22}) & \cdots & g(\frac{T-1}{T},\bm{\theta}_{22}) \\
\cdots & \cdots & \cdots & \cdots \\
g(0,\bm{\theta}_{2K}) & g(\frac{1}{T},\bm{\theta}_{2K}) & \cdots & g(\frac{T-1}{T},\bm{\theta}_{2K}) \\
\end{bmatrix}\;.
\end{equation}
We denote $\bm{a}_k\in\mathbb{R}^{hw}$ as the $k$-th column vector of $\bm{A}_{\bm{\Theta}_1}$ and $\bm{b}_k\in\mathbb{R}^{T}$ as the $k$-th row vector of $\bm{B}_{\bm{\Theta}_2}$ for $k=1,2,\dots,K$. Then, the dynamic image $\bm{U}$ can be approximated as $\bm{A}_{\bm{\Theta}_1}\bm{B}_{\bm{\Theta}_2}.$

To reconstruct $\bm{U}$ from noisy sinogram data $\bm{Z}\in\mathbb{R}^{N \times T}$, we propose to solve the following minimization problem:
\begin{equation} \label{reconstruction loss}
    \min_{\bm{\Theta}_1,\bm{\Theta}_2} \mathcal{L}(\bm{\Theta}_1, \bm{\Theta}_2) := D_{\text{KL}}(\bm{Z}\Vert\bm{P}\bm{A}_{\bm{\Theta}_1}\bm{B}_{\bm{\Theta}_2}+\bm{S}) + \lambda_1 R_1(\bm{A}_{\bm{\Theta}_1}) + \lambda_2 R_2(\bm{B}_{\bm{\Theta}_2}).
\end{equation}
Here, $R_1(\bm{A}_{\bm{\Theta}_1})=\sum_{k=1}^{K}{\text{TV}(\bm{a}_k)}$, where $\bm{a}_k$ is the $k$-th column of $\bm{A}_{\bm{\Theta}_1}$. By reshaping $\bm{a}_k$ into a matrix form, $\text{TV}(\bm{a}_k)$ is isotropic total variation (TV) \cite{rudin1992nonlinear} of $\bm{a}_k$. Similarly, $R_2(\bm{B}_{\bm{\Theta}_2})=\sum_{k=1}^{K}{\|\nabla_t(\bm{b}_k)\|_2^2}$, where $\bm{b}_k$ is the $k$-th row of $\bm{B}_{\bm{\Theta}_2}$, and $\nabla_t(\bm{b}_k)=[\bm{b}_{k,2}-\bm{b}_{k,1},\dots,\bm{b}_{k,T}-\bm{b}_{k,T-1}]^{\top}$. The regularization term $R_1(\bm{A}_{\bm{\Theta}_1})$ is adopted to suppress excessive oscillations in the resulted spatial bases, and $R_2(\bm{B}_{\bm{\Theta}_2})$ is used to enforce smoothness of the temporal basis. The regularization hyperparameters $\lambda_1>0$ and $\lambda_2>0$ are used to control the effects of the two regularization terms.

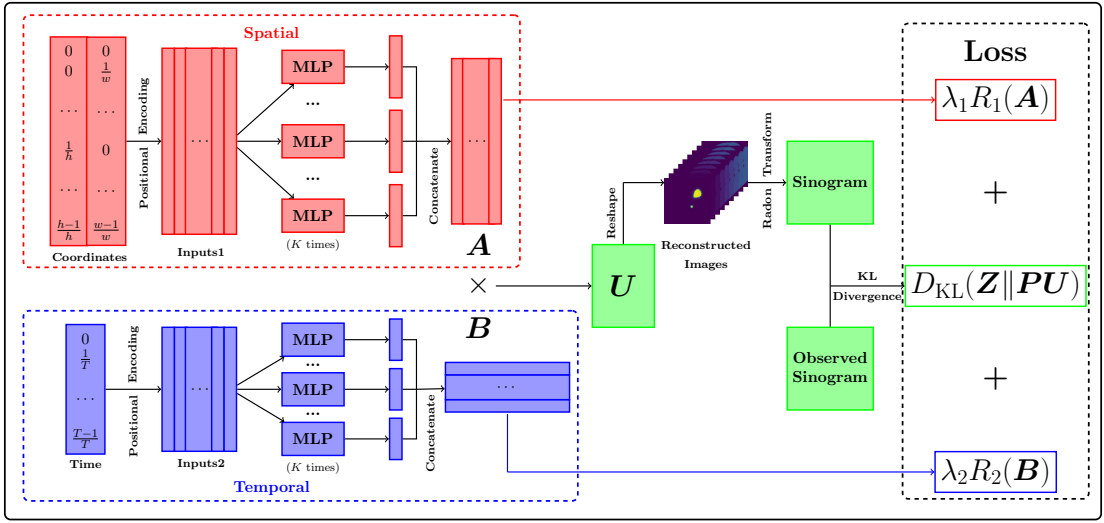
\begin{figure}[htbp!]
    \centering
    \scalebox{0.55}{
    \begin{tikzpicture}

    % \node[rectangle,fill=black!0,draw=black!100, line width=1.2pt, rounded corners, minimum width=26cm, minimum height=13cm] (Box) at (12,0.5) {};
    \node[rectangle, draw=black!100, line width=1.2pt, rounded corners, minimum width=28.1cm, minimum height=12.5cm] (Box) at (13.1,0.1) {};
    
    \node[rectangle,draw=red!100, line width=1.2pt, dashed, rounded corners, minimum width=12cm, minimum height=6.1cm] (spatial block) at (5.5,3.0) {};
    \node[rectangle] (Spatial) at (5.5,5.6) {\textcolor{red}{\textbf{Spatial}}};
    
    \node[rectangle,draw=blue!100, line width=1.2pt, dashed, rounded corners, minimum width=13.4cm, minimum height=4.6cm] (temporal block) at (6.2,-3.4) {};
    \node[rectangle] (Temporal) at (5.5,-5.4) {\textcolor{blue}{\textbf{Temporal}}};

    \node[rectangle,draw=black!100, line width=1.2pt, dashed, rounded corners, minimum width=5.9cm, minimum height=11.5cm] (loss block) at (24,0.1) {};
    \node[rectangle] (Loss) at (24,5.2) {\huge{\textbf{Loss}}};
    
    \node [rectangle,fill=red!40,draw=red!100, line width=1pt, minimum width=0.9cm, minimum height=4.5cm] (input_h) at (0.6,3) {$\begin{matrix}
        0 \\ 0 \\ \\  \cdots \\ \\  \frac{1}{h} \\ \\ \cdots \\ \\ \frac{h-1}{h}
    \end{matrix}$};		
    \node [rectangle,fill=red!40,draw=red!100, line width=1pt, minimum width=0.9cm, minimum height=4.5cm] (input_w) at (1.5,3) {$\begin{matrix}
        0 \\ \frac{1}{w} \\ \\  \cdots \\ \\  0 \\ \\ \cdots \\ \\ \frac{w-1}{w}
    \end{matrix}$};	
    \node [rectangle] (Coordinates) at (1.1,0.2) {\scriptsize\textbf{Coordinates}};
    \node [rectangle,fill=red!40,draw=red!100, line width=1pt, minimum width=0.3cm, minimum height=4.5cm] (input11) at (3,3) {};
    \node [rectangle,fill=red!40,draw=red!100, line width=1pt, minimum width=0.3cm, minimum height=4.5cm] (input12) at (3.3,3) {};
    \node [rectangle,fill=red!40,draw=red!100, line width=1pt, minimum width=0.6cm, minimum height=4.5cm] (input13) at (3.8,3) {$\cdots$};
    \node [rectangle,fill=red!40,draw=red!100, line width=1pt, minimum width=0.3cm, minimum height=4.5cm] (input14) at (4.2,3) {};
    \node [rectangle,fill=red!40,draw=red!100, line width=1pt, minimum width=0.3cm, minimum height=4.5cm] (input15) at (4.5,3) {};
    \node [rectangle] (Inputs1) at (3.8,0.3) {\scriptsize\textbf{Inputs1}};
    \draw [->,color=black,line width=0.8pt] (input_w)--node[left, rotate=90]{\scriptsize\textbf{Positional}}node[right, rotate=90]{\scriptsize\textbf{Encoding}} (input11);
    
    \node [rectangle,fill=red!40,draw=red!100, line width=1pt, minimum width=1.5cm, minimum height=0.8cm] (MLP1) at (6.5,4.8) {\textbf{MLP}};
    \node [rectangle,fill=red!40,draw=red!100, line width=1pt, minimum width=1.5cm, minimum height=0.8cm] (MLP2) at (6.5,3.0) {\textbf{MLP}};
    \node [rectangle,fill=red!40,draw=red!100, line width=1pt, minimum width=1.5cm, minimum height=0.8cm] (MLP3) at (6.5,1.2) {\textbf{MLP}};
    
    \node [rectangle] (dots1) at (6.5,3.9) {\textbf{...}};
    \node [rectangle] (dots2) at (6.5,2.1) {\textbf{...}};
    \node [rectangle] (Ktimes1) at (6.5,0.5) {\scriptsize($K$ times)};
    
    \draw [->,color=black,line width=0.8pt] (input15)-- (MLP1);
    \draw [->,color=black,line width=0.8pt] (input15)-- (MLP2);
    \draw [->,color=black,line width=0.8pt] (input15)-- (MLP3);
    
    \node [rectangle,fill=red!40,draw=red!100, line width=1pt, minimum width=0.3cm, minimum height=1.5cm] (Spatial1) at (8.5,4.8){};
    \node [rectangle,fill=red!40,draw=red!100, line width=1pt, minimum width=0.3cm, minimum height=1.5cm] (Spatial2) at (8.5,3.0){};
    \node [rectangle,fill=red!40,draw=red!100, line width=1pt, minimum width=0.3cm, minimum height=1.5cm] (Spatial3) at (8.5,1.2){};
    
    \draw [->,color=black,line width=0.8pt] (MLP1)-- (Spatial1);
    \draw [->,color=black,line width=0.8pt] (MLP2)-- (Spatial2);
    \draw [->,color=black,line width=0.8pt] (MLP3)-- (Spatial3);
    
    \node [rectangle,fill=red!40,draw=red!100, line width=1pt, minimum width=0.3cm, minimum height=4cm] (A1) at (10.0,3) {};
    \node [rectangle,fill=red!40,draw=red!100, line width=1pt, minimum width=0.6cm, minimum height=4cm] (A2) at (10.5,3) {$\cdots$};
    \node [rectangle,fill=red!40,draw=red!100, line width=1pt, minimum width=0.3cm, minimum height=4cm] (A3) at (10.9,3) {};
    
    \coordinate (conn1) at (9.0,3.0);
    \draw [-,color=black,line width=0.8pt] (Spatial1)-| (conn1);
    \draw [-,color=black,line width=0.8pt] (Spatial2)-| (conn1);
    \draw [-,color=black,line width=0.8pt] (Spatial3)-| (conn1);
    \draw [->,color=black,line width=0.8pt] (conn1)--node[left, rotate=90]{\scriptsize\textbf{Concatenate}} (A1);
    
    \node [rectangle,fill=blue!40,draw=blue!100, line width=1pt, minimum width=3cm, minimum height=0.3cm] (B1) at (11.2,-3.4) {};
    \node [rectangle,fill=blue!40,draw=blue!100, line width=1pt, minimum width=3cm, minimum height=0.6cm] (B2) at (11.2,-2.95) {$\cdots$};
    \node [rectangle,fill=blue!40,draw=blue!100, line width=1pt, minimum width=3cm, minimum height=0.3cm] (B3) at (11.2,-2.5) {};
    
    \node [rectangle,fill=blue!40,draw=blue!100, line width=1pt, minimum width=0.9cm, minimum height=3cm] (input_t) at (1.0,-3) {$\begin{matrix}
        0 \\ \frac{1}{T} \\ \\  \cdots \\ \\   \frac{T-1}{T}
    \end{matrix}$};	
    \node [rectangle] (Time) at (1.0,-4.8) {\scriptsize\textbf{Time}};
    \node [rectangle,fill=blue!40,draw=blue!100, line width=1pt, minimum width=0.3cm, minimum height=3cm] (input21) at (3,-3) {};
    \node [rectangle,fill=blue!40,draw=blue!100, line width=1pt, minimum width=0.3cm, minimum height=3cm] (input22) at (3.3,-3) {};
    \node [rectangle,fill=blue!40,draw=blue!100, line width=1pt, minimum width=0.6cm, minimum height=3cm] (input23) at (3.8,-3) {$\cdots$};
    \node [rectangle,fill=blue!40,draw=blue!100, line width=1pt, minimum width=0.3cm, minimum height=3cm] (input24) at (4.2,-3) {};
    \node [rectangle,fill=blue!40,draw=blue!100, line width=1pt, minimum width=0.3cm, minimum height=3cm] (input25) at (4.5,-3) {};
    \node [rectangle] (Inputs2) at (3.8,-4.8) {\scriptsize\textbf{Inputs2}};
    \draw [->,color=black,line width=0.8pt] (input_t)--node[left, rotate=90]{\scriptsize\textbf{Positional}}node[right, rotate=90]{\scriptsize\textbf{Encoding}} (input21);
    
    \node [rectangle,fill=blue!40,draw=blue!100, line width=1pt, minimum width=1.5cm, minimum height=0.8cm] (MLPt1) at (6.5,-1.8) {\textbf{MLP}};
    \node [rectangle,fill=blue!40,draw=blue!100, line width=1pt, minimum width=1.5cm, minimum height=0.8cm] (MLPt2) at (6.5,-3) {\textbf{MLP}};
    \node [rectangle,fill=blue!40,draw=blue!100, line width=1pt, minimum width=1.5cm, minimum height=0.8cm] (MLPt3) at (6.5,-4.2) {\textbf{MLP}};
    \node [rectangle] (dotst1) at (6.5,-2.4) {\textbf{...}};
    \node [rectangle] (dotst2) at (6.5,-3.6) {\textbf{...}};
    \node [rectangle] (Ktimes2) at (6.5,-4.9) {\scriptsize($K$ times)};
    \draw [->,color=black,line width=0.8pt] (input25)-- (MLPt1);
    \draw [->,color=black,line width=0.8pt] (input25)-- (MLPt2);
    \draw [->,color=black,line width=0.8pt] (input25)-- (MLPt3);
    
    \node [rectangle,fill=blue!40,draw=blue!100, line width=1pt, minimum width=0.3cm, minimum height=1cm] (Time1) at (8.5,-1.8){};
    \node [rectangle,fill=blue!40,draw=blue!100, line width=1pt, minimum width=0.3cm, minimum height=1cm] (Time2) at (8.5,-3){};
    \node [rectangle,fill=blue!40,draw=blue!100, line width=1pt, minimum width=0.3cm, minimum height=1cm] (Time3) at (8.5,-4.2){};
    
    \draw [->,color=black,line width=0.8pt] (MLPt1)-- (Time1);
    \draw [->,color=black,line width=0.8pt] (MLPt2)-- (Time2);
    \draw [->,color=black,line width=0.8pt] (MLPt3)-- (Time3);
    
    \coordinate (conn2) at (9,-3);
    \draw [-,color=black,line width=0.8pt] (Time1)-| (conn2);
    \draw [-,color=black,line width=0.8pt] (Time2)-| (conn2);
    \draw [-,color=black,line width=0.8pt] (Time3)-| (conn2);
    \draw [->,color=black,line width=0.8pt] (conn2)--node[left, rotate=90]{\scriptsize\textbf{Concatenate}}(B2);
    
    \node [rectangle] (A) at (10.5,0.5) {\huge$\bm{A}$};
    \node [rectangle] (B) at (10.5,-1.5) {\huge$\bm{B}$};
    \node [rectangle] (mul) at (10.5,-0.5) {\huge$\times$};
    
    \node [rectangle,fill=green!40,draw=green!100, line width=1pt, minimum width=1.5cm, minimum height=2cm] (output_u) at (14,-0.5) {\huge$\bm{U}$};
    
    \draw [->,color=black,line width=0.8pt] (mul)--(output_u);
    
    \node[inner sep=0] (recimages) at (16,2) {\includegraphics[width=2cm]{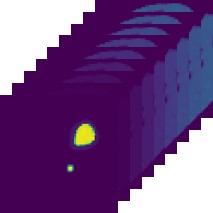}};
    \node [rectangle] (Recimages1) at (16,0.5) {\scriptsize\textbf{Reconstructed}};
    \node [rectangle] (Recimages2) at (16,0) {\scriptsize\textbf{Images}};

    \coordinate (conn3) at (14,2);
    \draw [-,color=black,line width=0.8pt] (output_u)--node[above,rotate=90]{\scriptsize\textbf{Reshape}}(conn3);
    \draw [->,color=black,line width=0.8pt] (conn3)--(recimages);
    
    \node [rectangle,fill=green!40,draw=green!100, line width=1pt, minimum width=1.5cm, minimum height=2cm] (recsino) at (19,2) {{\textbf{Sinogram}}};
    
    \draw [->,color=black,line width=0.8pt] (recimages)--node[left,rotate=90]{\scriptsize\textbf{Radon}}node[right,rotate=90]{\scriptsize\textbf{Transform}}(recsino);
    
    \node [rectangle,fill=green!40,draw=green!100, line width=1pt, minimum width=1.5cm, minimum height=2cm] (obsino) at (19,-2.5) {$\begin{matrix} \text{\textbf{Observed}} \\ \text{\textbf{Sinogram}} \end{matrix}$};

    \coordinate (conn4) at (19,-0.5);
    \draw [-,color=black,line width=0.8pt] (recsino)--(conn4);
    \draw [-,color=black,line width=0.8pt] (obsino)--(conn4);

    \node [rectangle,draw=green!100, line width=1pt] (KL) at (24,-0.5) {\huge$D_{\text{KL}}(\bm{Z}\Vert\bm{P}\bm{U}+\bm{S})$};

    \draw [->,color=black,line width=0.8pt] (conn4)--node[above]{\scriptsize\textbf{KL}}node[below]{\scriptsize\textbf{Divergence}}(KL);

    \node [rectangle,draw=red!100, line width=1pt] (R1) at (24,4) {\huge$\lambda_1 R_1(\bm{A})$};
    \node [rectangle,draw=blue!100, line width=1pt] (R2) at (24,-5) {\huge$\lambda_2 R_2(\bm{B})$};

    \coordinate (conn5) at (11.1,4);
    \draw [->,color=red,line width=0.8pt] (conn5)--(R1);
    \coordinate (conn61) at (11.2,-3.6);
    \coordinate (conn62) at (11.2,-5.0);
    \draw [-,color=blue,line width=0.8pt] (conn61)--(conn62);
    \draw [->,color=blue,line width=0.8pt] (conn62)--(R2);

    \node [rectangle] (plus1) at (24,1.75) {\huge$\bm{+}$};
    \node [rectangle] (plus2) at (24,-2.75) {\huge$\bm{+}$};
    
    % \draw [->,color=black,line width=0.8pt] (R1)--node[right]{\large\textbf{$+$}}(KL);
    % \draw [->,color=black,line width=0.8pt] (R2)--node[right]{\large\textbf{$+$}}(KL);
    	
    \end{tikzpicture}
    }
    \caption{Framework of the proposed method. The rescaled coordinates and time values, after positional encoding, are used for training two series of INRs. The resulting spatial and temporal bases are concatenated to form the matrices $\bm{A}_{\bm{\Theta}_1}$ and $\bm{B}_{\bm{\Theta}_2}$ in~\eqref{reconstruction loss}. These matrices are then combined to reconstruct the dynamic PET images. The primary loss function is the KL divergence between the projection of the reconstructed images, and the acquired sinograms contaminated by Poisson noise (as described in \eqref{sinogram2}). Our method minimizes this loss while incorporating regularization terms on $\bm{A}_{\bm{\Theta}_1}$ and $\bm{B}_{\bm{\Theta}_2}$ to ensure model stability and accuracy.}
    \label{f_inr}
\end{figure}

Upon solving~\eqref{reconstruction loss}, we obtain $\widehat{\bm{\Theta}}_1$ and $\widehat{\bm{\Theta}}_2$, and the reconstructed dynamic PET image is represented by
\begin{equation}
\widehat{u}(\bm{x},t) =\bm{A}(\bm{x};\widehat{\bm{\Theta}}_1)^\top\bm{B}(t;\widehat{\bm{\Theta}}_2)\;,\;~\forall (\bm{x},t)\in\Omega\times[0,1]\;.
\end{equation}
The overall framework of our proposed method is shown in \cref{f_inr}. 
\begin{remark} The proposed model~\eqref{reconstruction loss} is similar to the minimization of Equation~\eqref{MAPloss}, as both formulations incorporate the values of spatial and temporal bases at grid points.  However, they are fundamentally different. Minimizing Equation~\eqref{MAPloss} directly optimizes the grid point values, while the proposed model in Equation~\eqref{reconstruction loss} optimizes the parameters of the involved INRs. The selection of grid parameters  $h,w$ and $T$ influences the reconstruction accuracy, but the proposed method always provides a continuous representation of the reconstruction rather than a finite set of values. 
\end{remark}

\subsection{Implementation Details} \label{implementation}
In this section, we describe the implementation details of the proposed method.  

\subsubsection{Architectures of INRs} 
We employ fully connected networks (FCNs) to construct the INRs in the proposed model. For $L\geq 1$, an $L$-layer FCN $\bm{\phi}(\mathbf{x}, \bm{\theta})$ is described recursively as follows:
\begin{equation}\label{FCN}
\begin{aligned}
\bm{\phi}_{\bm{\theta}}(\mathbf{x}) &= \eta^{(L)}(\mathbf{W}^{(L)}\mathbf{x}^{(L-1)}+\mathbf{q}^{(L)}),\\
    \mathbf{x}^{(l)} &= \eta^{(l-1)}(\mathbf{W}^{(l)}\mathbf{x}^{(l-1)}+\mathbf{q}^{(l)})    , l=1,2,\dots,L-1\\
    \mathbf{x}^{(0)} &= \mathbf{x}, \\
\end{aligned}
\end{equation}
where $\{\eta^{(1)}(\cdot),\dots,\eta^{(L)}(\cdot)\}$ are nonlinear activation functions, and the network parameters are given by $\bm{\theta}=\{\mathbf{W}^{(1)}, \dots,\mathbf{W}^{(L)},\mathbf{q}^{(1)},\dots,\mathbf{q}^{(L)}\}$. We take all the activation functions as ReLU~\cite{glorot2011deep}. This specification is also compatible with the non-negativity of the proposed INR factorization. 

\noindent\textbf{Positional encoding.}
FCNs with ReLU activation functions exhibit spectral bias~\cite{rahaman2019spectral,basri2020frequency}, that is, they converge slower for high frequency components of the signals. To address this, we employ Fourier feature mappings~\cite{tancik2020fourier}, which encode the input coordinates into high dimensional vectors before feeding them into the networks. Specifically, we define the spatial encoding function as:
\begin{equation}\label{fbe-1}
    \psi_1(\bm{x}) = [\sin{(2\pi \bm{W}_1 \bm{x})}, \cos{(2\pi \bm{W}_1 \bm{x})}]^{\top}\in\mathbb{R}^{2d_1},
\end{equation}
where $\bm{W}_1\in\mathbb{R}^{d_1\times2}$ is a matrix with entries sampled from a normal distribution $\mathcal{N}(0,\sigma_1^2)$ with standard deviation $\sigma_1>0$. The time value $\tau$ is encoded as
\begin{equation}\label{fbe-2}
    \psi_2(\tau) = [\sin{(2\pi \bm{w}_2 \tau)}, \cos{(2\pi \bm{w}_2 \tau)}]^{\top}\in\mathbb{R}^{2d_2},
\end{equation}
where $\bm{w}_2\in\mathbb{R}^{d_2}$ is a vector with entries sampled from a normal distribution $\mathcal{N}(0,\sigma_2^2)$ with standard deviation $\sigma_2>0$. The hyperparameters $d_1$, $d_2$, $\sigma_1^2$ and $\sigma_2^2$ are specified in \cref{experiment}. Thus, the $k$-th components of $H_f$ and $H_g$ described in \cref{Model2-1} can be represented as 
\begin{equation}\label{INRrep}
    f(\cdot,\bm{\theta}_{1k})=\bm{\phi}(\psi_1(\cdot),\bm{\theta}_{1k}),~g(\cdot,\bm{\theta}_{2k})=\bm{\phi}(\psi_2(\cdot),\bm{\theta}_{2k}).
\end{equation}

\subsubsection{Implementation} \label{sr and lr}
We utilize the Adam optimizer to update the neural network parameters. The parameters $\bm{\Theta}_1$ and $\bm{\Theta}_2$ are learned by separate neural networks, each with distinct learning rates $\alpha_1$ and $\alpha_2$. In the initial phase of training, model parameters often reside far from the optimal manifold. Premature regularization constraints may hinder the exploration of critical feature representation. During the first few iterations, we propose to exclude the regularization terms from the loss function for stability. By deferring regularization, the model gains flexibility to rapidly capture coarse-grained patterns without being overly penalized for large parameter updates. Specifically, we set $\lambda_1=\lambda_2=0$ for the first 1000 iterations and then introduce regularization by assigning appropriate values to $\lambda_1$ and $\lambda_2$. In addition, the learning rate follows a step decay schedule with a decay rate of 0.98 for the first 1000 iterations, and then decay with rate 0.95. In summary, pseudo code is given in Algorithm~\ref{alg}.

\begin{algorithm}[ht]
\caption{\texttt{NINRF} for dynamic PET reconstruction}
\label{alg}
    \SetAlFnt
    \SetAlgoLined
    \SetKwData{Left}{left}\SetKwData{This}{this}\SetKwData{Up}{up}\SetKwFunction{Union}{Union}\SetKwFunction{FindCompress}{FindCompress}\SetKwInOut{Input}{Input}\SetKwInOut{Output}{Output}
    \textbf{Hyperparameters}: maximal iterations $I_{\text{max}}$, regularization parameters $\lambda_1$ and $\lambda_2$, positional encoding dimension $d_1$ and $d_2$, standard deviation $\sigma_1$ and $\sigma_2$, and initial step size $\alpha_1$ and $\alpha_2$ \;
    \textbf{Input}: space grid points $\Omega_{h,w}$ and time grid points $\mathcal{D}_T$, and sinogram data $\bm{Z}\in\mathbb{R}^{n_an_l \times T}$\;
    \textbf{Initialize}: $i=0$, initialize $\bm{\Theta}_1=\{\bm{\theta}_{11},\dots,\bm{\theta}_{1K}\}$ and $\bm{\Theta}_2=\{\bm{\theta}_{21},\dots,\bm{\theta}_{2K}\}$ using Kaiming uniform distribution~\cite{he2015delving}\;
    Sample the matrix $\bm{W}_1$ in~\eqref{fbe-1} and vector $\bm{w}_2$ in~\eqref{fbe-2}\;
    \While{$i<I_{\text{max}}$}{
        Compute spatial bases matrix $\bm{A}_{\bm{\Theta}_1}$ and temporal bases matrix $\bm{B}_{\bm{\Theta}_2}$ in~\eqref{MAT}\;
        \textcolor{blue}{
        \If{$i<1000$}
        {Set $\lambda_1=0$, $\lambda_2=0$\;}
        }
        % {Set regularization parameters $\lambda_1>0$,$\lambda_2>0$\;} 
        Update $\bm{\Theta}_1$ and $\bm{\Theta}_2$ by minimizing \eqref{reconstruction loss} via Adam optimizer with step size $\alpha_1$ and $\alpha_2$, respectively\;
        Decay step size $\alpha_1$ and $\alpha_2$\;
        $i \leftarrow i+1$\;
    }
    \textbf{Output}: reconstructed dynamic PET $\widehat{u}(\bm{x},t) =\bm{A}(\bm{x};\widehat{\bm{\Theta}}_1)^\top\bm{B}(t;\widehat{\bm{\Theta}}_2),~\forall (\bm{x},t)\in\Omega\times[0,1]$.
\end{algorithm}

\section{Experimental Results}\label{experiment}
We implement the methods in Pytorch interface on a NVIDIA Titan GPU.

\subsection{Methods in Comparison}
To evaluate the performance of our proposed method for dynamic PET reconstruction, we compare it against several representative methods, including \texttt{EM}~\cite{shepp1982maximum}, \texttt{EM-NMF}, \texttt{MAP-TV}, \texttt{DIP-B}~\cite{yokota2019dynamic}, and \texttt{INR-B}.
\begin{itemize}
    \item \texttt{EM}: For each frame $\bm{u}$ in the dynamic PET image, EM algorithm is used to perform maximum likelihood estimation~\cite{shepp1982maximum,lange1984reconstruction}. Given the sinogram $\bm{z}$ and the projection matrix $\bm{P}$, the update scheme is
    \begin{equation*}
        \bm{u}^{(k+1)} = \frac{\bm{u}^{(k)}}{\bm{P}^{\top}\bm{1}} \circ \bm{P}^{\top}\frac{\bm{z}}{\bm{P}\bm{u}^{(k)}+\bm{S}},
    \end{equation*}
    where $\circ$ represents Hadamard product and the division is element-wise. All the reconstructed frames constitute the dynamic PET image.
    \item \texttt{EM-NMF}: This method incorporates the NMF model with maximum likelihood estimation. The dynamic PET image $\bm{U}$ can be represented by \eqref{NMF_disc}, and $D_{\text{KL}}(\bm{Z}\Vert\bm{P}\bm{U}+\bm{S})$ is minimized where sinogram $\bm{Z}$ and $\bm{P}$ are given. EM algorithm is applied to alternately update $\bm{A}$ and $\bm{B}$. The update equations are: 
    \begin{equation*}
        \begin{aligned}
            \bm{A}^{(k+1)} &= \bm{A}^{(k)} \circ \frac{\bm{P}^{\top}(\frac{\bm{Z}}{\bm{P}\bm{A}^{(k)}\bm{B}^{(k)}+\bm{S}}){\bm{B}^{(k)}}^{\top}}{\bm{P}^{\top}\bm{1}{\bm{B}^{(k)}}^{\top}}, \\
            \bm{B}^{(k+1)} &= \bm{B}^{(k)} \circ \frac{(\bm{P}\bm{A}^{(k+1)})^{\top} \frac{\bm{Z}}{\bm{P}\bm{A}^{(k+1)}\bm{B}^{(k+1)}+\bm{S}}}{(\bm{P}\bm{A}^{(k+1)})^{\top}\bm{1}}.
        \end{aligned}
    \end{equation*}
    \item \texttt{MAP-TV} This method uses MAP estimation and minimizes the KL divergence with spatial TV regularization for every frame $\bm{u}_i$ and time quadratic variation for each pixel $\bm{v}_j$, where $\bm{u}_i$ and $\bm{v}_j$ are the $i$-th column and $j$-th row of $\bm{U}$, respectively. Given the sinogram $\bm{Z}$ and the projection matrix $\bm{P}$, the optimization problem is
    \begin{equation*}
        \min_{\bm{U}\ge 0} D_{\text{KL}}(\bm{Z}\Vert\bm{P}\bm{U}+\bm{S}) + \lambda_{\text{TV}_1}\sum_{i=1}^{T}\text{TV}(\bm{u}_i) + \lambda_{\text{TV}_2}\sum_{j=1}^{hw}\|\nabla_t(\bm{v}_j)\|_2^2,
    \end{equation*}
    which can be solved using the ADMM algorithm \cite{boyd2011distributed, wen2016primal}.
    \item \texttt{DIP-B}: This method uses NMF incorporated with DIP as proposed in \cite{yokota2019dynamic}. Additionally, the TV norm of $\bm{A}$ is used for regularization similar to $R_1(\bm{A})$ in~\eqref{reconstruction loss}.
    \item \texttt{INR-B}: In this approach, $\bm{A}$ is modeled using INR, while $\bm{B}$ is optimized using the EM-based update rule which is same as \texttt{DIP}. It is a compromise between \texttt{DIP-B} and our proposed method.
    % \item \texttt{NINRF}: Our proposed method.
\end{itemize}

\begin{figure}[htbp!]
    \centering
    \centering
    \begin{minipage}[t]{0.2\textwidth}
    \subfigure[Phantom]{
    \label{TACshowa}
    \includegraphics[width=0.88\textwidth]{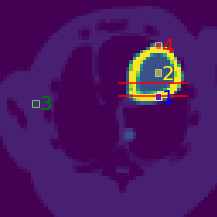}
    }
    \end{minipage}
    \begin{minipage}[t]{0.3\textwidth}
    \subfigure[TACs]{
    \includegraphics[width=0.92\textwidth]{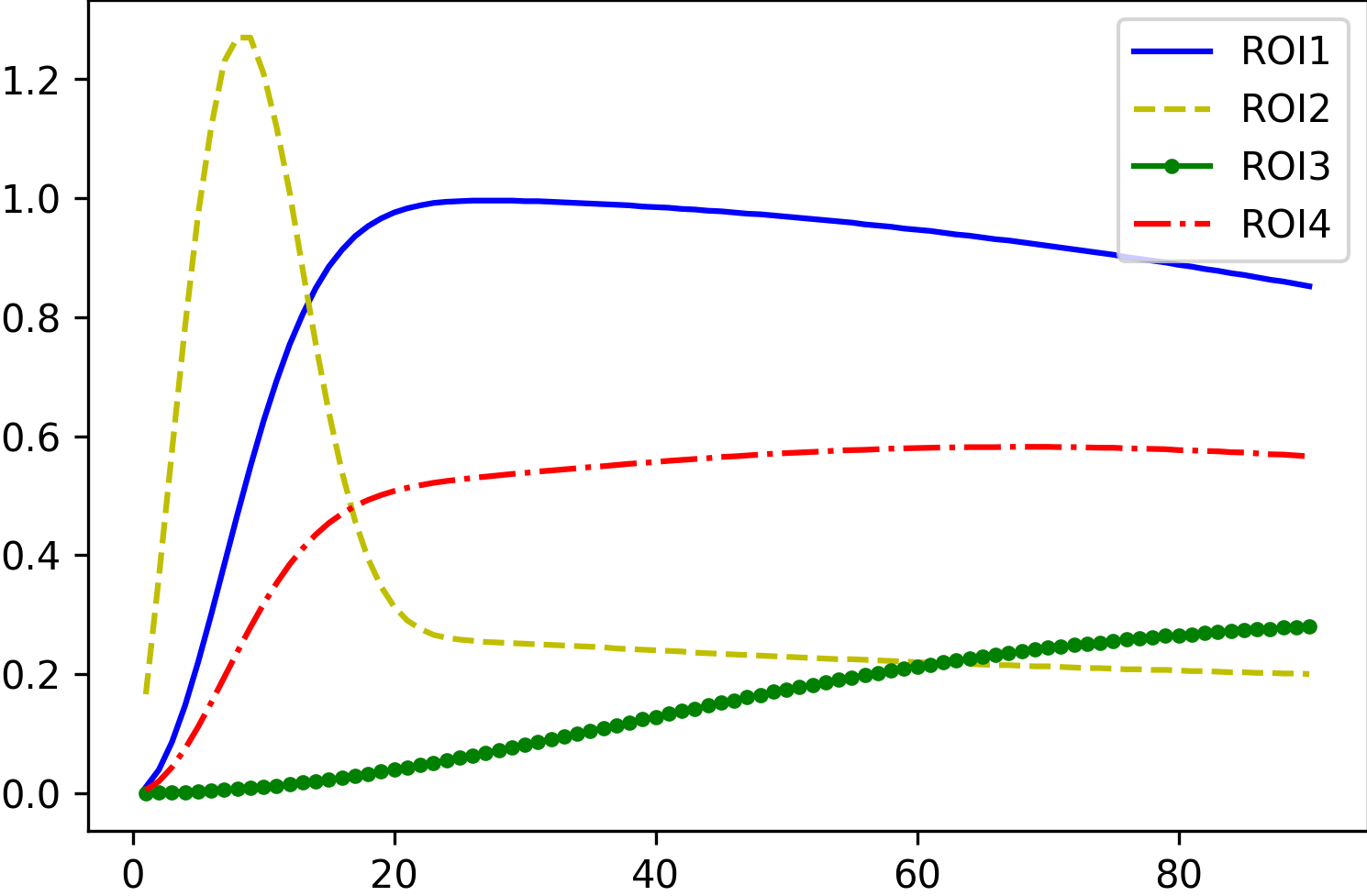}
    \label{TACshowb}
    }
    \end{minipage}
    \begin{minipage}[t]{0.4\textwidth}
    \subfigure[sinogram without noise]{
    \includegraphics[width=.85\textwidth]{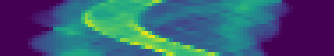}
    \label{sinoshow1}
    }
    \subfigure[sinogram with Poisson noise (SNR=20dB)]{
    \includegraphics[width=.85\textwidth]{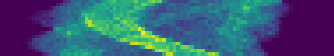}
    \label{sinoshow2}
    }
    \end{minipage}
    \caption{The data of rat abdomen. (a): The simulated image phantom of rat abdomen; (b): The  TACs for different ROIs annotated in (a) (The horizontal axis represents time, while the vertical axis denotes the intensity); (c) and (d): The sinogram without and with noise, where the number of projection angles $n_a=16$.}
    \label{TACshow}
\end{figure}

\subsection{Dynamic PET Reconstruction on Simulated Data}

We present the experimental results for reconstruction tasks using two different types of data: simulated rat abdomen and human brain. 

\subsubsection{Simulated Rat Abdomen Image Reconstruction}\label{exp_rat}
We conduct experiments using synthetic data based on an image phantom of a simulated rat abdomen ($h=w=64$), containing four regions of interest (ROIs) as shown in \cref{TACshowa}. The image intensity values in different ROIs are modulated according to TACs shown in \cref{TACshowb} over $T=90$ frames. We set the expectation of randoms and scatters to be 0. The sinogram is generated by applying the Radon transform to each frame, followed by the introduction of Poisson noise. The Radon transform is performed with $n_a = 16$ projection angles evenly spaced between 0 and 180 degrees. The length of every sinogram is $n_l=95$. The sinograms before and after adding Poisson noise for a given frame are shown in ~\cref{sinoshow1} and \cref{sinoshow2}. The objective is to reconstruct the image from the noisy sinogram.

% \noindent\textbf{Model Setting}.
For the methods \texttt{DIP-B}, \texttt{INR-B} and \texttt{NINRF}, which all employ the NMF model, we set the model rank $K=5$. The network architecture and training strategy used for \texttt{DIP-B} are identical to those in \cite{yokota2019dynamic}. For the proposed method \texttt{NINRF}, the network consists of 3 hidden layers, each with 256 units. The hyperparameters for positional encoding are set to $d_1=d_2=256$ and $\sigma_1=\sigma_2=8$. The initial learning rate $\alpha_1$ and $\alpha_2$ are both set to $5 \times 10^{-4}$. For the method \texttt{INR-B}, the network structure is identical to that in \texttt{NINRF}. The experiments are conducted with different noise levels in the observed sinogram, with signal-to-noise ratios (SNR) set at 30dB, 20dB, 10dB and 5dB. The regularization parameters of \texttt{MAP-TV} for different noise levels from low to high are $\lambda_{\text{TV}_1}=0.01,0.05,0.2,0.5$ and $\lambda_{\text{TV}_2}=5,30,120,150$. Regularization terms of $\bm{A}$ and $\bm{B}$ are all considered in \texttt{DIP-B}, \texttt{INR-B} and \texttt{NINRF}. In \texttt{DIP-B} and \texttt{INR-B} for consistency, and the  coefficients are $\lambda_1=0.1,0.8,5,5$ and $\lambda_2=1,10,50,50$ . In \texttt{NINRF}, the coefficients are $\lambda_1=4,10,50,50$ and $\lambda_2=0.1,0.5,5,5$.

\begin{figure}[htbp!]
    \centering
    \resizebox{\textwidth}{!}{
    \begin{tabular}{c@{\hspace{2pt}}c@{\hspace{1pt}}c@{\hspace{1pt}}c@{\hspace{1pt}}c@{\hspace{1pt}}c@{\hspace{1pt}}c}
    &
   	\includegraphics[width=.15\linewidth]{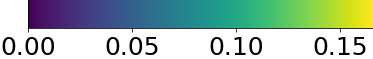}&
		\includegraphics[width=.15\linewidth]{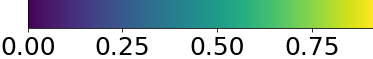}&
		\includegraphics[width=.15\linewidth]{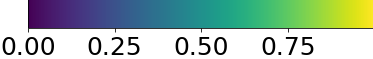}&
		\includegraphics[width=.15\linewidth]{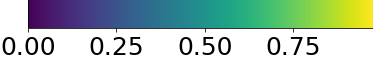}&
		\includegraphics[width=.15\linewidth]{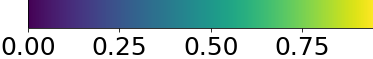}&
		\includegraphics[width=.15\linewidth]{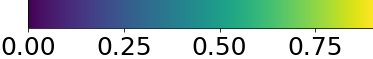}\\
  \put(-5,10){\rotatebox{90}{\scriptsize{Truth}}}&
		\includegraphics[width=.15\linewidth]{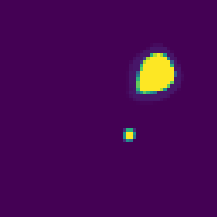}&
		\includegraphics[width=.15\linewidth]{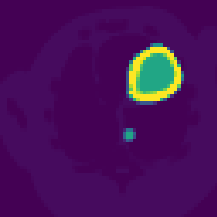}&
		\includegraphics[width=.15\linewidth]{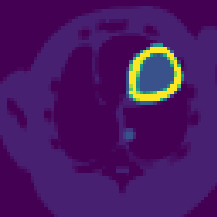}&
		\includegraphics[width=.15\linewidth]{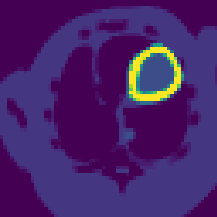}&
		\includegraphics[width=.15\linewidth]{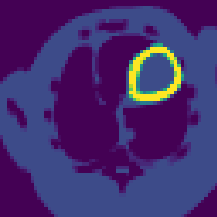}&
		\includegraphics[width=.15\linewidth]{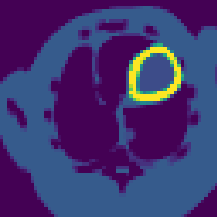}\\
  \put(-8,10){ \rotatebox{90}{\scriptsize{\texttt{EM}}}}&
	\includegraphics[width=.15\linewidth]{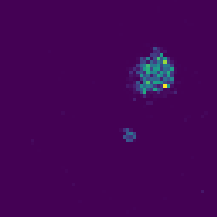}&
		\includegraphics[width=.15\linewidth]{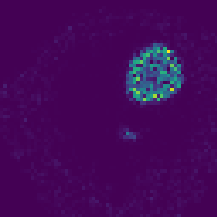}&
		\includegraphics[width=.15\linewidth]{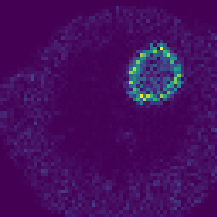}&
		\includegraphics[width=.15\linewidth]{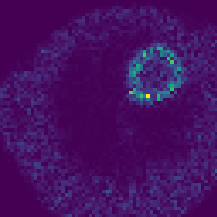}&
		\includegraphics[width=.15\linewidth]{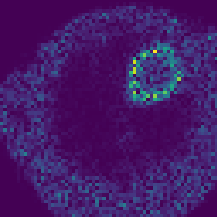}&
		\includegraphics[width=.15\linewidth]{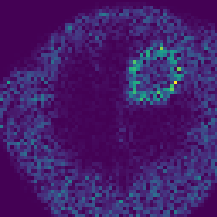}\\
 \put(-5,10){\rotatebox{90}{\scriptsize{\texttt{EM-NMF}}}}&
		\includegraphics[width=.15\linewidth]{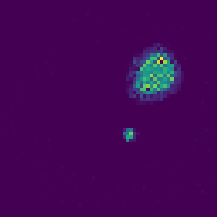}&
		\includegraphics[width=.15\linewidth]{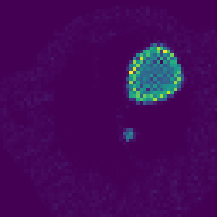}&
		\includegraphics[width=.15\linewidth]{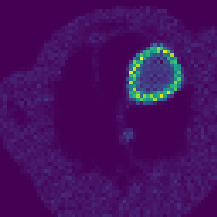}&
		\includegraphics[width=.15\linewidth]{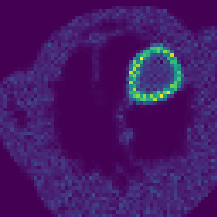}&
		\includegraphics[width=.15\linewidth]{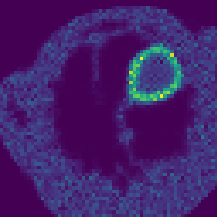}&
		\includegraphics[width=.15\linewidth]{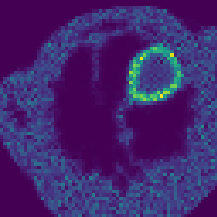}\\
\put(-5,10){\rotatebox{90}{\scriptsize{\texttt{MAP-TV}}}}&
		\includegraphics[width=.15\linewidth]{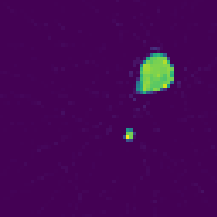}&
		\includegraphics[width=.15\linewidth]{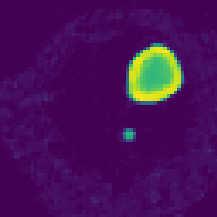}&
		\includegraphics[width=.15\linewidth]{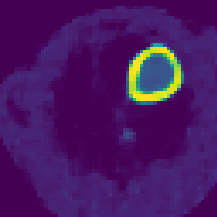}&
		\includegraphics[width=.15\linewidth]{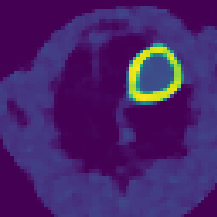}&
		\includegraphics[width=.15\linewidth]{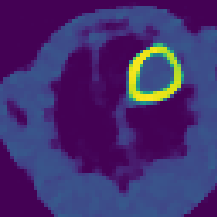}&
		\includegraphics[width=.15\linewidth]{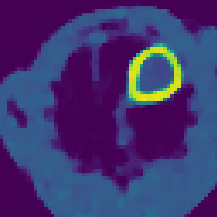}\\        
\put(-5,10){\rotatebox{90}{\scriptsize{\texttt{DIP-B}}}}&
		\includegraphics[width=.15\linewidth]{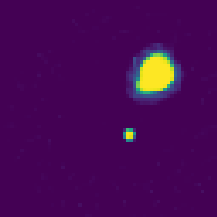}&
		\includegraphics[width=.15\linewidth]{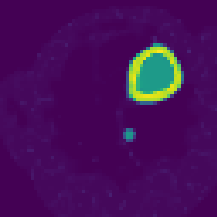}&
		\includegraphics[width=.15\linewidth]{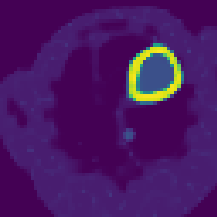}&
		\includegraphics[width=.15\linewidth]{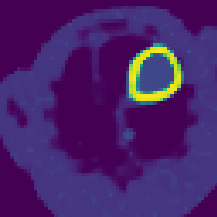}&
		\includegraphics[width=.15\linewidth]{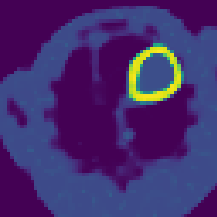}&
		\includegraphics[width=.15\linewidth]{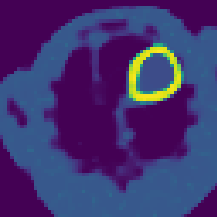}\\
 \put(-5,10){\rotatebox{90}{\scriptsize{\texttt{INR-B}}}}&
		\includegraphics[width=.15\linewidth]{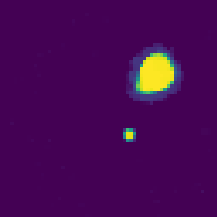}&
		\includegraphics[width=.15\linewidth]{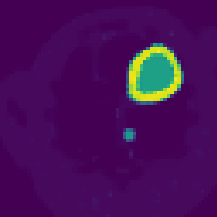}&
		\includegraphics[width=.15\linewidth]{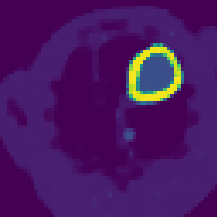}&
		\includegraphics[width=.15\linewidth]{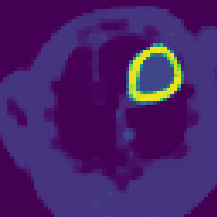}&
		\includegraphics[width=.15\linewidth]{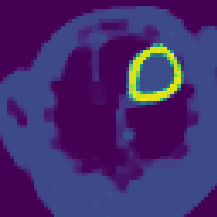}&
		\includegraphics[width=.15\linewidth]{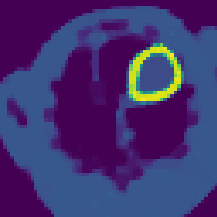}\\
\put(-5,10){\rotatebox{90}{\scriptsize{\texttt{NINRF}}}}&
		\includegraphics[width=.15\linewidth]{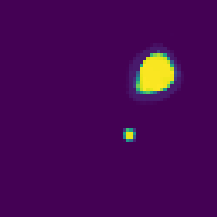}&
		\includegraphics[width=.15\linewidth]{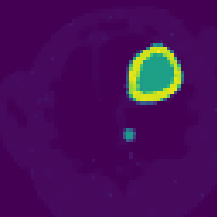}&
		\includegraphics[width=.15\linewidth]{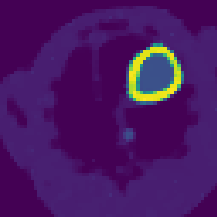}&
		\includegraphics[width=.15\linewidth]{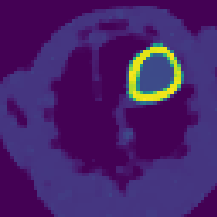}&
		\includegraphics[width=.15\linewidth]{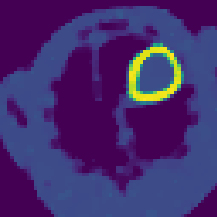}&
		\includegraphics[width=.15\linewidth]{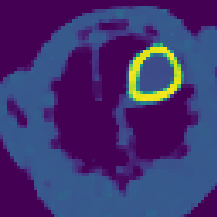}\\
  &Frame 1& Frame 16&Frame 31&Frame 46&Frame 61&Frame 76
    \end{tabular}
    }
\caption{Reconstructed dynamic PET images at Frame 1, 16, 31, 46, 61, and 76 of the rat abdomen using different methods. (sinogram SNR=20)}
\label{rec32_a90_snr20_num16}
\end{figure}

\begin{figure}[htbp!]
    \centering
    \subfigure[PSNR]{
    \includegraphics[width=0.3\textwidth]{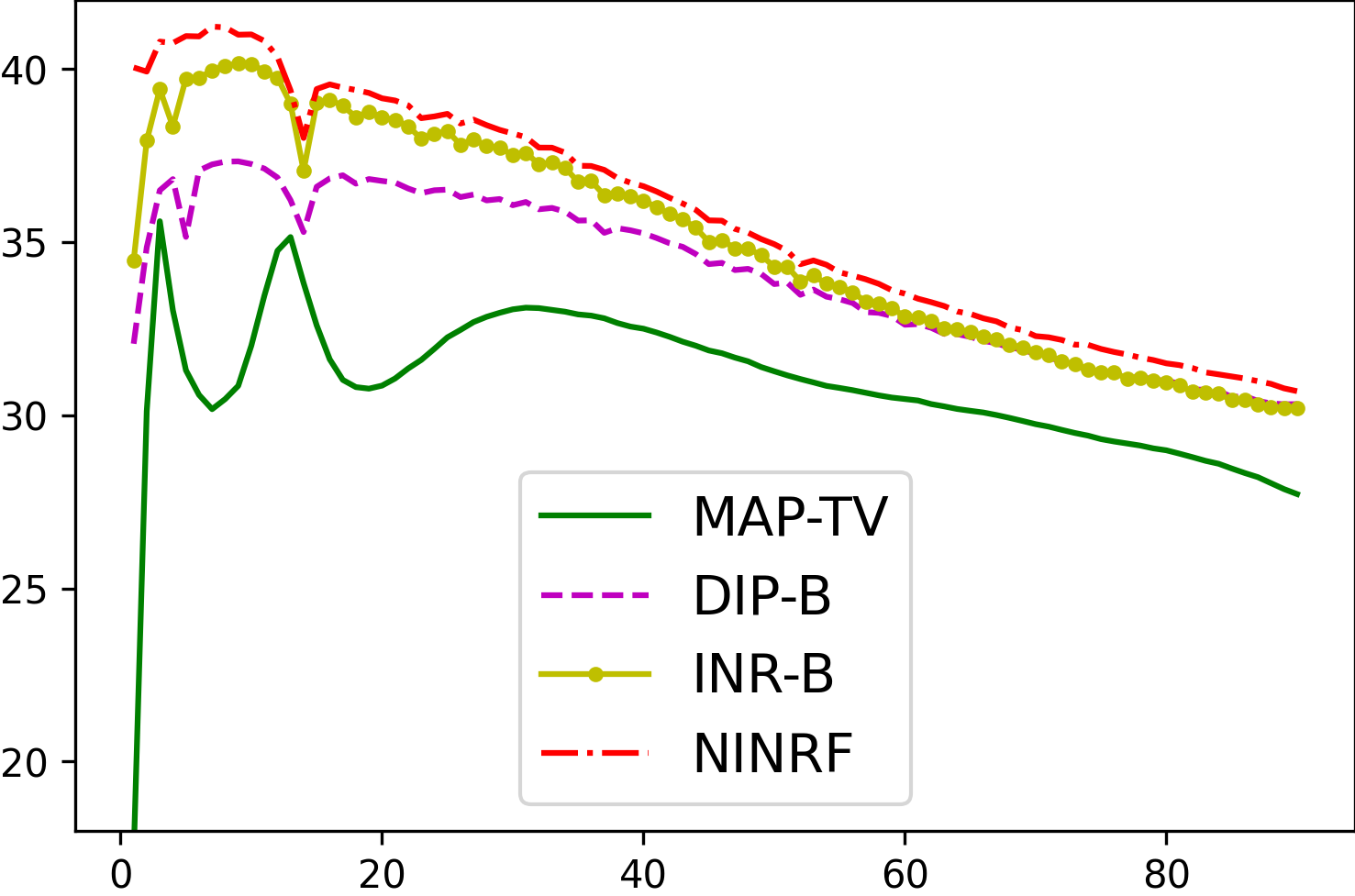}
    }
    \subfigure[SSIM]{
    \includegraphics[width=0.3\textwidth]{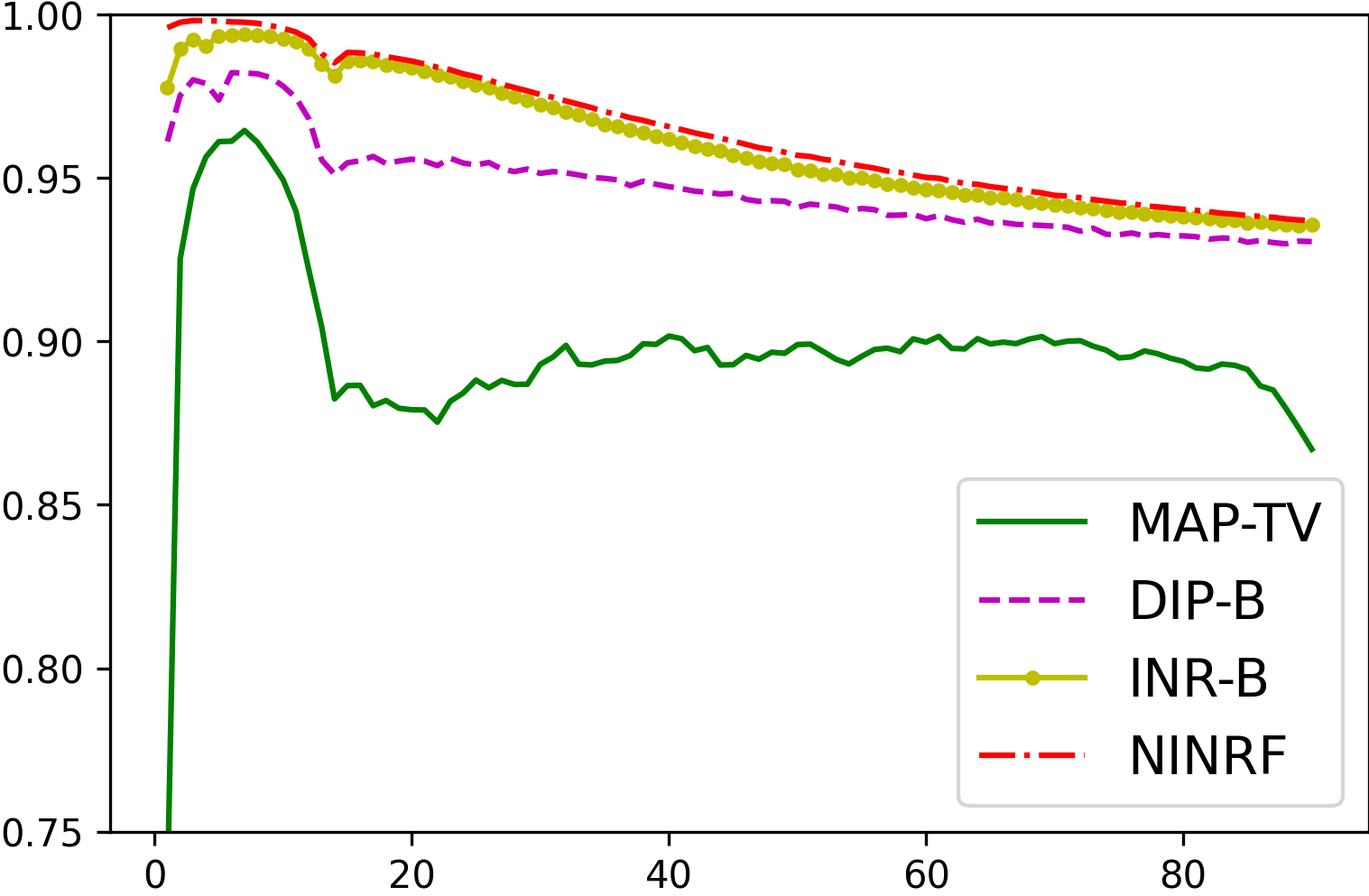}
    }
    \caption{PSNR and SSIM of every frame of the reconstructed rat abdomen image(sinogram SNR=20). The horizontal axis represents time, while the vertical axis denotes the corresponding values.}
    \label{PSNRSSIM_a90_snr20_num16}
\end{figure}

\begin{figure}[htbp!]
    \centering
    % \subfigure[PSNR]{
    % \label{PSNR_a90_snr20_num16}
    % \includegraphics[width=0.21\textwidth]{img/rec/PSNRnum16a90snr/PSNR_num=16_snr=20.png}
    % }
    % \subfigure[SSIM]{
    % \label{SSIM_a90_snr20_num16}
    % \includegraphics[width=0.21\textwidth]{img/rec/SSIMnum16a90snr/SSIM_num=16_snr=20.png}
    % }
    \subfigure[Line 1]{
    \label{rec_profile1}
    \includegraphics[width=0.3\textwidth]{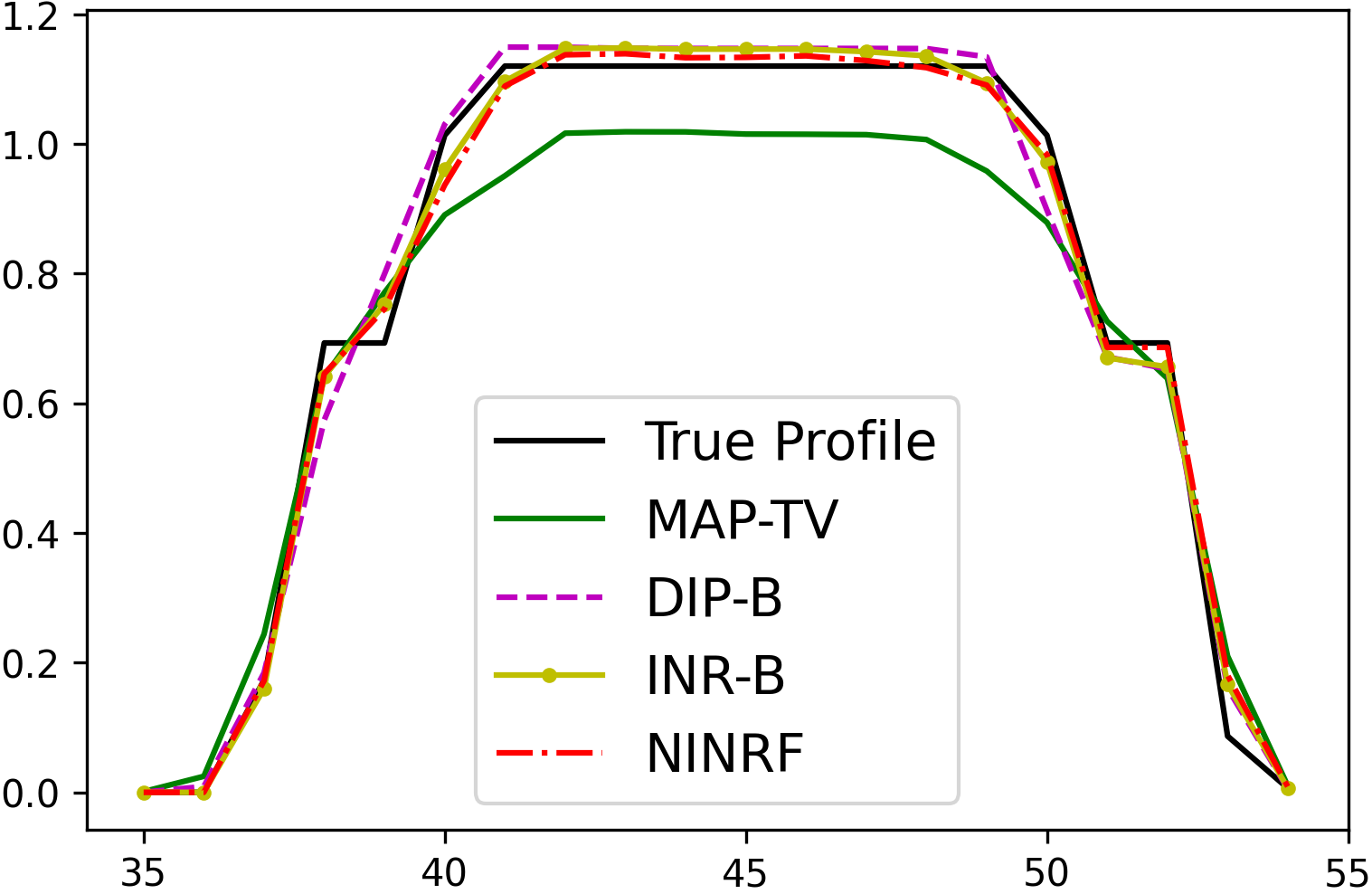}
    }
    \subfigure[Line 2]{
    \label{rec_profile2}
    \includegraphics[width=0.3\textwidth]{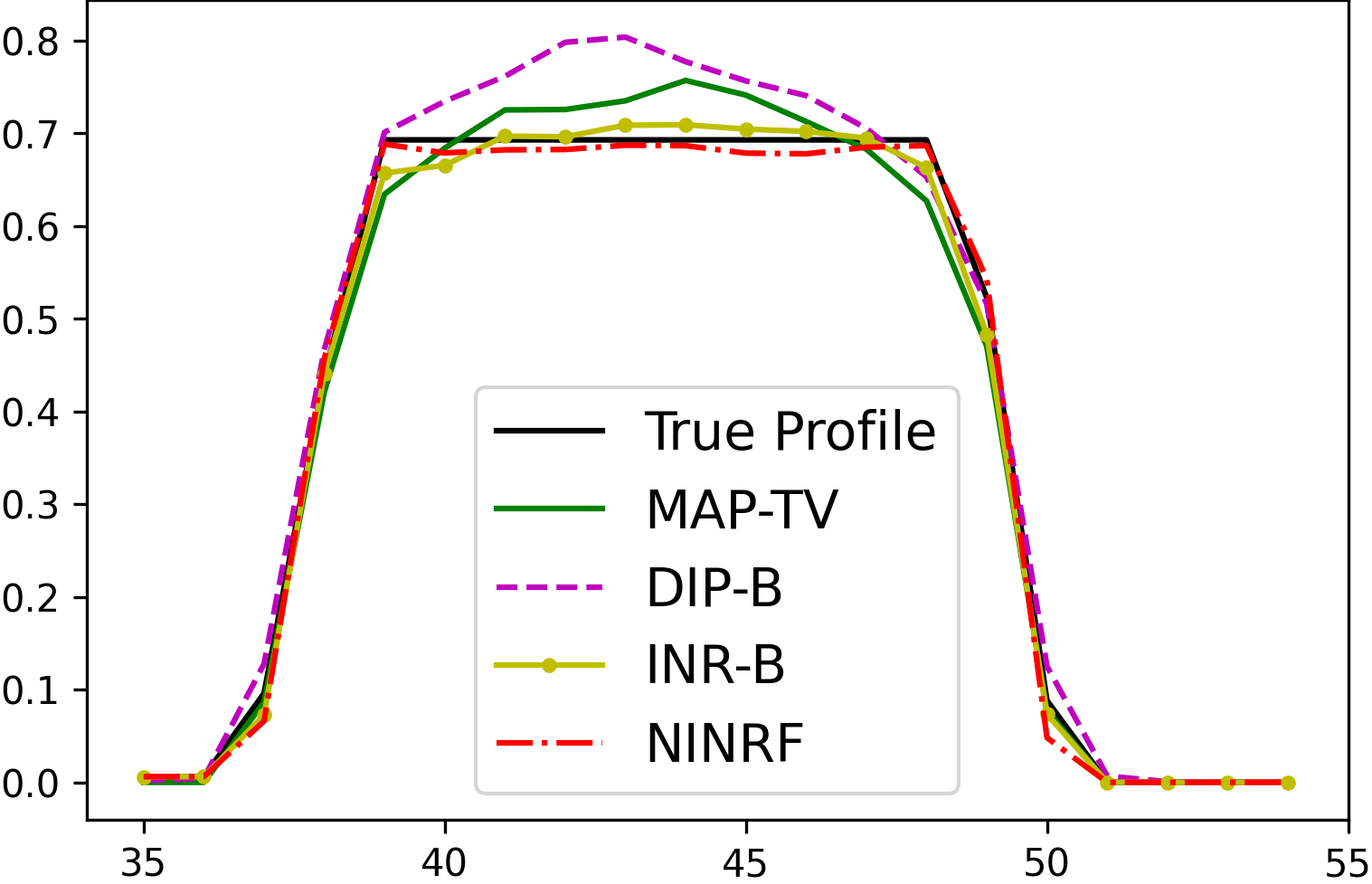}
    }
    \centering
    \subfigure[TAC of ROI1]{
    \label{TAC1_a90_snr20_num16}
    \includegraphics[width=0.3\textwidth]{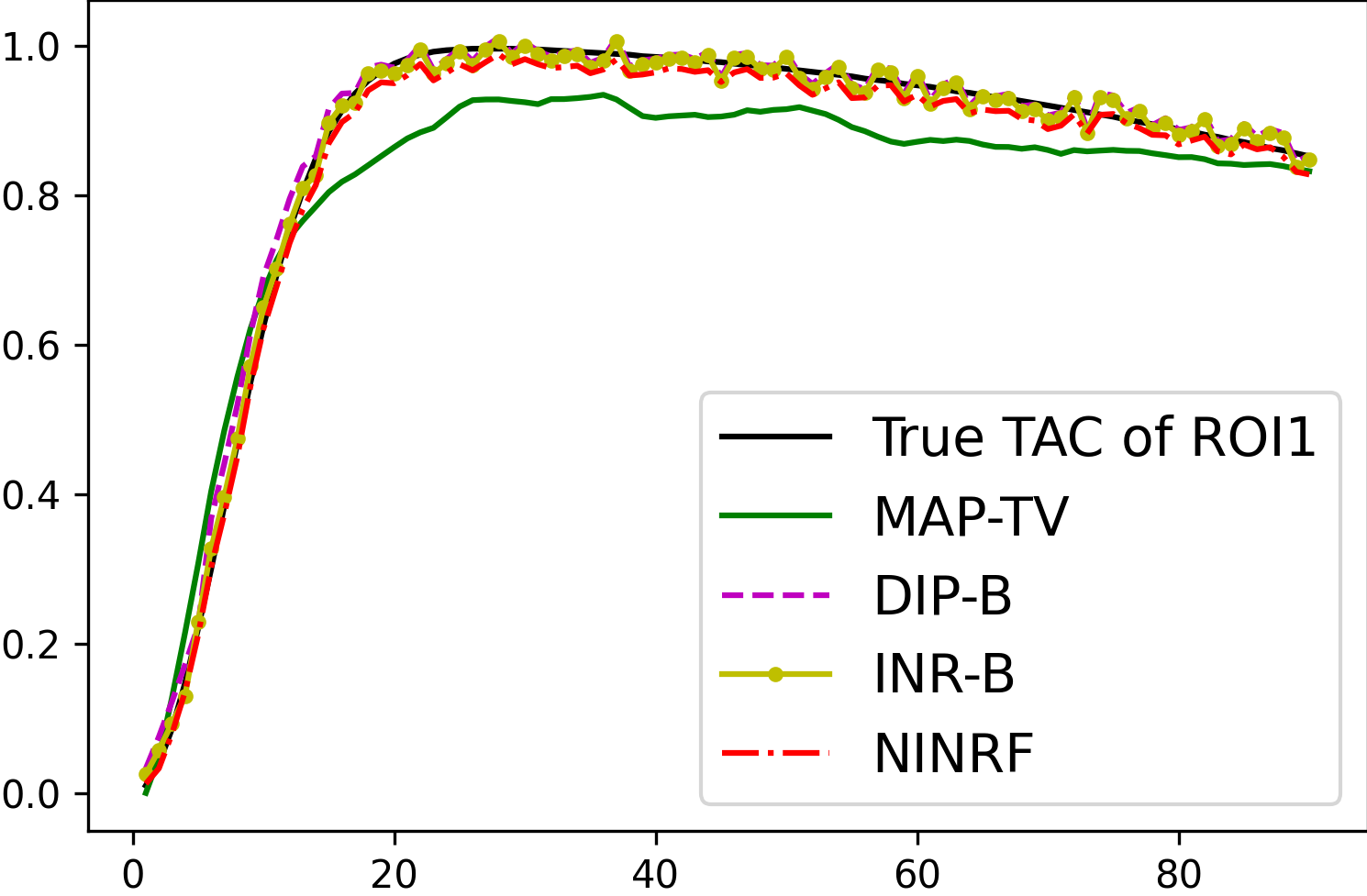}
    }
    \subfigure[TAC of ROI2]{
    \label{TAC2_a90_snr20_num16}
    \includegraphics[width=0.3\textwidth]{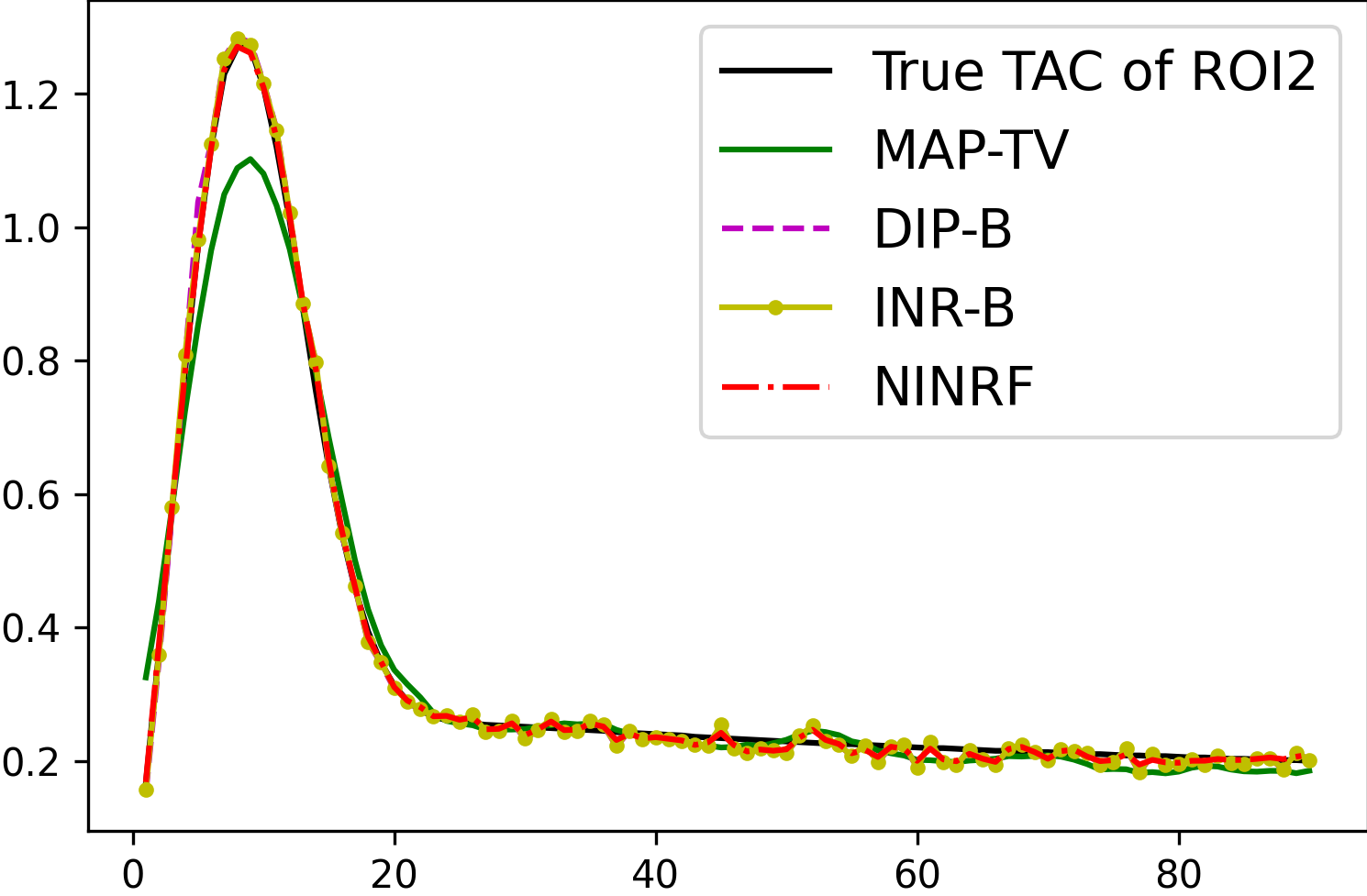}
    }
    \subfigure[TAC of ROI3]{
    \label{TAC3_a90_snr20_num16}
    \includegraphics[width=0.3\textwidth]{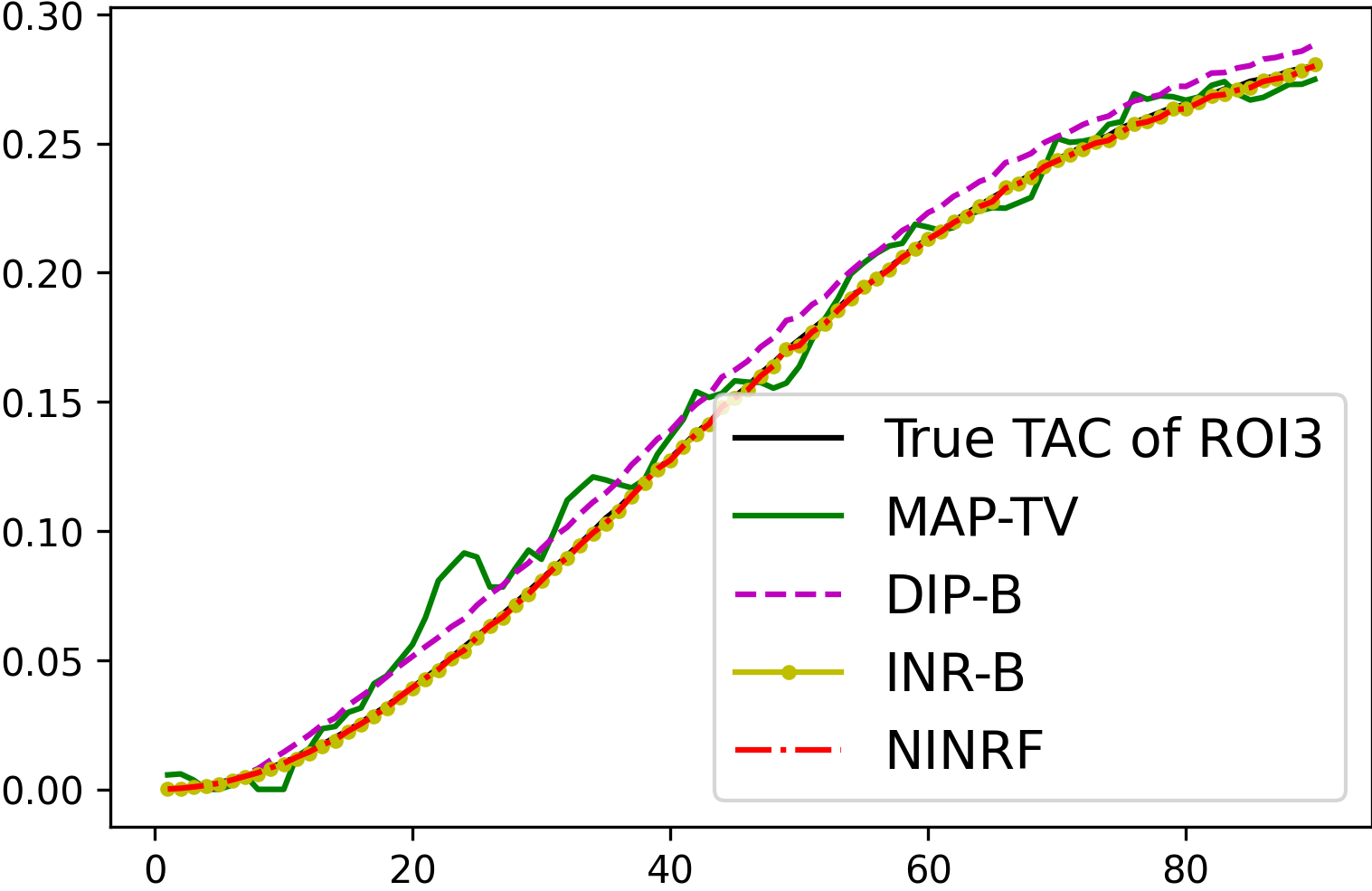}
    }
    \subfigure[TAC of ROI4]{
    \label{TAC4_a90_snr20_num16}
    \includegraphics[width=0.3\textwidth]{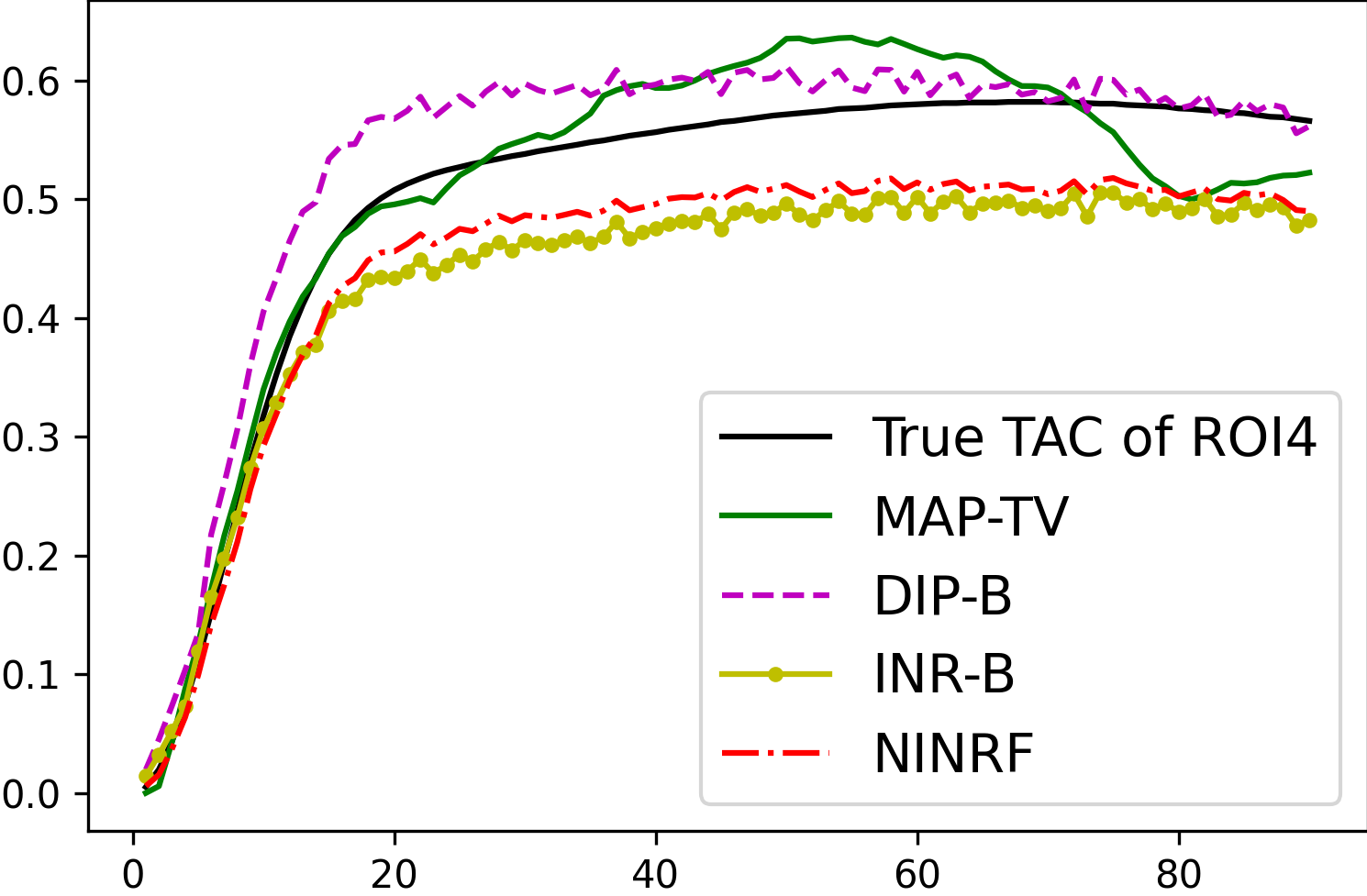}
    }
    \caption{Intensity profiles, and TACs of the reconstructed rat abdomen image(sinogram SNR=20). (a) and (b): Intensity profiles of Frame 11 along the red lines in \cref{TACshowa}, the horizontal axis represents the horizontal position coordinates, while the vertical axis denotes the corresponding intensity values; (c) to (f): TACs at the locations in different ROIs shown in \cref{TACshowa}, the horizontal axis represents the time, while the vertical axis denotes the corresponding values.}
    \label{PSpT_a90_snr20_num16}
\end{figure}

% \noindent\textbf{Numerical Results}
In \cref{rec32_a90_snr20_num16}, we present frames of the reconstructed image from the sinogram with SNR of 20dB at different time points. Treating the whole dynamic PET image as a signal, we compute the peak signal-to-noise ratio (PSNR) and structural similarity index measure (SSIM) of the reconstructed image for each reconstruction method, which are summarized in \cref{SNRnum16}. Our proposed method \texttt{NINRF} achieves the highest scores in both PSNR and SSIM, demonstrating superior image quality. Since the results of \texttt{EM} and \texttt{EM-NMF} are notably poor, we omit them from further discussions. To provide a more detailed evaluation, we compute the PSNR and SSIM for each frame and present them in \cref{PSNRSSIM_a90_snr20_num16}. As shown, \texttt{NINRF} consistently achieves the best PSNR and SSIM across all frames. \cref{rec_profile1} and \cref{rec_profile2} show the profile comparisons. In the Frame 11, the profile of \texttt{NINRF} along the red lines is closest to the true profile. Additionally, we reconstruct the TACs for each method using the mean values within the marked regions shown in \cref{TACshowa}. The corresponding reconstructed TACs are displayed in \cref{TAC1_a90_snr20_num16} through \cref{TAC4_a90_snr20_num16}. We can see that \texttt{NINRF} achieves the comparable results with the other methods in ROI1 and ROI2, and outperforms the other methods in ROI3. 
ROI4 is the transitional zone between ROI1 and ROI2. All the methods can not get a perfect result in this specific region, while  our  method preserves the right tendency for the TAC, \textit{i.e.} increasing rapidly and then keeping stable after Frame 20.
%In ROI1 and ROI2, the results from \texttt{NINRF}, \texttt{INR-B} and \texttt{DIP-B} are nearly identical. In ROI3, \texttt{NINRF} and \texttt{INR-B} yield much better results compared to other methods and \texttt{NINRF} outperforms \texttt{INR-B}. In ROI4, \texttt{NINRF} preserves the correct TAC dynamics: increase first then stabilize after around frame 20.

\begin{figure}[htbp!]
    \centering
    \resizebox{\textwidth}{!}{
    \begin{tabular}{c@{\hspace{2pt}}c@{\hspace{1pt}}c@{\hspace{1pt}}c@{\hspace{1pt}}c@{\hspace{1pt}}c@{\hspace{1pt}}c}
		\put(-15,10){\rotatebox{90}{30dB}}&
        \includegraphics[width=2cm]{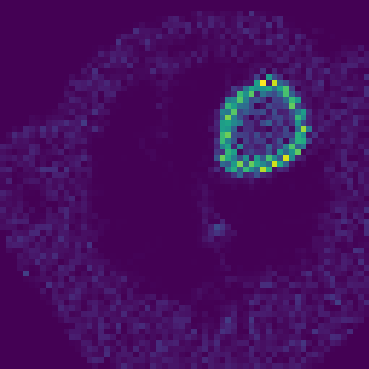}&
		\includegraphics[width=2cm]{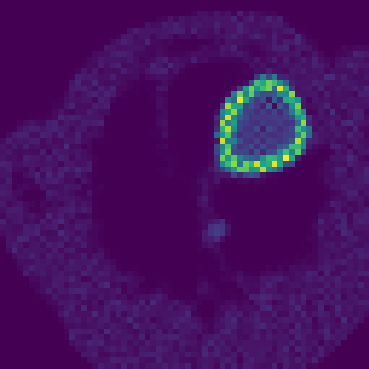}&
		\includegraphics[width=2cm]{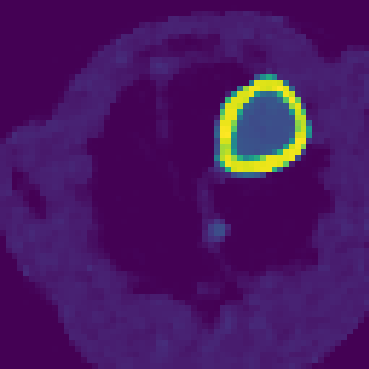}&
		\includegraphics[width=2cm]{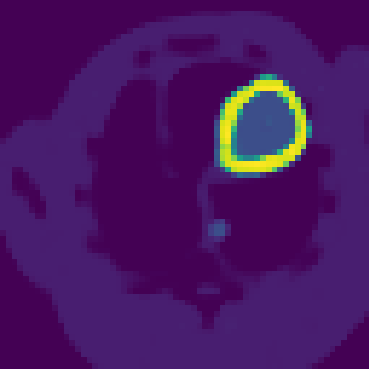}&
		\includegraphics[width=2cm]{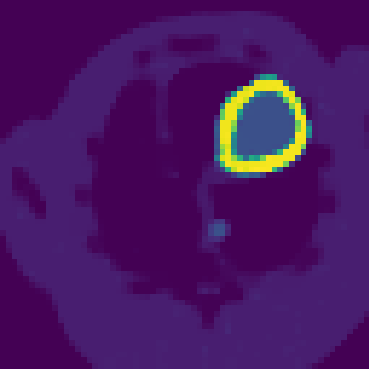}&
		\includegraphics[width=2cm]{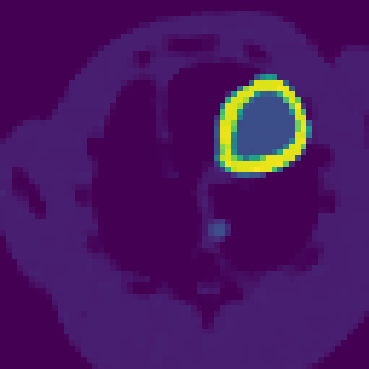}\\
        \put(-15,10){\rotatebox{90}{20dB}}&
	\includegraphics[width=2cm]{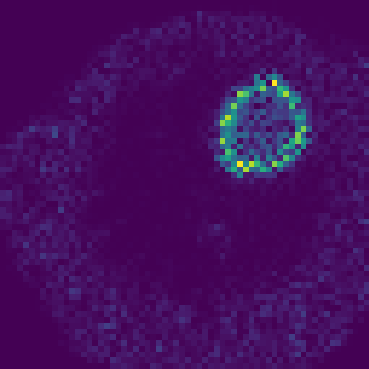}&
		\includegraphics[width=2cm]{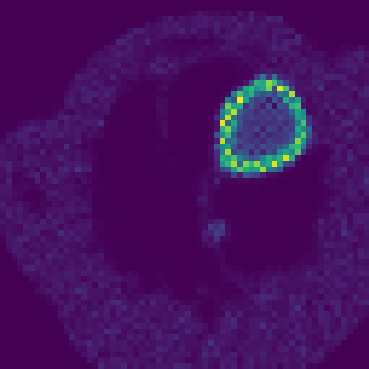}&
		\includegraphics[width=2cm]{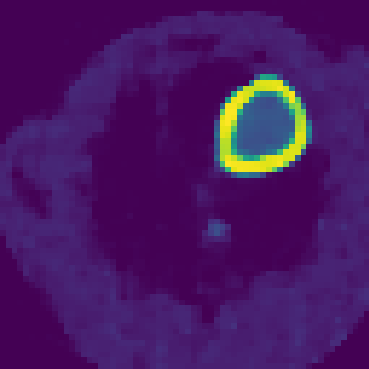}&
		\includegraphics[width=2cm]{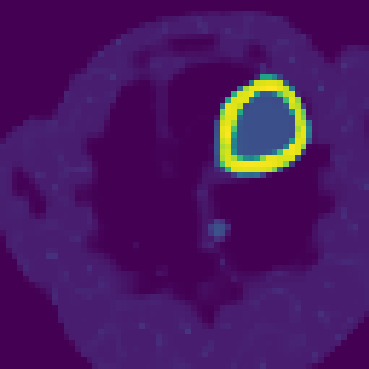}&
		\includegraphics[width=2cm]{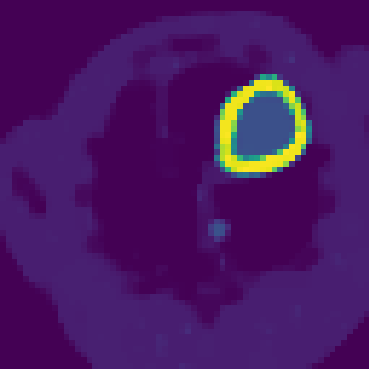}&
		\includegraphics[width=2cm]{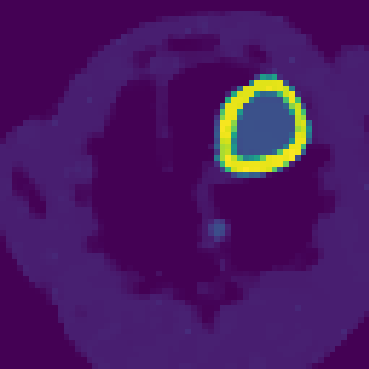}\\
  \put(-15,10){\rotatebox{90}{10dB}}&
   \includegraphics[width=2cm]{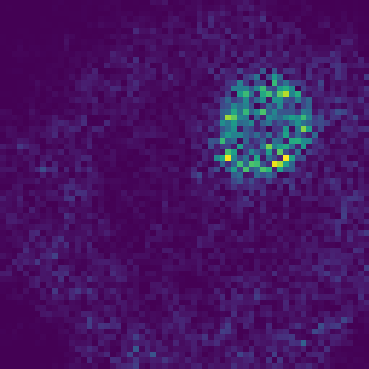}&
		\includegraphics[width=2cm]{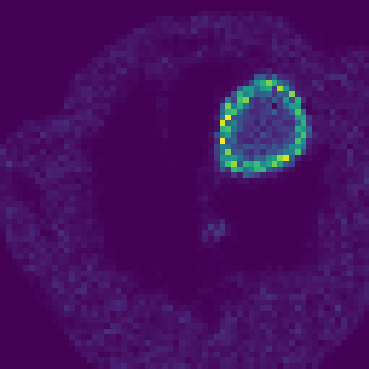}&
		\includegraphics[width=2cm]{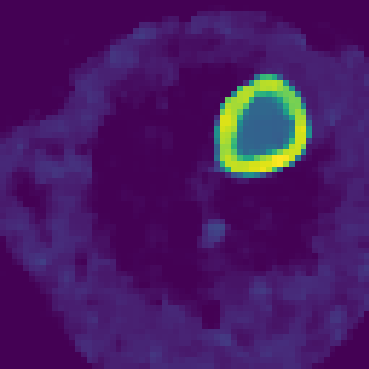}&
		\includegraphics[width=2cm]{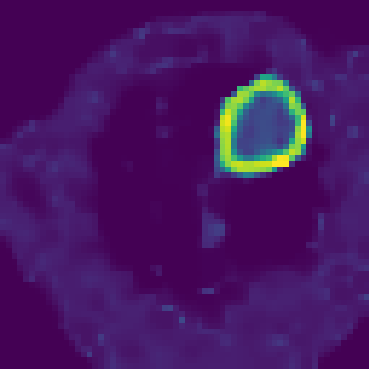}&
		\includegraphics[width=2cm]{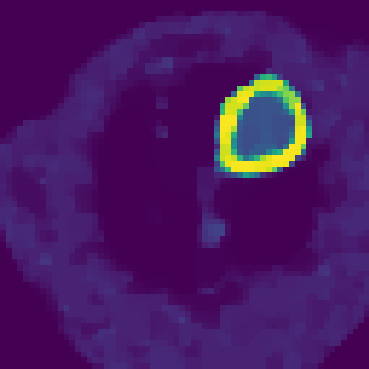}&
		\includegraphics[width=2cm]{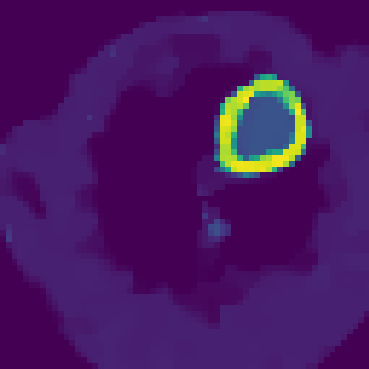}\\
    \put(-15,10){\rotatebox{90}{5dB}}&
   \includegraphics[width=2cm]{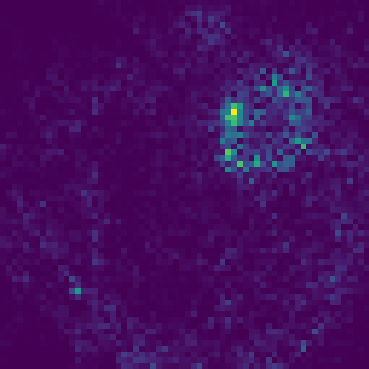}&
		\includegraphics[width=2cm]{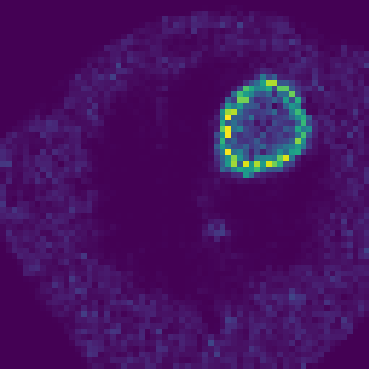}&
		\includegraphics[width=2cm]{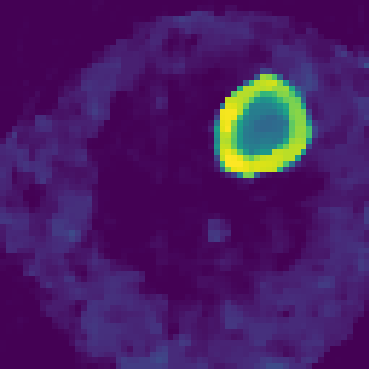}&
		\includegraphics[width=2cm]{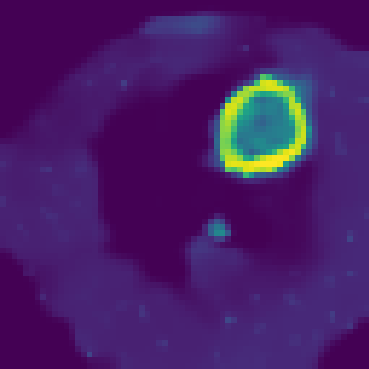}&
		\includegraphics[width=2cm]{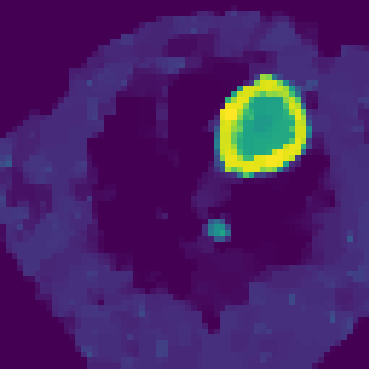}&
		\includegraphics[width=2cm]{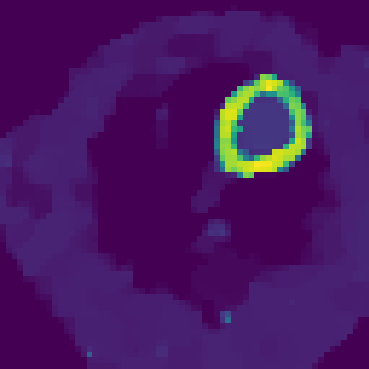}\\
    &\texttt{EM}& \texttt{EM-NMF}&\texttt{MAP-TV}& \texttt{DIP-B}& \texttt{INR-B}& \texttt{NINRF}
    \end{tabular}
    }  
\caption{Reconstruction of the dynamic PET image of the rat abdomen at Frame 31 under various levels of noise and using different methods. The proposed method outperform the others in preserving the intensity levels and details of structures.}
\label{rec32_a90_t61_num16}
\end{figure}

\noindent\textbf{Reconstruction from different noise level.} 
For sinograms with different levels of noise (SNR=30dB, 20dB, 10dB and 5dB), we present Frame 31 of the reconstructed image obtained from all the methods in \cref{rec32_a90_t61_num16}. The PSNR and SSIM values for the reconstructed dynamic images across the different methods and noise levels are shown in \cref{SNRnum16}. The results demonstrate that our proposed method consistently performs well across all noise levels, with particularly strong performance when the noise level in the sinogram is high. 

\begin{table}[htbp!]
\caption{PSNR and SSIM of the reconstructed dynamic rat abdomen image across the different methods and noise levels of the sinogram. The bold numbers mark the best performances.}
\centering
\begin{tabular}{cc|cccccc}
\hline
\multicolumn{2}{c|}{}                             & \texttt{EM}     & \texttt{EM-NMF} & \texttt{MAP-TV}     & \texttt{DIP-B}           & \texttt{INR-B}  & \texttt{NINRF}         \\ \hline
\multicolumn{1}{c|}{\multirow{2}{*}{30db}} & PSNR & 22.34  & 26.68  & 37.04  & 39.61           & 40.98  & \textbf{41.47}  \\
\multicolumn{1}{c|}{}                      & SSIM & 0.6679 & 0.8071 & 0.9588 & \textbf{0.9893} & 0.9868 & 0.9879          \\ \hline
\multicolumn{1}{c|}{\multirow{2}{*}{20db}} & PSNR & 21.21  & 26.08  & 33.32  & 36.23           & 36.74  & \textbf{37.33}  \\
\multicolumn{1}{c|}{}                      & SSIM & 0.5602 & 0.7904 & 0.9315 & 0.9597          & 0.9680 & \textbf{0.9715} \\ \hline
\multicolumn{1}{c|}{\multirow{2}{*}{10db}} & PSNR & 20.02  & 24.62  & 28.97  & 31.27           & 31.63  & \textbf{32.17}  \\
\multicolumn{1}{c|}{}                      & SSIM & 0.3908 & 0.7355 & 0.8487 & 0.8849          & 0.8934 & \textbf{0.9145} \\ \hline
\multicolumn{1}{c|}{\multirow{2}{*}{5db}} & PSNR & 19.04  & 23.26  & 26.54  & 27.56          & 27.94  & \textbf{29.15}  \\
\multicolumn{1}{c|}{}                      & SSIM & 0.3297 & 0.6773 & 0.7202 & 0.8431          & 0.8463 & \textbf{0.8835} \\ \hline
\end{tabular}
\label{SNRnum16}
\end{table}

\noindent\textbf{Computational cost of different methods.} 
Computational complexity is important for the application  of proposed method in large dataset; see \cref{CompCost} for the computational costs of different methods.  
Even the deep learning-based methods \texttt{DIP-B}, \texttt{INR-B}, and \texttt{NINRF} exhibit significantly better performance compared to traditional iteration method, the computational costs of three deep learning methods 
have no advantage.
As an iterative method,  there is no advantage of our method over the \texttt{EM}, \texttt{EM-NMF} and \texttt{MAP-TV} methods in terms of the computation time for a large number of iterations is needed in reconstruction. 
Since \texttt{INR-B} models only the matrix $\bm{A}$ using neural networks, it has a smaller model size than \texttt{NINRF}. However, both \texttt{INR-B} and \texttt{DIP-B} require multiple inner iterations using an EM-based algorithm to update $\bm{B}$, which increases their computational burden. In contrast, \texttt{NINRF} parametrizes both $A$ and $B$ and updates them simultaneously, resulting in reduced total computational time. As shown in \cref{CompCost}, \texttt{NINRF} achieves the lowest FLOPs and least runtime among the three deep learning methods.

\begin{table}[htbp!]
\caption{Computational cost of the compared methods. The bold numbers mark the best performances.}
\centering
\begin{tabular}{c|cccccc}
\hline
       & \texttt{EM}    & \texttt{EM-NMF} & \texttt{MAP-TV} & \texttt{DIP-B}  & \texttt{INR-B}  & \texttt{NINRF}  \\ \hline
params & -     & -      & -      & 14.36M & 1.64M  & 3.29M  \\
FLOPs  & $\approx0.34$G & $\approx0.44$G  & $\approx93.50$G  & $\approx55.59$G & $\approx26.15$G & $\approx20.58$G \\
time   & 2.17s & 3.32s  & 61.44s & 1452s  & 426s   & 224s   \\ \hline
\end{tabular}
\label{CompCost}
\end{table}

\noindent\textbf{Reconstruction for different $K$.} 
To investigate the optimal values of $K$, we performed studies by setting $K$ with different values; see \cref{Kvalue} for the quantitative results. It can be found that performance gains little with $K>5$. Besides, increasing the value of $K$ will introduce more parameters and aggravate the computational burden of the network. Thus, we chose $K = 5$ in this task.

% For methods based on the NMF model, we study the reconstruction performance under different values of $K$. As shown in \cref{Kvalue}, both PSNR and SSIM consistently improve as K increases. However, the improvements become small when $K>5$, indicating that further increasing of $K$ brings limited benefits. This suggests that $K=5$ provides a good balance between model complexity and reconstruction quality for this example.

\begin{table}[htbp!]
\caption{PSNR and SSIM of the reconstructed dynamic rat abdomen image for different $K$ in the methods based on the NMF model.}
\centering
\begin{tabular}{cc|cccc}
\hline
\multicolumn{2}{c|}{$K$}                             & 4      & 5      & 6      & 7      \\ \hline
\multicolumn{1}{c|}{\multirow{2}{*}{\texttt{DIP-B}}} & PSNR & 24.98  & 25.64  & 26.29  & 26.31  \\
\multicolumn{1}{c|}{}                       & SSIM & 0.8184 & 0.8447 & 0.8605 & 0.8620 \\ \hline
\multicolumn{1}{c|}{\multirow{2}{*}{\texttt{INR-B}}} & PSNR & 26.71  & 27.50  & 29.63 & 29.76  \\
\multicolumn{1}{c|}{}                       & SSIM & 0.8750 & 0.8876 & 0.9392 & 0.9398 \\ \hline
\multicolumn{1}{c|}{\multirow{2}{*}{\texttt{NINRF}}} & PSNR & 28.14  & 28.56  & 30.56  & 30.60  \\
\multicolumn{1}{c|}{}                       & SSIM & 0.9166 & 0.9284 & 0.9392 & 0.9398 \\ \hline
\end{tabular}
\label{Kvalue}
\end{table}

\begin{figure}[htbp!]
    \centering
    \subfigure[Phantom]{
    \includegraphics[width=0.4\textwidth]{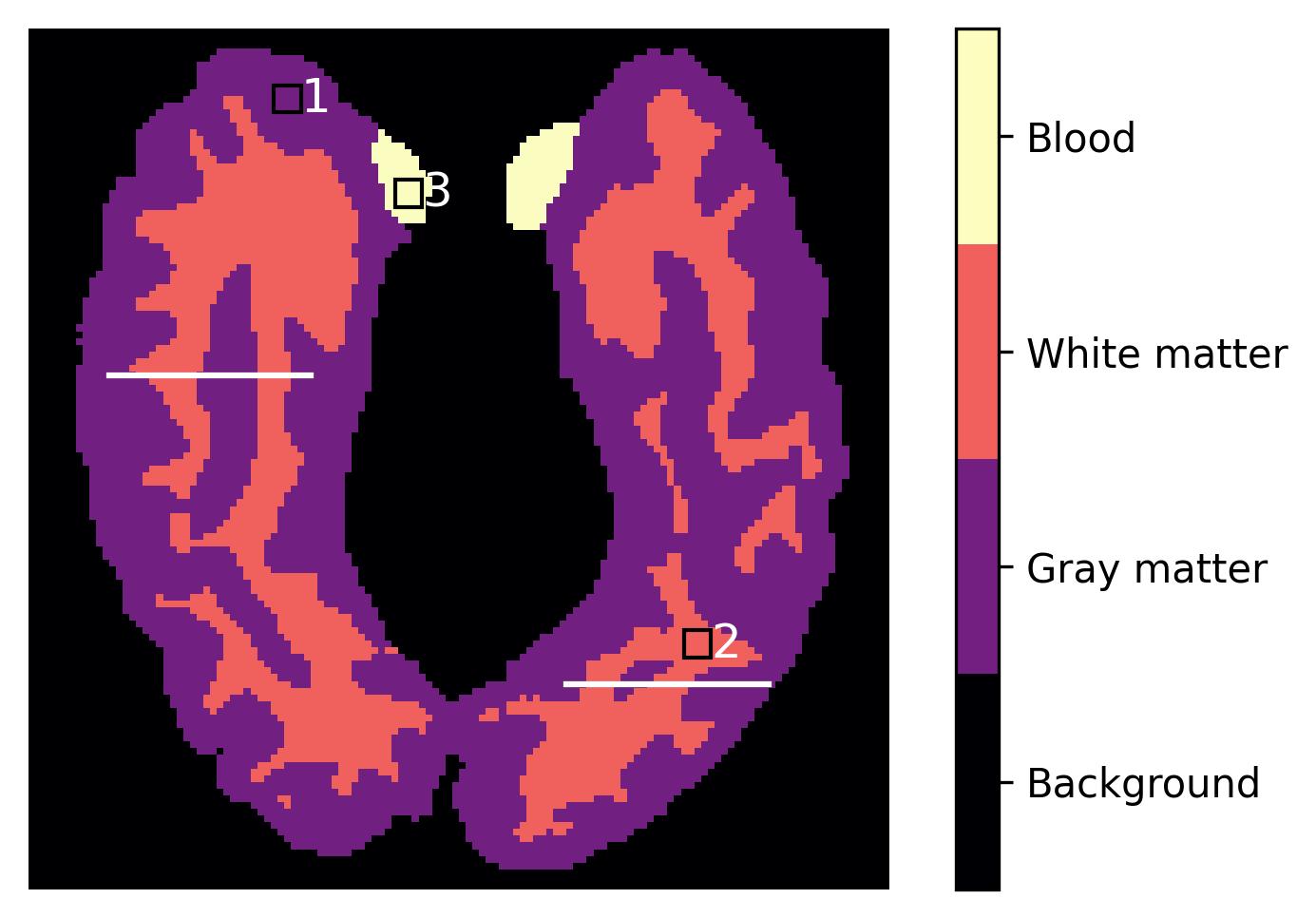}
    \label{ph201}
    }
    % \hspace{1cm}
    \subfigure[TACs]{
    \includegraphics[width=0.4\textwidth]{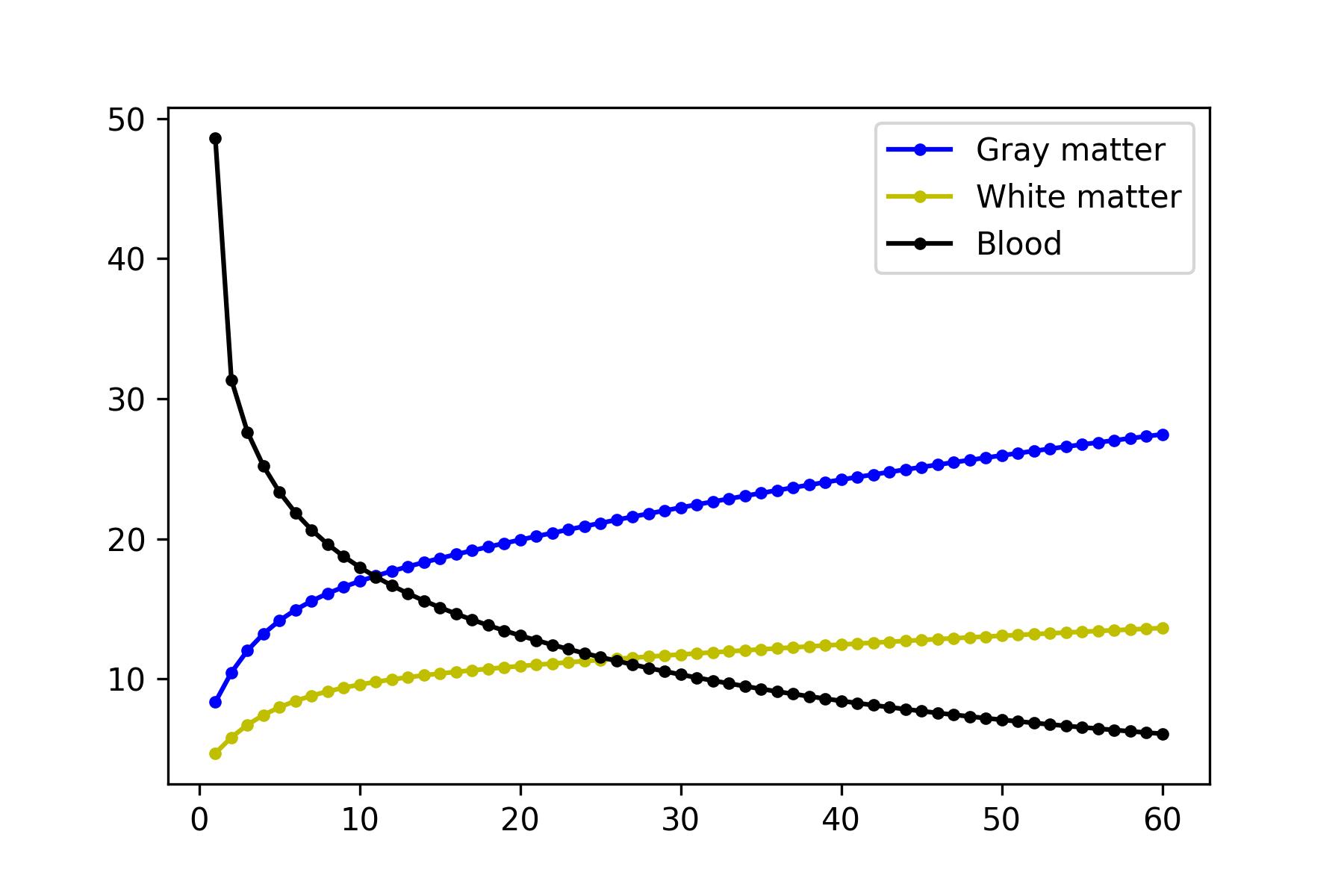}
    \label{tac201}
    }
    \caption{The simulated brain image phantom and the corresponding TACs. In (b), the horizontal axis represents time, while the vertical axis denotes the corresponding value.}
    \label{TACshow201}
\end{figure}

\subsubsection{Simulated Brain Phantom Image Reconstruction} \label{exp_brain1}
We further conduct experiments on the simulated brain data with a complex structure. The Zubal brain phantom~\cite{zubal1994computerized} is used to simulate dynamic PET data. The size of the phantom is $h=128$ and $w=128$, and it contains 3 main regions: blood, white matter, and gray matter as shown in \cref{ph201}. The simulated tracer is \textsuperscript{18}F-FDG and a three compartment model~\cite{phelps1979tomographic} is employed to simulate its variations. As in \cite{feng1993models}, the input function, which is the TAC function of blood region, is 
\begin{equation}\label{Cp}
    C_p(t)=(A_1t-A_2-A_3)e^{-\mu_1 t} + A_2e^{-\mu_2 t} + A_3e^{-\mu_3 t},
\end{equation}
where $A_1=851.1225$, $A_2=21.8798$, $A_3=20.8113$, $\mu_1=-4.1339$, $\mu_2=-0.1191$, $\mu_3=-0.0104$. Model parameters of compartment model are set as follows: [$K_1 (\textrm{min}^{-1})$, $k_2 (min^{-1})$, $k_3 (min^{-1})$, $ k_4 (min^{-1})$, $V_b (unitless)$],
gray matter =[0.102, 0.130, 0.062, 0.007, 0.03],
white matter =[0.054, 0.109, 0.045, 0.006, 0.02].
The dynamic PET image consists of $T=60$ frames over 60 minutes, each with equal time intervals: 60 $\times$ 60s. The corresponding TACs used in the dataset are shown in \cref{tac201}. True images are obtained by filling the TACs into the brain phantom. We use the same way as it in the rat abdomen image to generate the sinogram of the simulated brain image. The acquired sinogram consists of $n_a = 30$ projections and $n_l=195$ bins. The noise level is set to an SNR of 20 dB.

% \noindent\textbf{Model Setting} \label{brain-ms2}
For the methods \texttt{DIP-B}, \texttt{INR-B} and \texttt{NINRF}, which all employ the NMF model, we set the model rank $K=6$. For our proposed method \texttt{NINRF}, the hyperparameters for positional encoding are set to $d_1=d_2=256$ and $\sigma_1=\sigma_2=8$. The initial learning rate $\alpha_1$ and $\alpha_2$ are both set to $5 \times 10^{-3}$. For the methods \texttt{NINRF} and \texttt{INR-B}, the network for $\bm{A}$ consists of 4 hidden layers, each with a width of 512. The regularization parameters in \texttt{MAP-TV} are set to $\lambda_{\text{TV}_1}=0.1$ and $\lambda_{\text{TV}_2}=5$. For the methods \texttt{DIP-B} and \texttt{INR-B}, the regularization parameters for $\bm{A}$ and $\bm{B}$ are set to $\lambda_1=100$ and $\lambda_2=1$. For \texttt{NINRF}, the values are $\lambda_1=2\times 10^4$ and $\lambda_2=5e-4$.

\begin{figure}[htbp!]
    \centering
    \resizebox{\textwidth}{!}{
    \begin{tabular}{c@{\hspace{2pt}}c@{\hspace{1pt}}c@{\hspace{1pt}}c@{\hspace{1pt}}c@{\hspace{1pt}}c@{\hspace{1pt}}c@{\hspace{1pt}}c}
    &\multicolumn{7}{c}{\includegraphics[width=14cm]{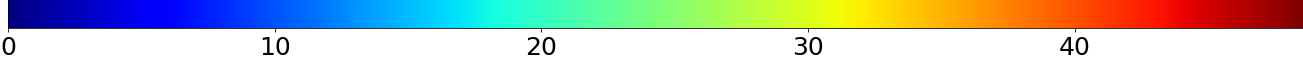}} \\
		\put(-15,10){\rotatebox{90}{Frame 1}}&
        \includegraphics[width=2cm]{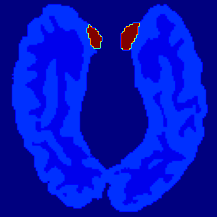}&
		\includegraphics[width=2cm]{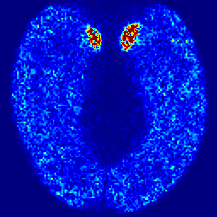}&
		\includegraphics[width=2cm]{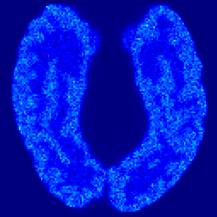}&
		\includegraphics[width=2cm]{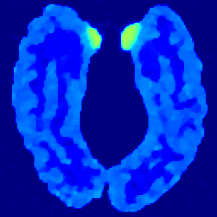}&
		\includegraphics[width=2cm]{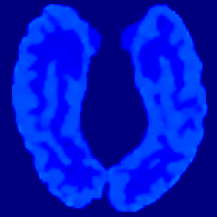}&
		\includegraphics[width=2cm]{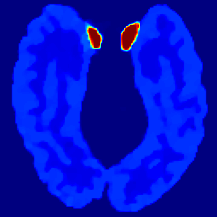}&
		\includegraphics[width=2cm]{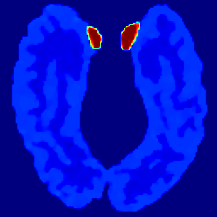}\\
        \put(-15,10){\rotatebox{90}{error map}}& &
        \includegraphics[width=2cm]{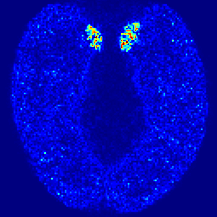}&
	\includegraphics[width=2cm]{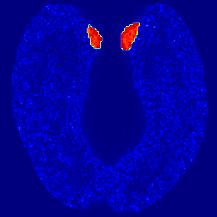}&
	\includegraphics[width=2cm]{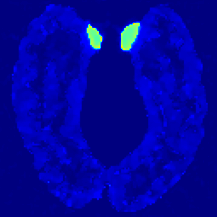}&
	\includegraphics[width=2cm]{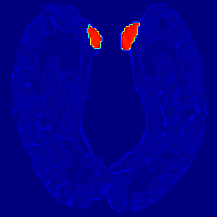}&
	\includegraphics[width=2cm]{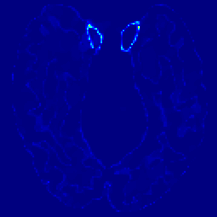}&
        \includegraphics[width=2cm]{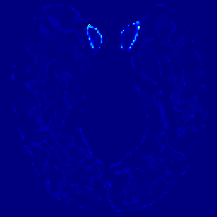}\\
        \put(-15,10){\rotatebox{90}{Frame 31}}&
	\includegraphics[width=2cm]{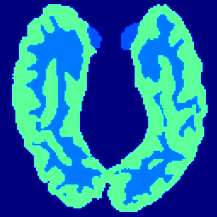}&
		\includegraphics[width=2cm]{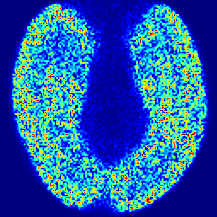}&
		\includegraphics[width=2cm]{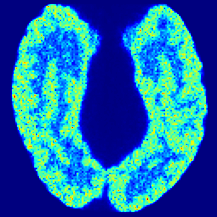}&
		\includegraphics[width=2cm]{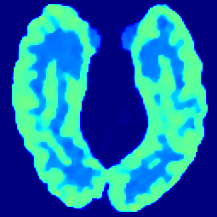}&
		\includegraphics[width=2cm]{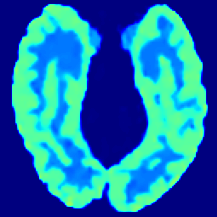}&
		\includegraphics[width=2cm]{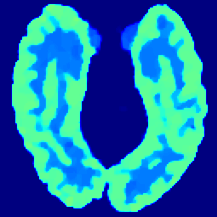}&
		\includegraphics[width=2cm]{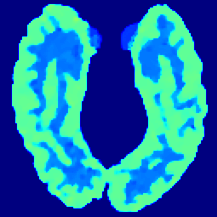}\\
  \put(-15,10){\rotatebox{90}{error map}}& &
        \includegraphics[width=2cm]{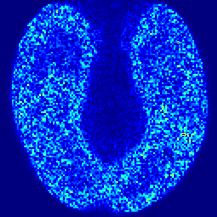}&
	\includegraphics[width=2cm]{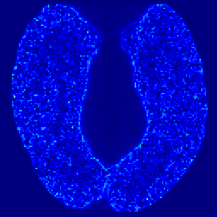}&
	\includegraphics[width=2cm]{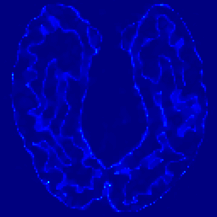}&
	\includegraphics[width=2cm]{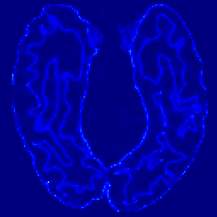}&
	\includegraphics[width=2cm]{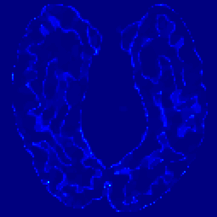}&
        \includegraphics[width=2cm]{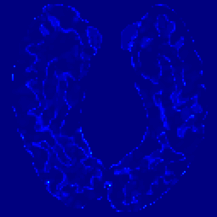}\\
  \put(-15,10){\rotatebox{90}{Frame 51}}&
      \includegraphics[width=2cm]{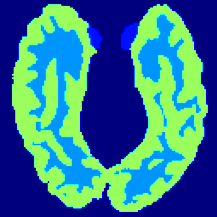}&
		\includegraphics[width=2cm]{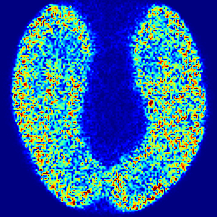}&
		\includegraphics[width=2cm]{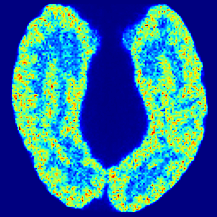}&
		\includegraphics[width=2cm]{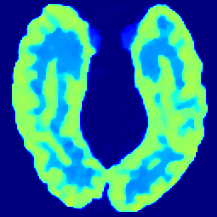}&
		\includegraphics[width=2cm]{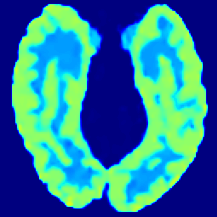}&
		\includegraphics[width=2cm]{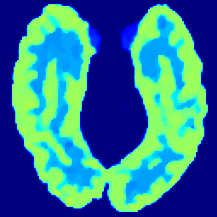}&
		\includegraphics[width=2cm]{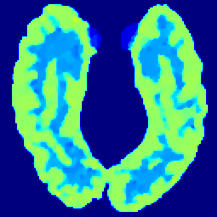}\\
  \put(-15,10){\rotatebox{90}{error map}}& &
        \includegraphics[width=2cm]{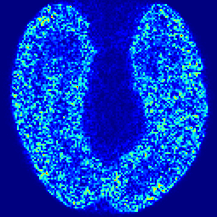}&
	\includegraphics[width=2cm]{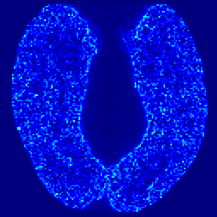}&
	\includegraphics[width=2cm]{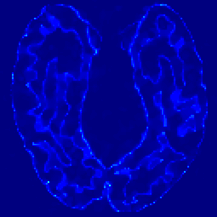}&
	\includegraphics[width=2cm]{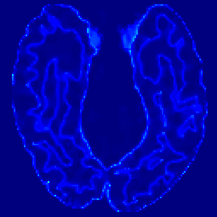}&
	\includegraphics[width=2cm]{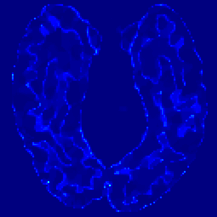}&
        \includegraphics[width=2cm]{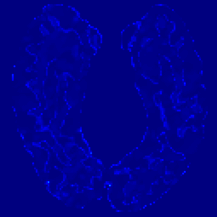}\\
%   PSNR       &   -   & 15.26  & 20.92  & 28.09   & 26.29  & 29.63  & \textbf{30.56}  \\ \hline
% SSIM         &   -  & 0.3179 & 0.6240 & 0.8853 & 0.8605 & 0.9247 & \textbf{0.9392}  \\ \hline
   &Truth &\texttt{EM}& \texttt{EM-NMF}&\texttt{MAP-TV}& \texttt{DIP-B}& \texttt{INR-B}& \texttt{NINRF}
   
    \end{tabular}
    }  
\caption{The reconstructed brain images and corresponding error maps at Frame 1, 31, and 51.}
\label{rec201_show}
\end{figure}

\begin{table}[htbp!]
\caption{PSNR and SSIM of the reconstructed dynamic brain image. The bold numbers mark the best performances.}
\centering
\begin{tabular}{c|cccccc}
\hline
\multicolumn{1}{l|}{} & EM     & EM-NMF & MAP-TV     & DIP-B  & INR-B  & NINRF \\ \hline
PSNR                  & 15.26  & 20.92  & 28.09   & 26.29  & 29.63  & \textbf{30.56}  \\ \hline
SSIM                  & 0.3179 & 0.6240 & 0.8853 & 0.8605 & 0.9247 & \textbf{0.9392}  \\ \hline
\end{tabular}
\label{SNRrec201}
\end{table}

\begin{figure}[htbp!]
    \centering
    \subfigure[PSNR]{
    \includegraphics[width=0.4\textwidth]{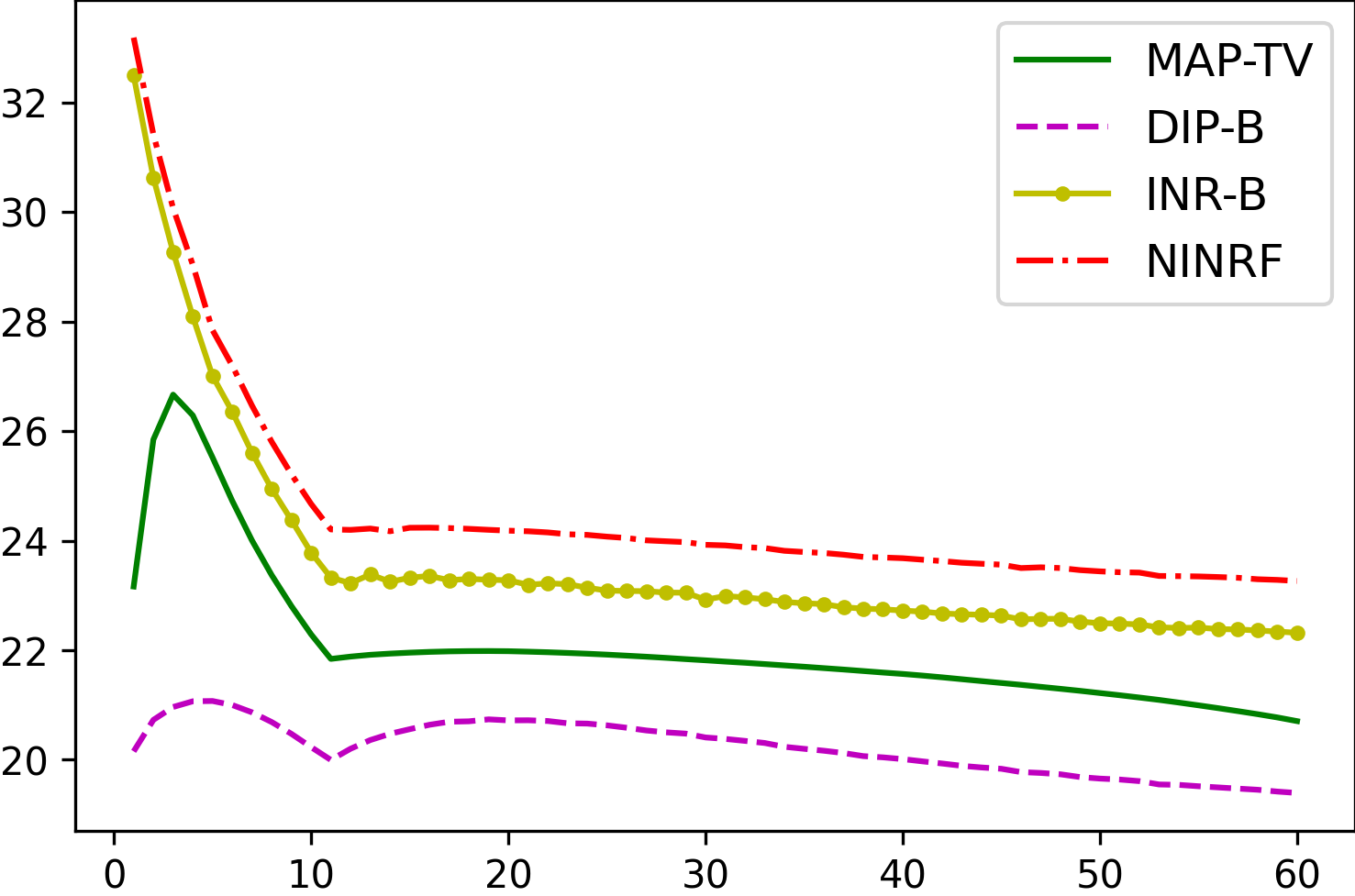}
    }
    \subfigure[SSIM]{
    \includegraphics[width=0.4\textwidth]{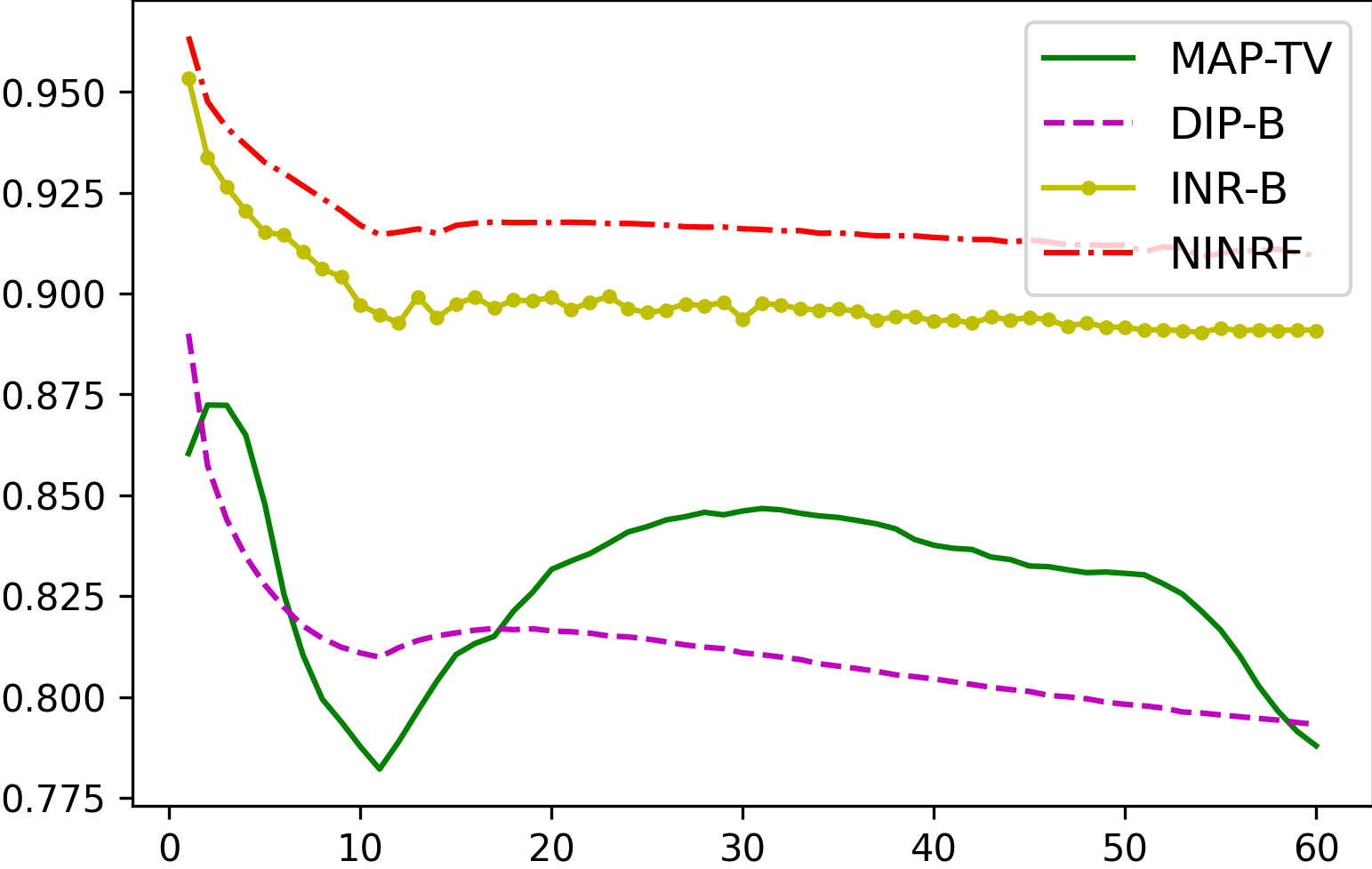}
    }
    \caption{PSNR and SSIM of every frame of the reconstructed brain image. The horizontal axis represents time, while the vertical axis denotes the corresponding values.}
    \label{PSNRSSIM_rec201}
\end{figure}

\begin{figure}[htbp!]
    \centering
    \subfigure[Line1]{
    \label{rec201_profile1}
    \includegraphics[width=0.4\textwidth]{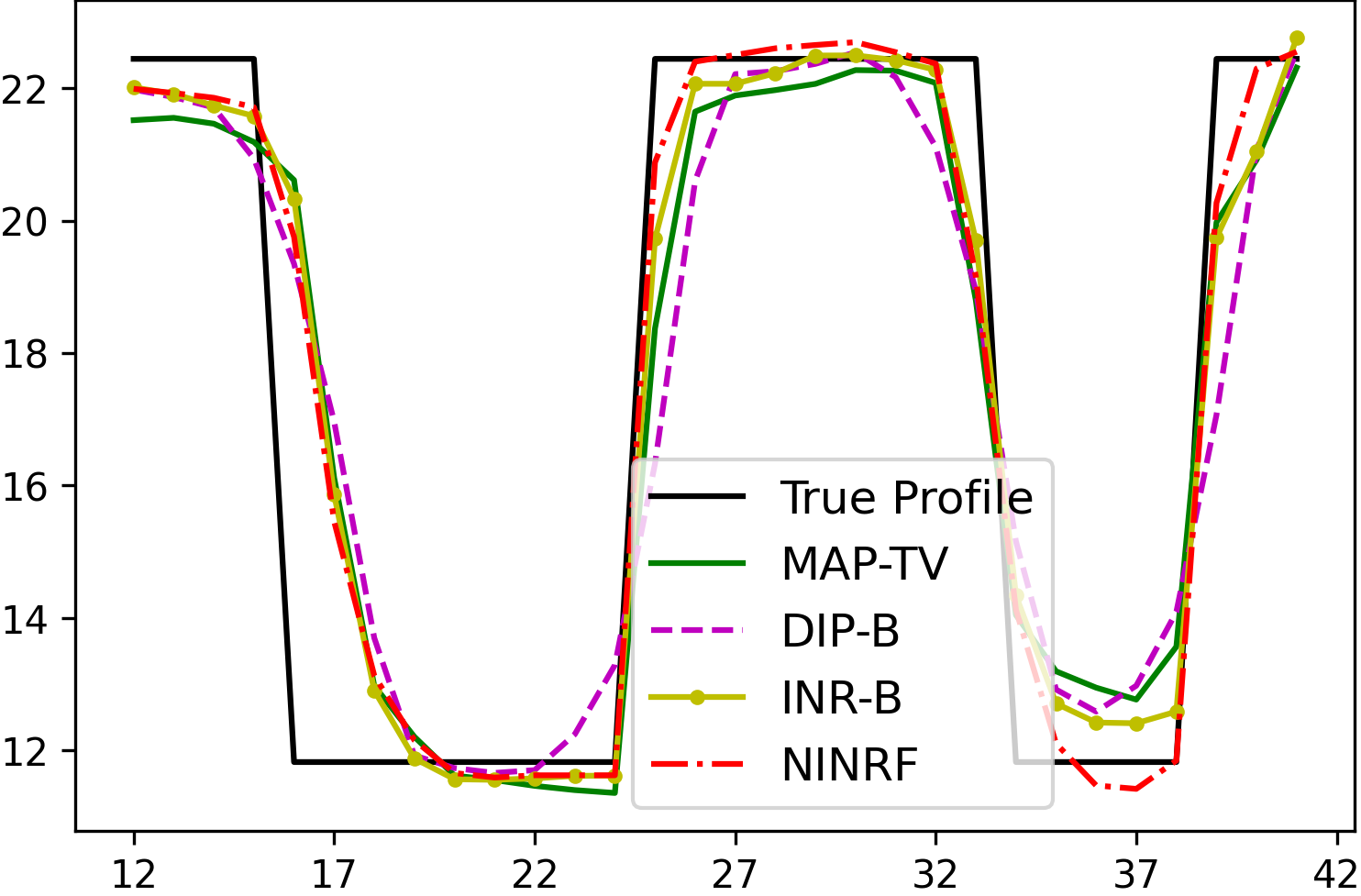}
    }
    \subfigure[Line2]{
    \label{rec201_profile2}
    \includegraphics[width=0.4\textwidth]{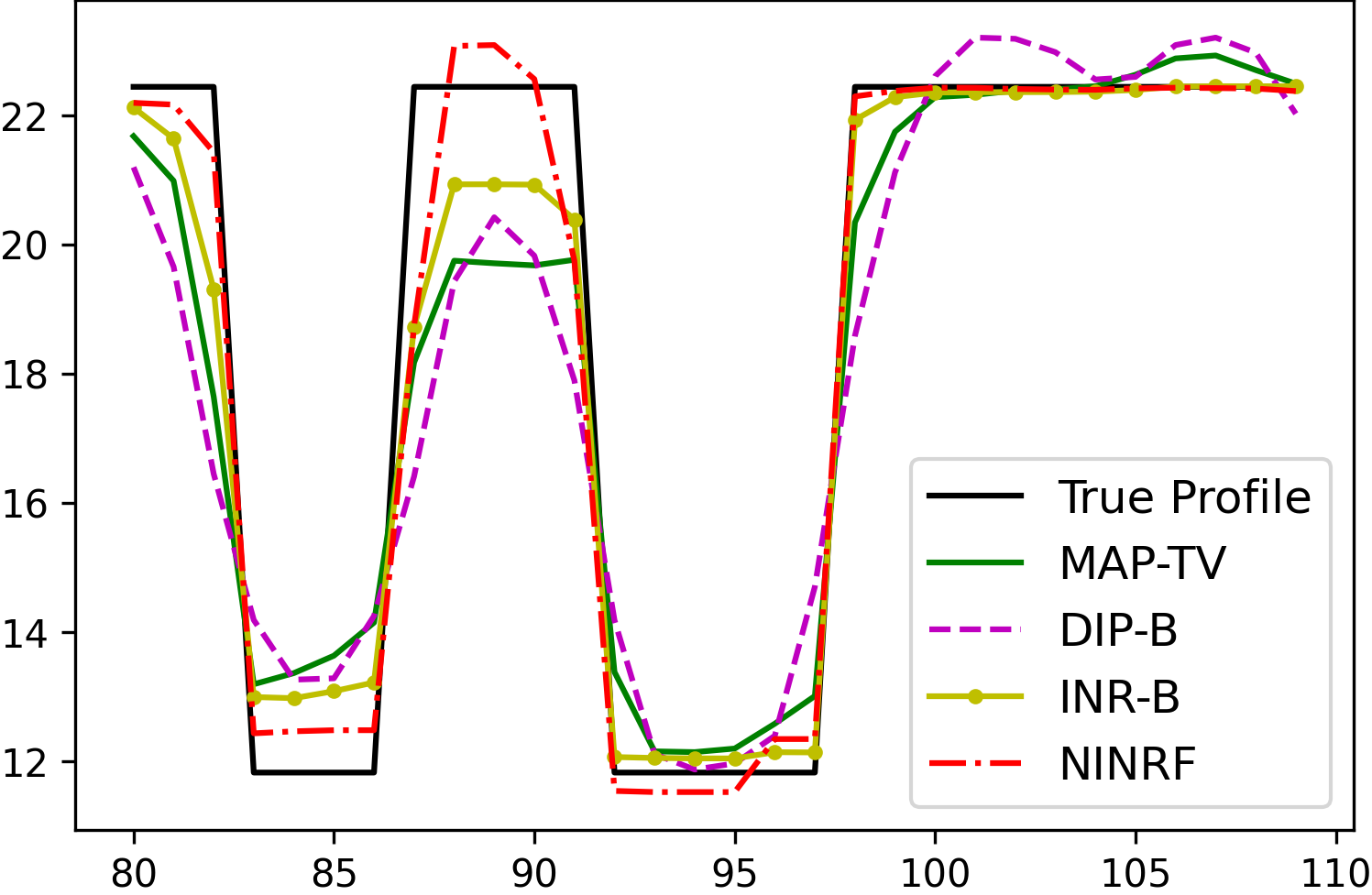}
    }
    \subfigure[Gray matter]{
    \label{TAC201_Ki1}
    \includegraphics[width=0.3\textwidth]{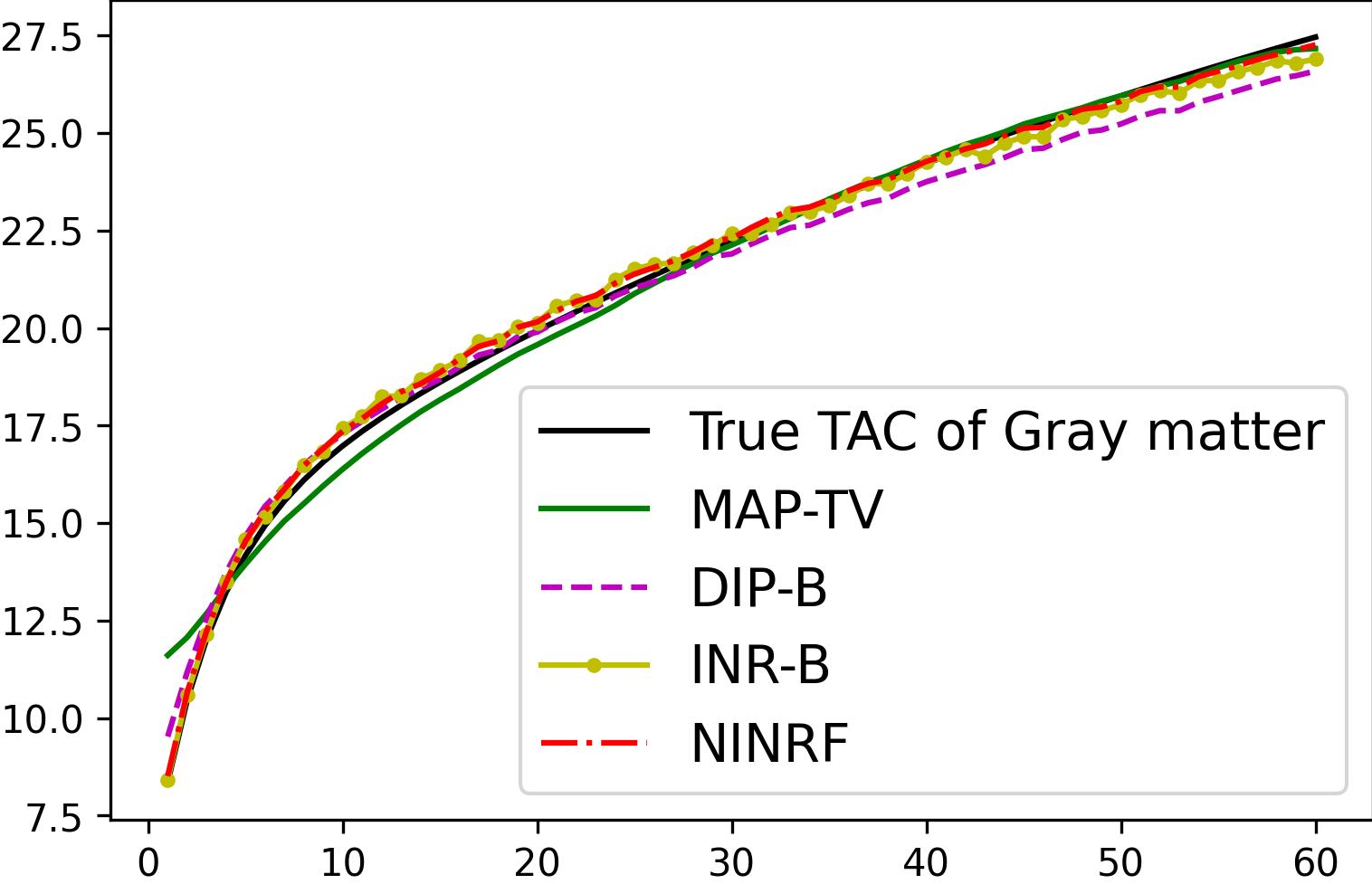}
    }
    \subfigure[White matter]{
    \label{TAC201_Ki2}
    \includegraphics[width=0.3\textwidth]{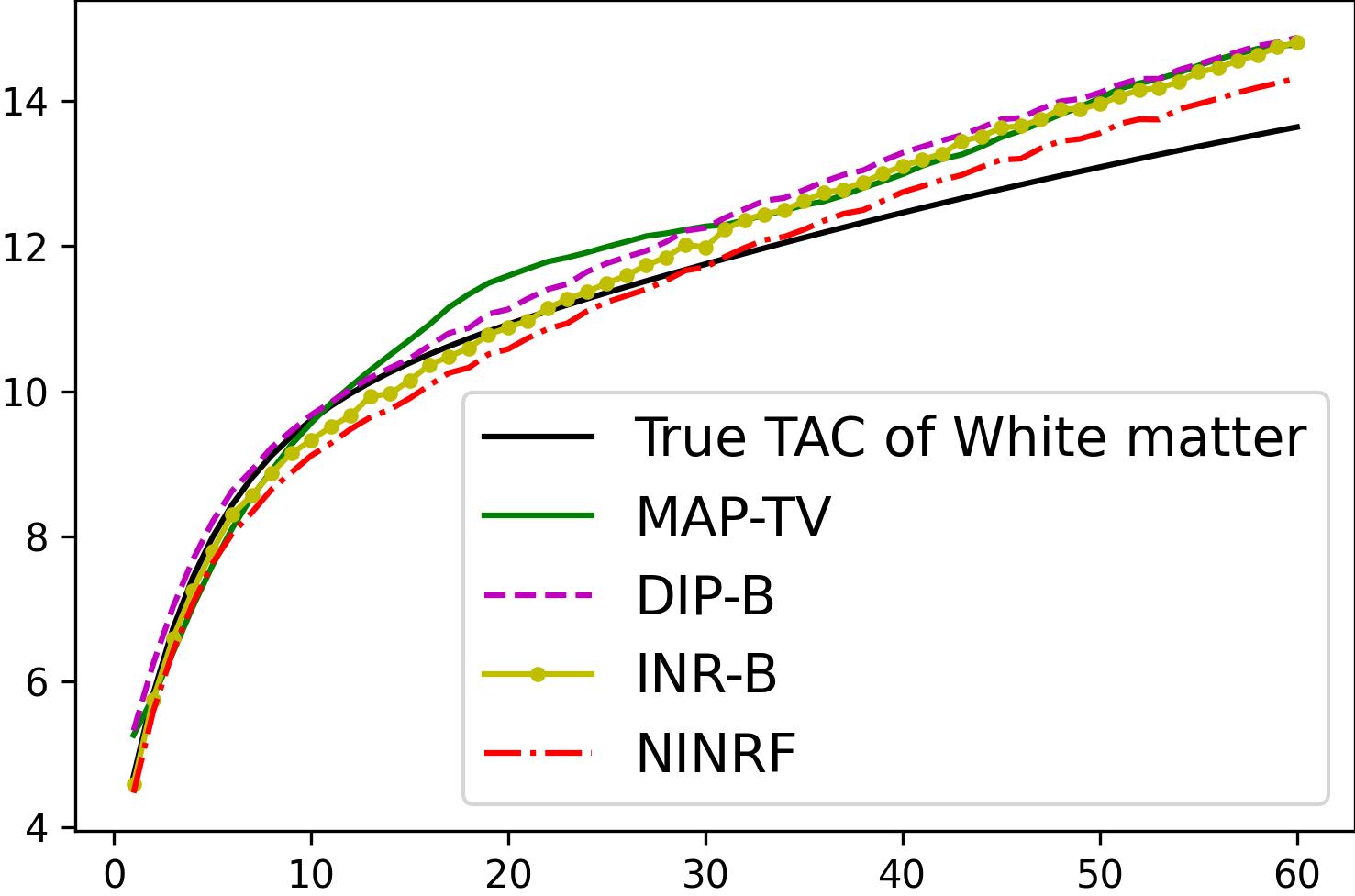}
    }
    \subfigure[Blood]{
    \label{TAC201_Ki3}
    \includegraphics[width=0.3\textwidth]{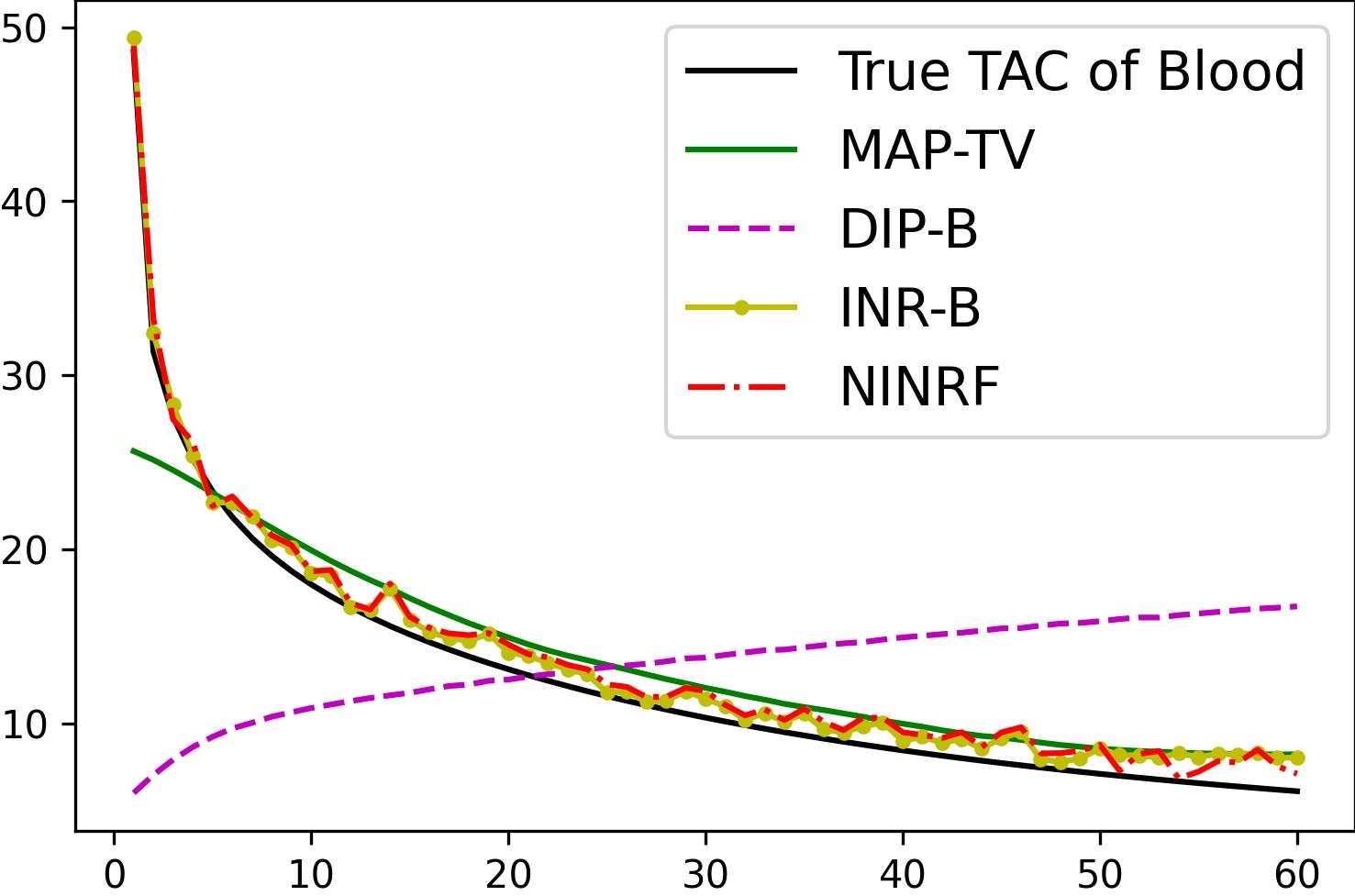}
    }
    \caption{Intensity profiles, and TACs of the reconstructed brain image. (a) and (b):  Intensity profiles of Frame 31 along the white lines in \cref{ph201}, the horizontal axis represents the horizontal position coordinates, while the vertical axis denotes the corresponding intensity values; (c) to (e): TACs at the locations in different ROIs shown in \cref{ph201}, the horizontal axis represents the time, while the vertical axis denotes the corresponding values.}
    \label{PSpT_rec201}
\end{figure}

% \noindent\textbf{Numerical Results} 
We present several frames of the reconstructed brain images and the corresponding error maps at different time points in \cref{rec201_show}. Among all methods, only \texttt{NINRF} and \texttt{INR-B} exhibited good performance in the blood region, particularly in Frame 1. Overall, the reconstructed image quality of \texttt{NINRF} is the best, which is further supported by the PSNR and SSIM values shown in \cref{SNRrec201}. Additionally, we compute the PSNR and SSIM for each frame, as displayed in \cref{PSNRSSIM_rec201}. Our proposed method, \texttt{NINRF}, achieved the highest PSNR and SSIM across all comparison methods. In \cref{rec201_profile1} and \cref{rec201_profile2}, we show the profiles along the white lines in \cref{ph201} for the Frame 31. The profile of \texttt{NINRF} is closest to the ground truth. Finally, we reconstruct the TACs using the mean values from the marked region in \cref{ph201}. The results shown in \cref{TAC201_Ki1} to \cref{TAC201_Ki3} indicate that \texttt{NINRF} achieves the closest approximation to the true TAC values.

\noindent\textbf{Reconstruction for different $K$.} 
Consistent with the previous example in \cref{exp_rat}, we evaluated models with varying rank $K$. The quantitative results in \cref{Kvalue2} indicate that performance improvement is negligible for $K>6$. Accordingly, we set $K = 6$ in this task.

\begin{table}[htbp!]
\caption{PSNR and SSIM of the reconstructed dynamic brain image for different $K$ in the methods based on the NMF model.}
\centering
\begin{tabular}{cc|cccc}
\hline
\multicolumn{2}{c|}{$K$}                             & 4      & 5      & 6      & 7      \\ \hline
\multicolumn{1}{c|}{\multirow{2}{*}{\texttt{DIP-B}}} & PSNR & 24.98  & 25.64  & 26.29  & 26.31  \\
\multicolumn{1}{c|}{}                       & SSIM & 0.8184 & 0.8447 & 0.8605 & 0.8620 \\ \hline
\multicolumn{1}{c|}{\multirow{2}{*}{\texttt{INR-B}}} & PSNR & 26.71  & 27.50  & 29.63 & 29.76  \\
\multicolumn{1}{c|}{}                       & SSIM & 0.8750 & 0.8876 & 0.9392 & 0.9398 \\ \hline
\multicolumn{1}{c|}{\multirow{2}{*}{\texttt{NINRF}}} & PSNR & 28.14  & 28.56  & 30.56  & 30.60  \\
\multicolumn{1}{c|}{}                       & SSIM & 0.9166 & 0.9284 & 0.9392 & 0.9398 \\ \hline
\end{tabular}
\label{Kvalue2}
\end{table}

\begin{figure}[htbp!]
    \centering
    \subfigure[Truth]{
    \label{Ki201pix0}
    \includegraphics[width=0.15\textwidth]{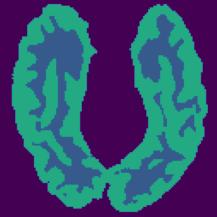}
    }
    \subfigure[\texttt{MAP-TV}]{
    \label{Ki201pix1}
    \includegraphics[width=0.15\textwidth]{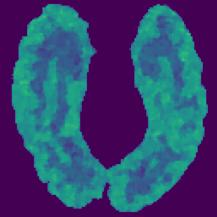}
    }
    \subfigure[\texttt{DIP-B}]{
    \label{Ki201pix2}
    \includegraphics[width=0.15\textwidth]{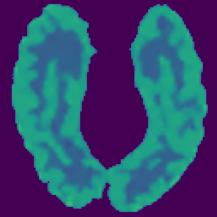}
    }
    \subfigure[\texttt{INR-B}]{
    \label{Ki201pix3}
    \includegraphics[width=0.15\textwidth]{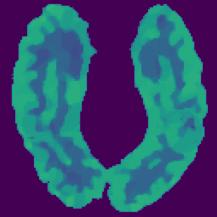}
    }
    \subfigure[\texttt{NINRF}]{
    \label{Ki201pix4}
    \includegraphics[width=0.15\textwidth]{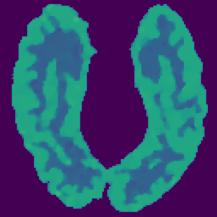}
    }
    \subfigure{
    \label{Ki201pixcolorbar}
    \includegraphics[width=0.045\textwidth]{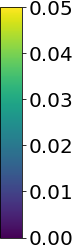}
    }
    \caption{$K_i$ calculated from the tracer at every pixel.}
    \label{Ki201pix}
\end{figure}

\noindent\textbf{Kinetic Parameter Analysis.}
The net influx rate constant~(denoted as $K_i$) is a pivotal pharmacokinetic parameter in dynamic PET imaging, quantifying the net transfer rate of a tracer from plasma to tissue and reflecting underlying metabolic or exchange processes through kinetic modeling. It serves as a biomarker for disease diagnosis and physiological assessment, enabling precise quantification of metabolic activity in oncology, cerebral perfusion in neurology, and myocardial blood flow in cardiology. Using the reconstructed tracer in different regions, we can use Patlak plot method~\cite{patlak1983graphical} to compute it. When the tracer distribution among the compartments reaches equilibrium, the tracer $C_T(t)$ and the input function $C_p(t)$ satisfy the following equation:
\begin{equation*}
    \frac{C_T(t)}{C_p(t)} = K_i \frac{\int_0^t{C_p(\tau)}d\tau}{C_p(t)}+V,
\end{equation*}
where $K_i$ is the slope and $V$ is the intercept. In practice, we apply the least squares method to calculate $K_i$ using all the points in the TAC when $t\ge50$. Based on the results in \cref{TAC201_Ki1} to \cref{TAC201_Ki3}, we compute $K_i$ using images reconstructed with different methods in both the gray matter and white matter regions, as shown in \cref{Ki201}. Our proposed method \texttt{NINRF} performs the best in both regions. Furthermore, $K_i$ can be calculated for every pixel in the image, and we visualize these pixel-wise estimates for each method in \cref{Ki201pix}. The $K_i$ map of \texttt{NINRF} best approximates the truth. For different methods, the mean absolute error (MAE) and mean relative error (MRE) of $K_i$ are displayed in \cref{Ki201} and our proposed method achieves the smaller errors compared to other methods.

\begin{table}[htbp!]
\caption{The reconstructed $K_i$ and corresponding errors calculated from the TACs in \cref{PSpT_rec201} and for all pixels. The bold numbers mark the best performances.}
\centering
\scalebox{0.8}[0.8]{
\begin{tabular}{ccc|ccccc}
\hline
         &                         &              & True    & \texttt{MAP-TV}  & \texttt{DIP-B}   & \texttt{INR-B}   & \texttt{NINRF}            \\ \hline
\multicolumn{1}{c|}{\multirow{6}{*}{\begin{tabular}[c]{@{}c@{}}Based\\ on\\ TACs\end{tabular}}}   & \multicolumn{1}{c|}{\multirow{2}{*}{$K_i$}}                                                         & Gray Matter  & 0.03060 & 0.02944 & 0.02912 & 0.02826 & \textbf{0.02994} \\
\multicolumn{1}{c|}{}                                                                             & \multicolumn{1}{c|}{}                                                                               & White Matter & 0.01419 & 0.01608 & 0.01628 & 0.01674 & \textbf{0.01571} \\ \cline{2-8} 
\multicolumn{1}{c|}{}                                                                             & \multicolumn{1}{c|}{\multirow{2}{*}{Absolute Error}}                                                & Gray Matter  & -       & 0.00116 & 0.00148 & 0.00234 & \textbf{0.00066} \\
\multicolumn{1}{c|}{}                                                                             & \multicolumn{1}{c|}{}                                                                               & White Matter & -       & 0.00189 & 0.00210 & 0.00255 & \textbf{0.00152} \\ \cline{2-8} 
\multicolumn{1}{c|}{}                                                                             & \multicolumn{1}{c|}{\multirow{2}{*}{Relative Error}}                                                & Gray Matter  & -       & 0.03785 & 0.04827 & 0.07639 & \textbf{0.02157} \\
\multicolumn{1}{c|}{}                                                                             & \multicolumn{1}{c|}{}                                                                               & White Matter & -       & 0.13341 & 0.14771 & 0.17960 & \textbf{0.10747} \\ \hline
\multicolumn{1}{c|}{\multirow{4}{*}{Pixel-wise}} & \multicolumn{1}{c|}{\multirow{2}{*}{\begin{tabular}[c]{@{}c@{}}Mean Absolute\\ Error\end{tabular}}} & Gray Matter  & -       & 0.00467 & 0.00306 & 0.00214 & \textbf{0.00169} \\
\multicolumn{1}{c|}{}                                                                             & \multicolumn{1}{c|}{}                                                                               & White Matter & -       & 0.00306 & 0.00353 & 0.00310 & \textbf{0.00269} \\ \cline{2-8} 
\multicolumn{1}{c|}{}                                                                             & \multicolumn{1}{c|}{\multirow{2}{*}{\begin{tabular}[c]{@{}c@{}}Mean Relative\\ Error\end{tabular}}} & Gray Matter  & -       & 0.15251 & 0.09992 & 0.07000 & \textbf{0.05532} \\
\multicolumn{1}{c|}{}                                                                             & \multicolumn{1}{c|}{}                                                                               & White Matter & -       & 0.21556 & 0.24862 & 0.21818 & \textbf{0.18952} \\ \hline
\end{tabular}
}
\label{Ki201}
\end{table}

\begin{figure}[htbp!]
    \centering
    \subfigure[Phantom]{
    \includegraphics[width=0.4\textwidth]{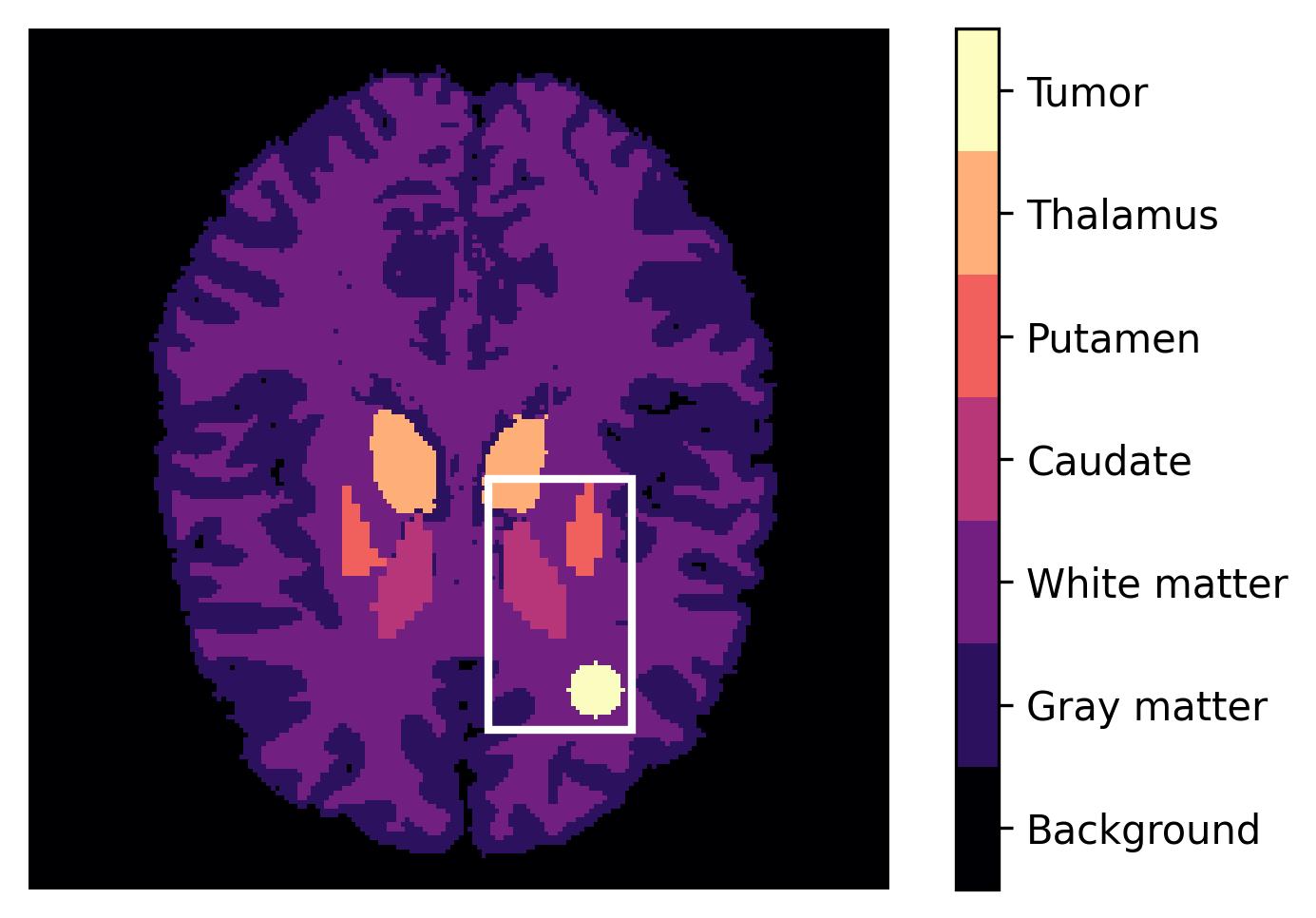}
    \label{ph2}
    }
    % \hspace{1cm}
    \subfigure[TACs]{
    \includegraphics[width=0.4\textwidth]{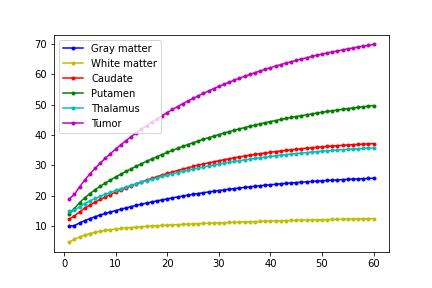}
    \label{tac2}
    }
    \caption{The simulated  brain image phantom  with tumor and the corresponding TACs. In (b), the horizontal axis represents time, while the vertical axis denotes the corresponding value.}
    \label{TACshow2}
\end{figure}

\begin{table}[htbp!]
\caption{Kinetic parameters of different regions in simulated complicated brain images.}
\centering
\begin{tabular}{c|ccccc}
\hline
             & $K_1$ & $k_2$ & $k_3$ & $k_4$ & $V_b$    \\ \hline
Gray matter  & 0.080 & 0.140 & 0.150 & 0.013 & 0.103 \\
White matter & 0.050 & 0.110 & 0.050 & 0.006 & 0.026 \\
Caudate      & 0.120 & 0.170 & 0.190 & 0.016 & 0.101 \\
Putamen      & 0.150 & 0.160 & 0.170 & 0.010 & 0.092 \\
Thalamus     & 0.130 & 0.160 & 0.140 & 0.012 & 0.152 \\
Tumor        & 0.180 & 0.100 & 0.200 & 0.015 & 0.173 \\ \hline
\end{tabular}
\label{kinetic parameters}
\end{table}

\subsubsection{Simulated Brain Tumor Phantom Reconstruction}
The experiment is conducted on a more complicated example of brain phantom with more regions and a tumor is added. The region masks for the phantom are derived from AAL3 dataset~\cite{rolls2020automated}. The size of this phantom is $h=192$ and $w=192$, and it contains 6 main regions: tumor, thalamus, putamen, caudate, white matter, and gray matter, as shown in \cref{ph2}. The same three compartment model explained in \cref{exp_brain1} is employed to generate the TACs. The kinetic parameters for each region are given in \cref{kinetic parameters}. The resulting dynamic PET image consists of $T=60$ frames over 60 minutes, each with equal time intervals: 60 $\times$ 60s. The corresponding TACs are shown in \cref{tac2}, and true images are obtained by filling the TACs into the brain phantom. Similar to the previous experiments, the sinogram of the simulated brain image is generated using Radon transform. It consists of $n_a=60$ projections and $n_l=275$ bins. The sinogram noise level is set to an SNR of 20 dB.

For the methods \texttt{DIP-B}, \texttt{INR-B} and \texttt{NINRF}, which all employ the NMF model, we set the model rank $K=6$. For our proposed method \texttt{NINRF}, the hyperparameters for positional encoding, the initial learning rates, and the network structures are same as that in \cref{exp_brain1}. The regularization parameters in \texttt{MAP-TV} are set to $\lambda_{\text{TV}_1}=0.5$ and $\lambda_{\text{TV}_2}=5$. For the methods \texttt{DIP-B} and \texttt{INR-B}, the regularization parameters for $\bm{A}$ and $\bm{B}$ are set to $\lambda_1=100$ and $\lambda_2=10$. For \texttt{NINRF}, the corresponding values are $\lambda_1=400$ and $\lambda_2=5$.

\begin{figure}[htbp!]
    \centering
    \resizebox{0.9\textwidth}{!}{
    \begin{tabular}{c@{\hspace{2pt}}c@{\hspace{1pt}}c@{\hspace{1pt}}c@{\hspace{1pt}}c@{\hspace{1pt}}c@{\hspace{1pt}}c@{\hspace{1pt}}c}
    &\multicolumn{7}{c}{\includegraphics[width=14cm]{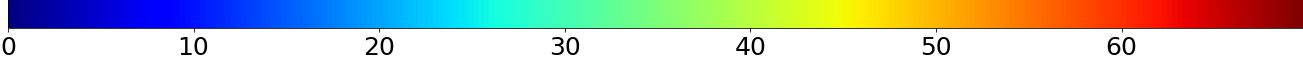}} \\
		\put(-15,10){\rotatebox{90}{Frame 11}}&
        \includegraphics[width=2cm]{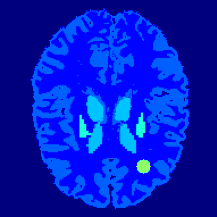}&
       \includegraphics[width=2cm]{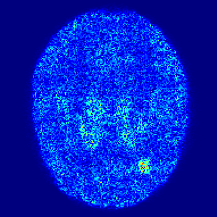}&
       \includegraphics[width=2cm]{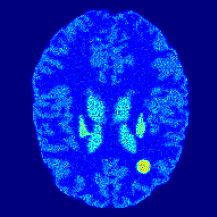}&
       \includegraphics[width=2cm]{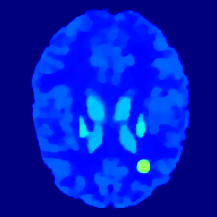}&
       \includegraphics[width=2cm]{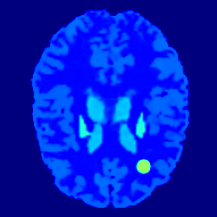}&
      \includegraphics[width=2cm]{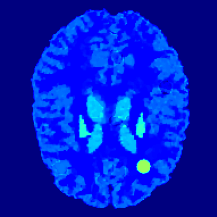}&
     \includegraphics[width=2cm]{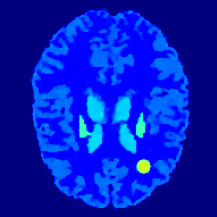}\\
     \put(-15,10){\rotatebox{90}{error map}}& &
        \includegraphics[width=2cm]{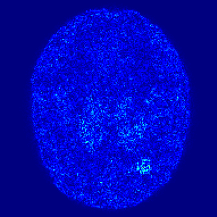}&
	\includegraphics[width=2cm]{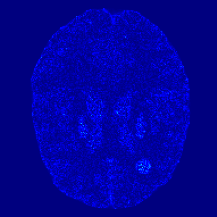}&
	\includegraphics[width=2cm]{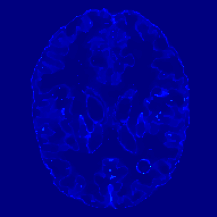}&
	\includegraphics[width=2cm]{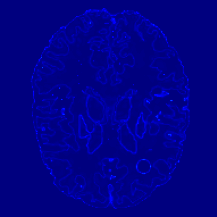}&
	\includegraphics[width=2cm]{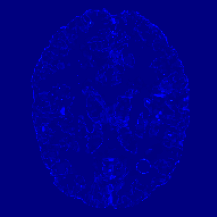}&
        \includegraphics[width=2cm]{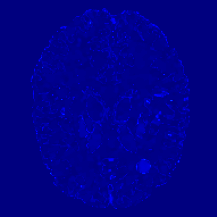}\\
        \put(-15,10){\rotatebox{90}{Frame 31}}&
	\includegraphics[width=2cm]{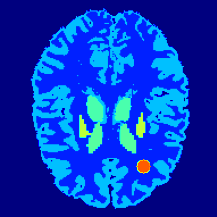}&
       \includegraphics[width=2cm]{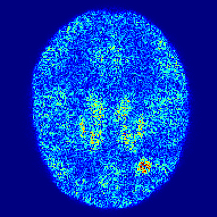}&
       \includegraphics[width=2cm]{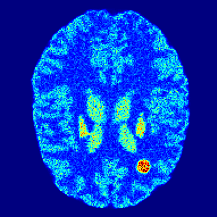}&
      \includegraphics[width=2cm]{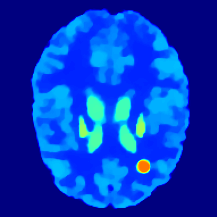}&
     \includegraphics[width=2cm]{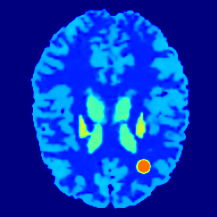}&
     \includegraphics[width=2cm]{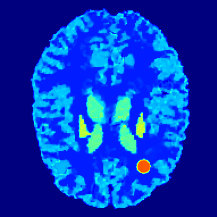}&
    \includegraphics[width=2cm]{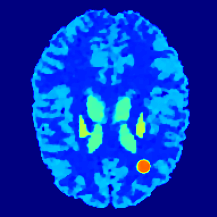}\\
    \put(-15,10){\rotatebox{90}{error map}}& &
        \includegraphics[width=2cm]{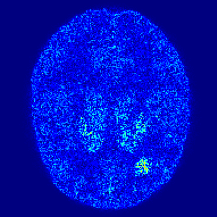}&
	\includegraphics[width=2cm]{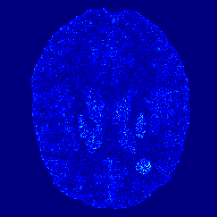}&
	\includegraphics[width=2cm]{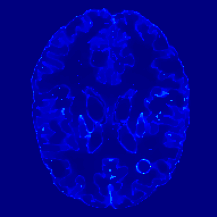}&
	\includegraphics[width=2cm]{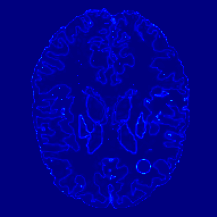}&
	\includegraphics[width=2cm]{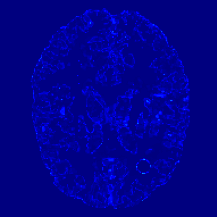}&
        \includegraphics[width=2cm]{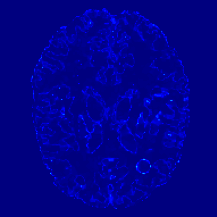}\\
  \put(-15,10){\rotatebox{90}{Frame 51}}&
      \includegraphics[width=2cm]{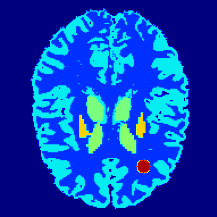}&
     \includegraphics[width=2cm]{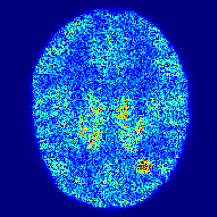}&
     \includegraphics[width=2cm]{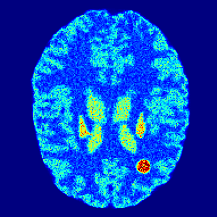}&
    \includegraphics[width=2cm]{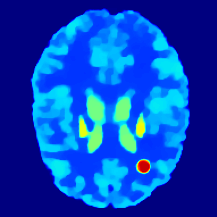}&
    \includegraphics[width=2cm]{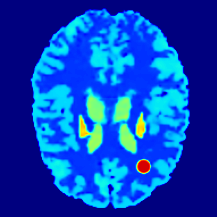}&
   \includegraphics[width=2cm]{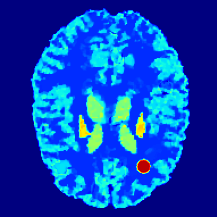}&
   \includegraphics[width=2cm]{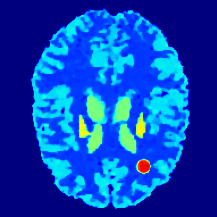}\\
   \put(-15,10){\rotatebox{90}{error map}}& &
        \includegraphics[width=2cm]{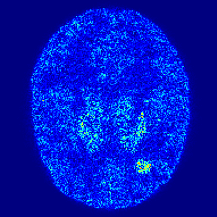}&
	\includegraphics[width=2cm]{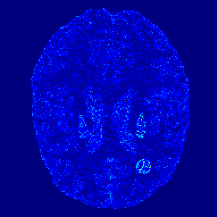}&
	\includegraphics[width=2cm]{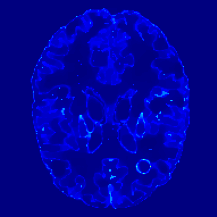}&
	\includegraphics[width=2cm]{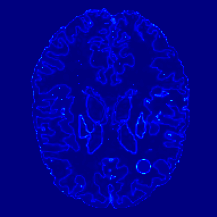}&
	\includegraphics[width=2cm]{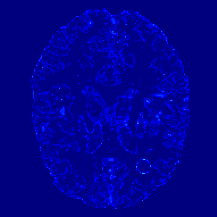}&
        \includegraphics[width=2cm]{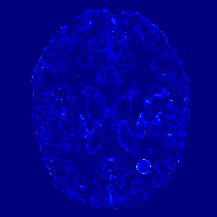}\\
%    PSNR         &   -   & 19.1  & 25.05  & 28.29   & 30.13  & 30.99  & \textbf{31.30}  \\ \hline
% SSIM            &  -  & 0.4725 & 0.7282 & 0.8370 & 0.8981 & 0.9150 & \textbf{0.9239}  \\ \hline
   &Truth &\texttt{EM}& \texttt{EM-NMF}&\texttt{MAP-TV}& \texttt{DIP-B}& \texttt{INR-B}& \texttt{NINRF}
    \end{tabular}
    }  
\caption{The reconstructed brains image with more regions and corresponding error maps at Frame 11, 31, and 51.}
\label{rec32_a60_snr20_num60}
\end{figure}

\begin{figure}[htbp!]
    \centering
    \subfigure[PSNR]{
    \includegraphics[width=0.3\textwidth]{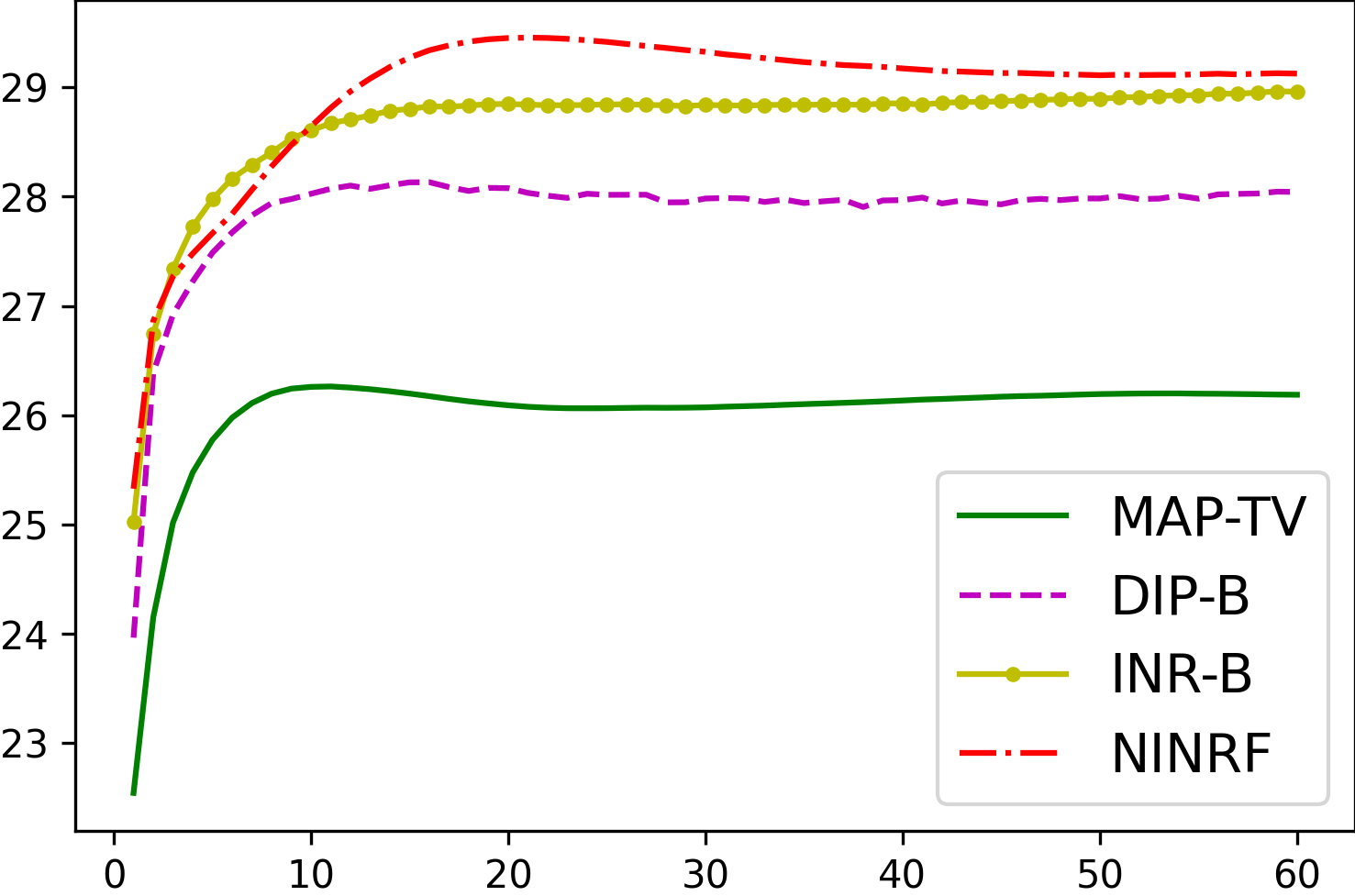}
    }
    \subfigure[SSIM]{
    \includegraphics[width=0.3\textwidth]{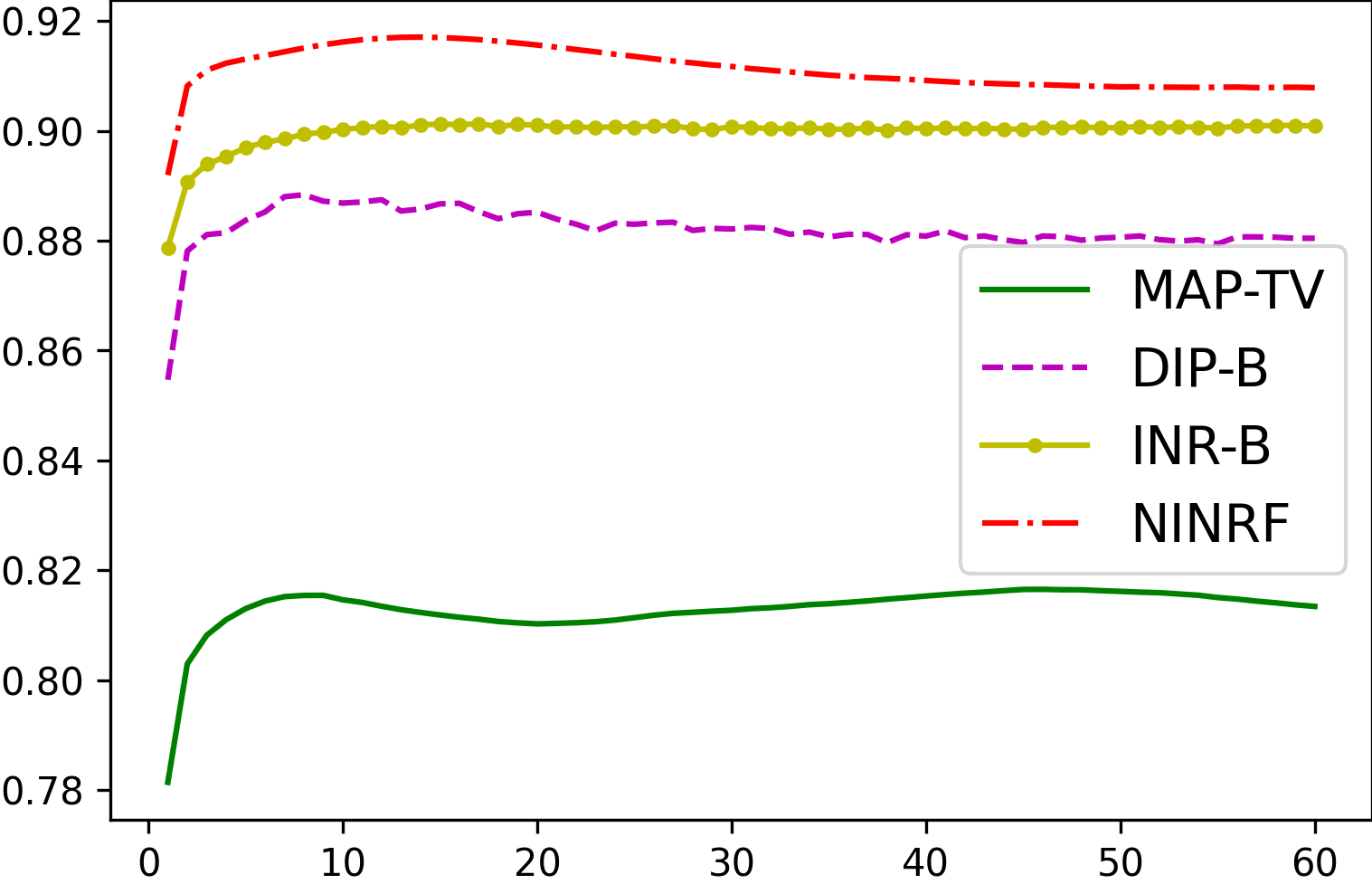}
    }
    \caption{PSNR and SSIM of every frame of the reconstructed complicated brain image. The horizontal axis represents time, while the vertical axis denotes the corresponding values.}
    \label{PSNRSSIM_a60_snr20_num60}
\end{figure}

\begin{figure}[htbp!]
    \centering
    \begin{tabular}{c@{\hspace{2pt}}c@{\hspace{2pt}}c@{\hspace{2pt}}c}
    \includegraphics[width=0.15\textwidth]{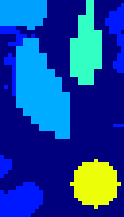}
    & \includegraphics[width=0.15\textwidth]{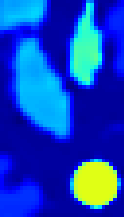}
    & \includegraphics[width=0.15\textwidth]{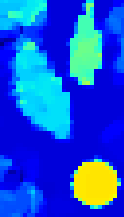}
    & \includegraphics[width=0.15\textwidth]{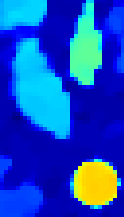} \\
    PSNR & 23.65 & 25.53 & \textbf{25.73} \\
    SSIM & 0.8541 & 0.8868 & \textbf{0.9105}\\
    Truth & \texttt{DIP-B} & \texttt{INR-B} & \texttt{NINRF} 
    \end{tabular}
    \caption{The comparison of the details at the marked region in the white box in \cref{ph2} using different methods.}
    
    \label{zoom_a60_snr20_num60}
\end{figure}

In \cref{rec32_a60_snr20_num60}, we present the reconstructed brain image along with the corresponding error maps. The results from \texttt{NINRF} and \texttt{INR-B} outperform those of the other methods. As shown in the \cref{SNRnum60}, which lists the PSNR and SSIM values for the entire reconstructed image, \texttt{NINRF} provides the best performance. We also compute the PSNR and SSIM for each frame and display the results in \cref{PSNRSSIM_a60_snr20_num60}. Our proposed \texttt{NINRF} outperforms the other methods in terms of PSNR, except for the first few frames, and achieves the highest SSIM across all frames. %\cref{zoom_a60_snr20_num60} provides a comparison of the reconstructed images for the marked region in \cref{ph2} using \texttt{DIP-B}, \texttt{INR-B} and \texttt{NINRF}. 
\textcolor{blue}{The reconstructed images for the marked region in \cref{ph2} using \texttt{DIP-B}, \texttt{INR-B} and \texttt{NINRF} are given in \cref{zoom_a60_snr20_num60}, where the PSNR and SSIM of ROI are also given. The ROI contains the lesion region and is indeed overestimated by the proposed method and underestimated by \texttt{DIP-B}. However,  \texttt{NINRF} achieves significantly higher SSIM in this region, which yields more accurate visual reconstruction.
}
% It shows that \texttt{NINRF} restores the details of this region with higher accuracy.

\begin{table}[htbp!]
\caption{PSNR and SSIM of the reconstructed complicated brain image. The bold numbers mark the best performances.}
\centering
\begin{tabular}{c|cccccc}
\hline
\multicolumn{1}{l|}{} & EM     & EM-NMF & MAP-TV     & DIP-B  & INR-B  & NINRF \\ \hline
PSNR                  & 19.1  & 25.05  & 28.29   & 30.13  & 30.99  & \textbf{31.30}  \\ \hline
SSIM                  & 0.4725 & 0.7282 & 0.8370 & 0.8981 & 0.9150 & \textbf{0.9239}  \\ \hline
\end{tabular}
\label{SNRnum60}
\end{table}

\subsection{Dynamic PET Reconstruction on Clinical Data}

\begin{figure}[htbp!]
    \centering
    \resizebox{0.9\textwidth}{!}{
    \begin{tabular}{c@{\hspace{2pt}}c@{\hspace{1pt}}c@{\hspace{1pt}}c@{\hspace{1pt}}c@{\hspace{1pt}}c@{\hspace{1pt}}c@{\hspace{1pt}}c}
		\put(-15,10){\rotatebox{90}{Frame 6}}&
        \includegraphics[width=2cm]{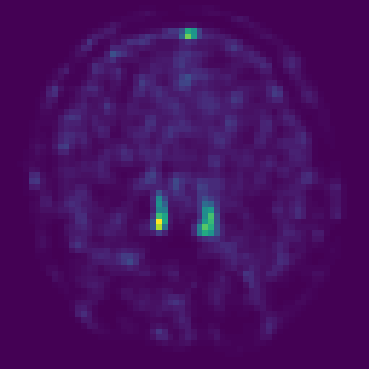}&
		\includegraphics[width=2cm]{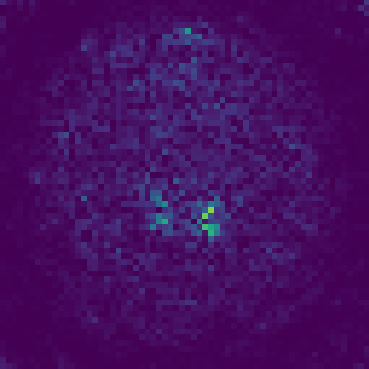}&
		\includegraphics[width=2cm]{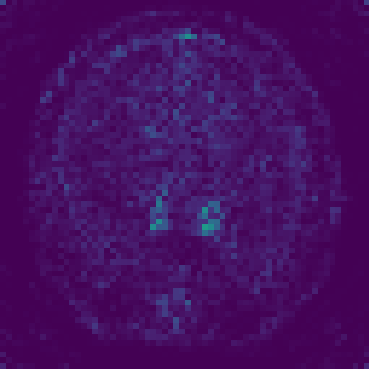}&
		\includegraphics[width=2cm]{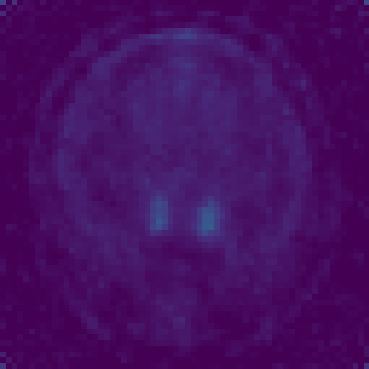}&
		\includegraphics[width=2cm]{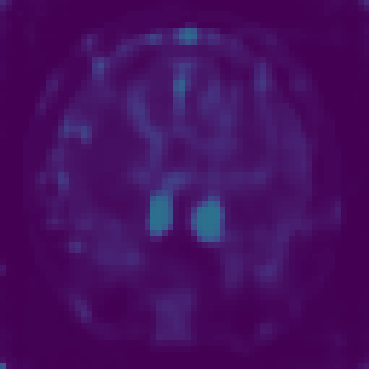}&
		\includegraphics[width=2cm]{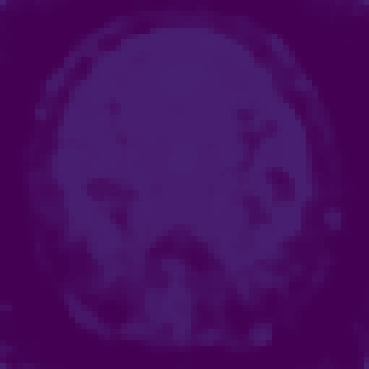}&
		\includegraphics[width=2cm]{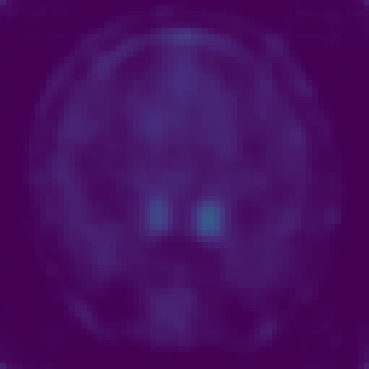}\\
        \put(-15,10){\rotatebox{90}{error map}}& &
        \includegraphics[width=2cm]{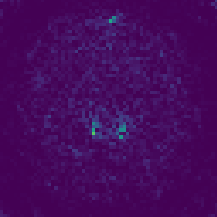}&
	\includegraphics[width=2cm]{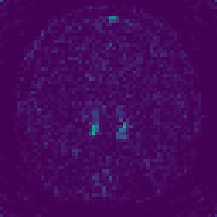}&
	\includegraphics[width=2cm]{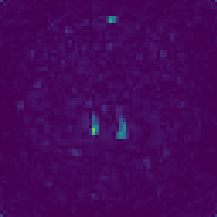}&
	\includegraphics[width=2cm]{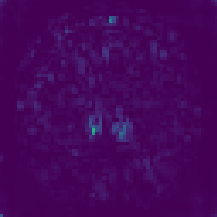}&
	\includegraphics[width=2cm]{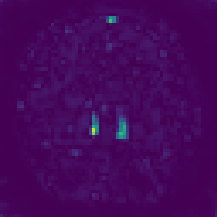}&
        \includegraphics[width=2cm]{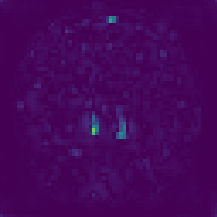}\\
        \put(-15,10){\rotatebox{90}{Frame 16}}&
	\includegraphics[width=2cm]{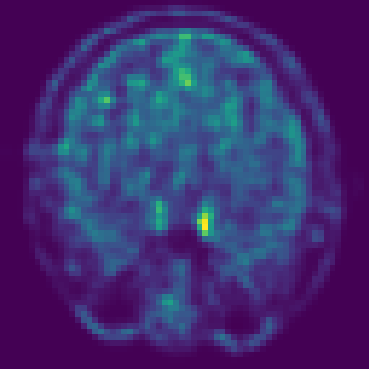}&
		\includegraphics[width=2cm]{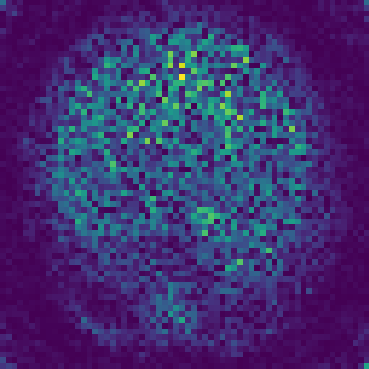}&
		\includegraphics[width=2cm]{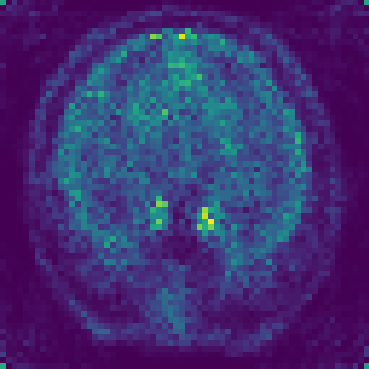}&
		\includegraphics[width=2cm]{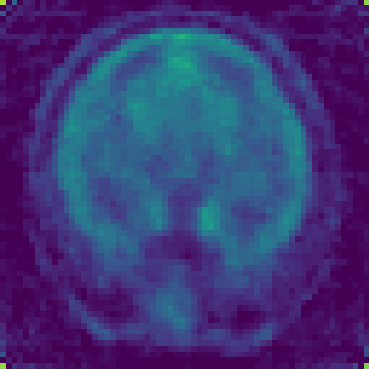}&
		\includegraphics[width=2cm]{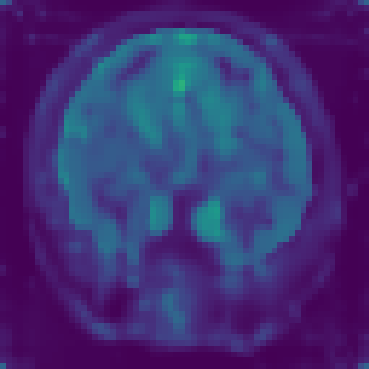}&
		\includegraphics[width=2cm]{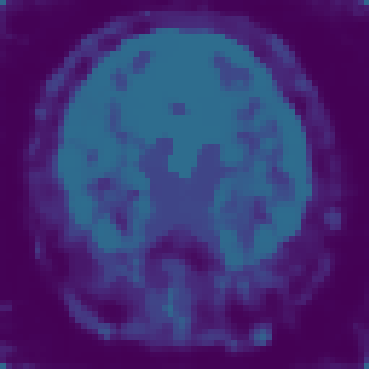}&
		\includegraphics[width=2cm]{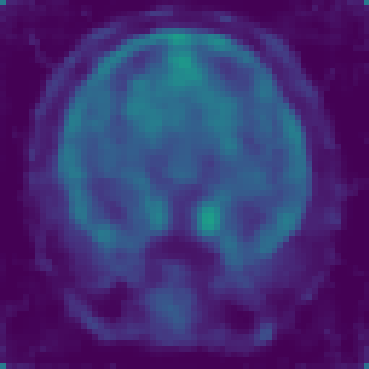}\\
  \put(-15,10){\rotatebox{90}{error map}}& &
        \includegraphics[width=2cm]{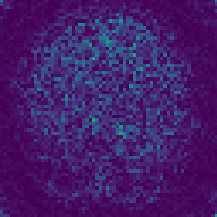}&
	\includegraphics[width=2cm]{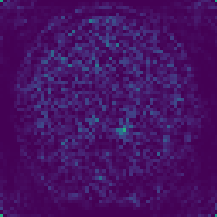}&
	\includegraphics[width=2cm]{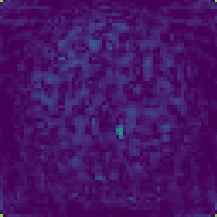}&
	\includegraphics[width=2cm]{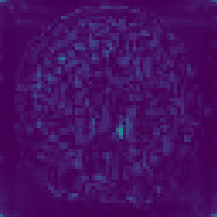}&
	\includegraphics[width=2cm]{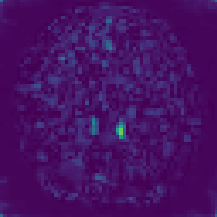}&
        \includegraphics[width=2cm]{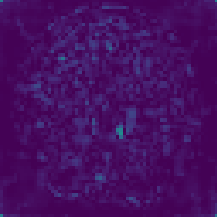}\\
 %  \put(-15,10){\rotatebox{90}{Frame 26}}&
 %      \includegraphics[width=2cm]{img/rec32/num16/utruth_t=26.png}&
	% 	\includegraphics[width=2cm]{img/rec32/num16/uEMU_t=26.png}&
	% 	\includegraphics[width=2cm]{img/rec32/num16/uEMNMF_t=26.png}&
	% 	\includegraphics[width=2cm]{img/rec32/num16/uADMMU_t=26.png}&
	% 	\includegraphics[width=2cm]{img/rec32/num16/uDIPEM_t=26.png}&
	% 	\includegraphics[width=2cm]{img/rec32/num16/uINREM_t=26.png}&
	% 	\includegraphics[width=2cm]{img/rec32/num16/uINRINR_t=26.png}\\
 %  \put(-15,10){\rotatebox{90}{error map}}& &
 %        \includegraphics[width=2cm]{img/rec32/showres2/reuEMU_t=26.png}&
	% \includegraphics[width=2cm]{img/rec32/showres2/reuEMNMF_t=26.png}&
	% \includegraphics[width=2cm]{img/rec32/showres2/reuADMMU_t=26.png}&
	% \includegraphics[width=2cm]{img/rec32/showres2/reuDIPEM_t=26.png}&
	% \includegraphics[width=2cm]{img/rec32/showres2/reuINREM_t=26.png}&
 %        \includegraphics[width=2cm]{img/rec32/showres2/reuINRINR_t=26.png}\\
    \put(-15,10){\rotatebox{90}{Frame 36}}&
      \includegraphics[width=2cm]{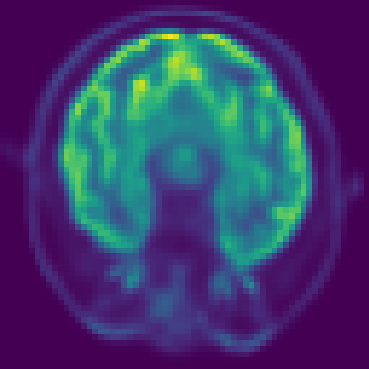}&
		\includegraphics[width=2cm]{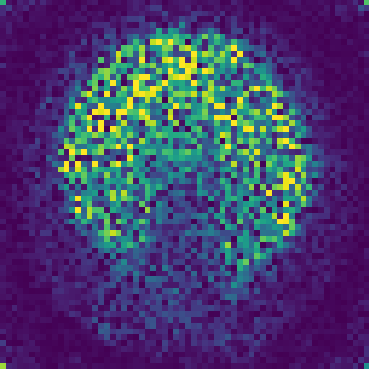}&
		\includegraphics[width=2cm]{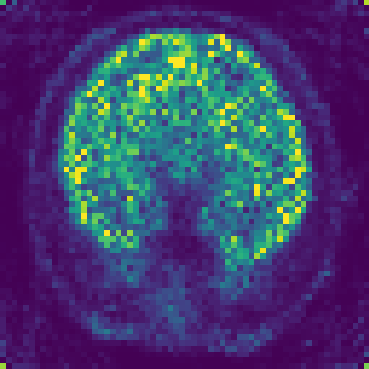}&
		\includegraphics[width=2cm]{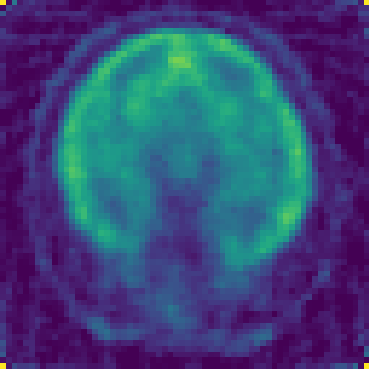}&
		\includegraphics[width=2cm]{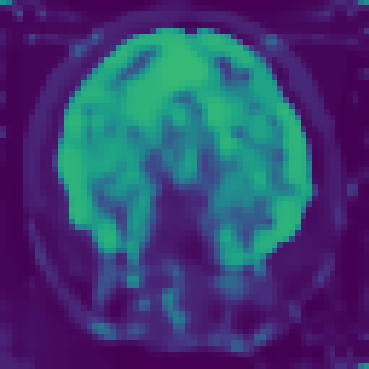}&
		\includegraphics[width=2cm]{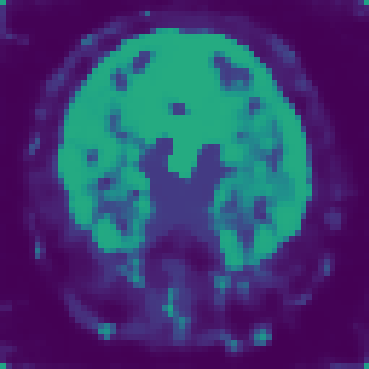}&
		\includegraphics[width=2cm]{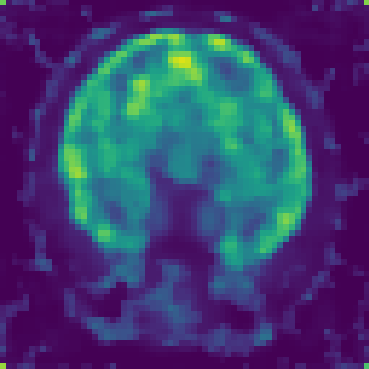}\\
  \put(-15,10){\rotatebox{90}{error map}}& &
        \includegraphics[width=2cm]{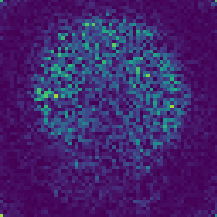}&
	\includegraphics[width=2cm]{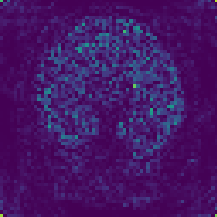}&
	\includegraphics[width=2cm]{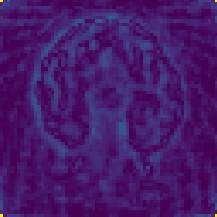}&
	\includegraphics[width=2cm]{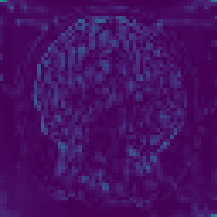}&
	\includegraphics[width=2cm]{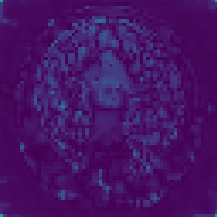}&
        \includegraphics[width=2cm]{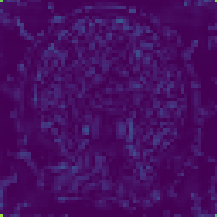}\\
   &Truth &\texttt{EM}& \texttt{EM-NMF}&\texttt{MAP-TV}& \texttt{DIP-B}& \texttt{INR-B}& \texttt{NINRF}
   
    \end{tabular}
    }  
\caption{The reconstructed images and corresponding error maps at Frame 6, 16, and 36 (with random events).}
\label{rec32_show}
\end{figure}

We present the experimental results for the reconstruction task using a clinical brain PET image. The image sequence consists of $T=41$ frames, with time intervals of $15\times15s$, $16\times60s$, $9\times300s$. Selected frames are shown in the first column of \cref{rec32_show}. Experiments are conducted under two conditions: with and without random events. The projection data are  generated following the same procedure as in previous experiments. Specifically, the Radon transform is applied with $n_a = 16$ projection angles uniformly distributed between 0 and 180 degrees. In the random events setting, uniform random events are simulated and account for 10\% of the noise free data in all frames. Poisson noise is then introduced to simulate measurement noise and noise level is set to an SNR of 20 dB.

For the methods \texttt{DIP-B}, \texttt{INR-B}, and \texttt{NINRF}, We set the model rank $K=8$, using the same network architecture described in \cref{exp_rat}. For scenarios without and with random events, the regularization parameters in \texttt{MAP-TV} are set to $\lambda_{\text{TV}_1}=0.002, 0.001$ and $\lambda_{\text{TV}_2}=10, 5$. For \texttt{DIP-B}, the regularization parameters for $\bm{A}$ and $\bm{B}$ are set to $\lambda_1=0.0001, 0.00001$ and $\lambda_2=10, 1$, and for \texttt{INR-B} they are $\lambda_1=0.01, 0.0001$ and $\lambda_2=0.001, 0.0001$. For \texttt{NINRF}, the corresponding values are $\lambda_1=5, 2$ and $\lambda_2=0.0005, 0.0001$. 

\begin{figure}[htbp!]
    \centering
    \subfigure[PSNR]{
    \includegraphics[width=0.4\textwidth]{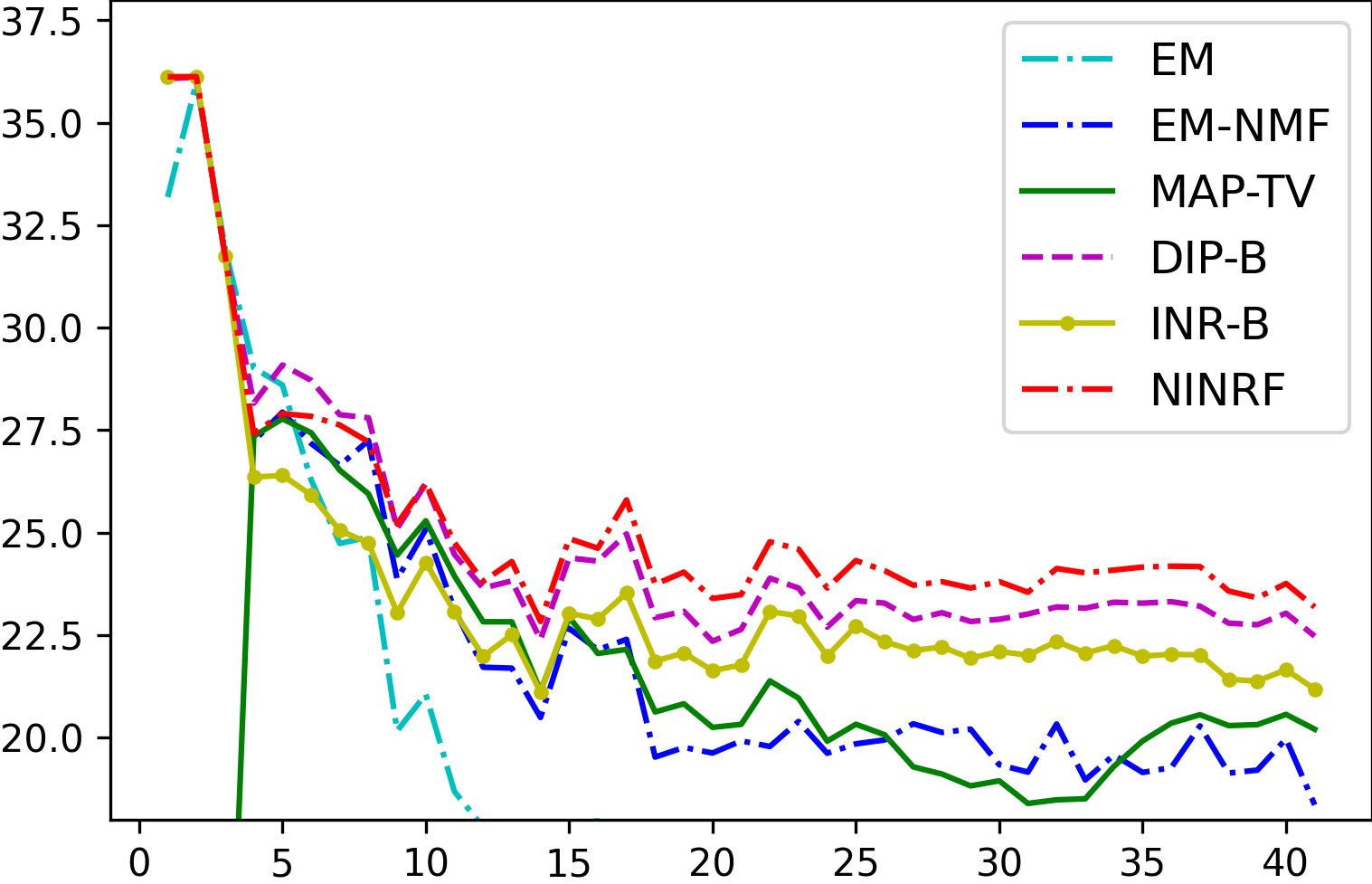}
    }
    \subfigure[SSIM]{
    \includegraphics[width=0.4\textwidth]{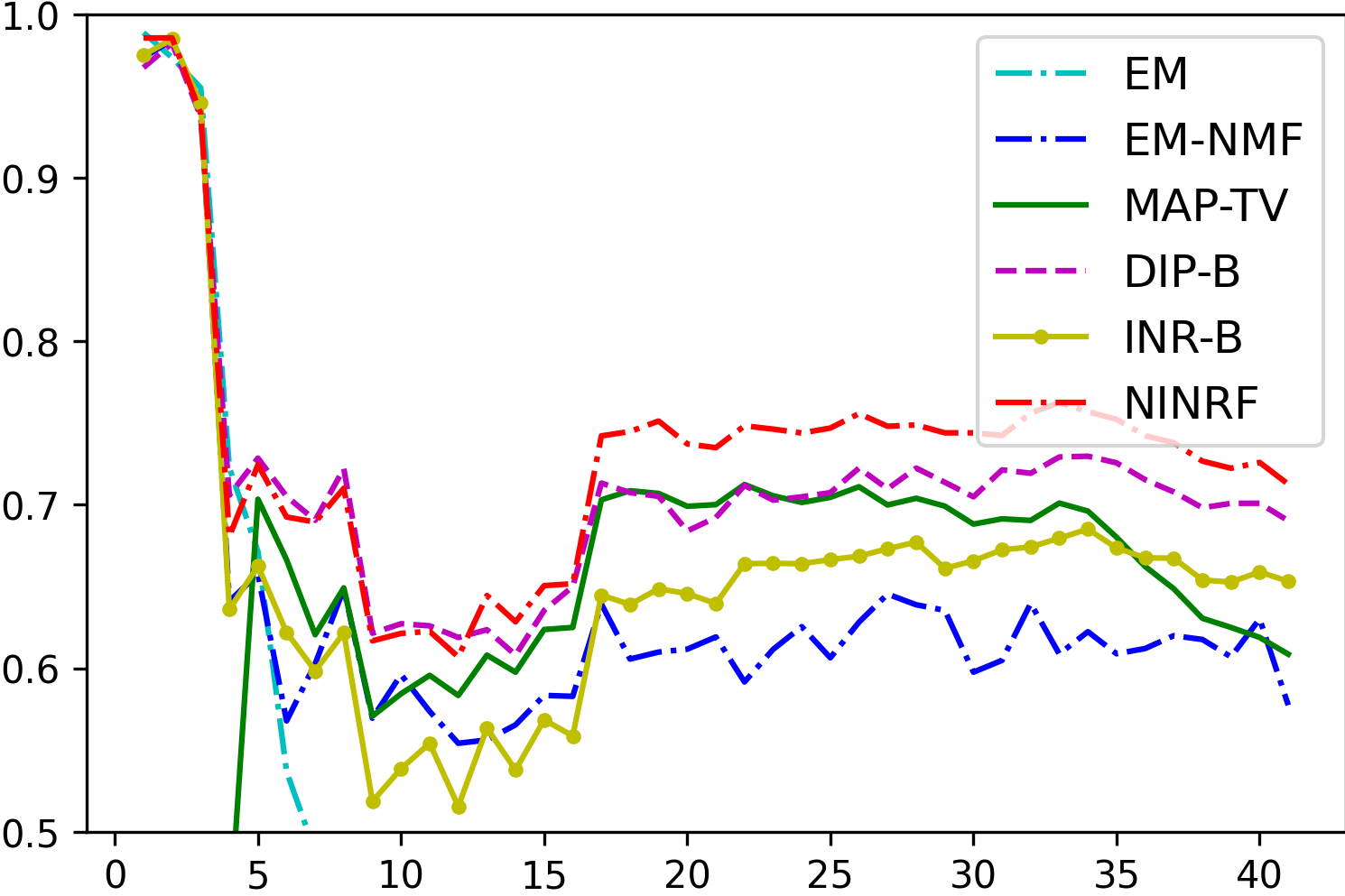}
    }
    \caption{PSNR and SSIM of every frame of the reconstructed clinical brain image (with random events). The horizontal axis represents time, while the vertical axis denotes the corresponding values.}
    \label{rec32_PS}
\end{figure}

Treating the data as a 3D image, we report PSNR and SSIM values for both settings in \cref{rec3psnr}. Several frames of the reconstructed images and corresponding error maps under random events are shown in \cref{rec32_show}. \texttt{NINRF} demonstrates superior performance, both visually and quantitatively. Frame-wise PSNR and SSIM values with random events are presented in \cref{rec32_PS}, where \texttt{NINRF} consistently achieves the highest metrics. In most frames, \texttt{NINRF} achieves the highest metrics. The results without random events are displayed in \cref{rec3res}. 
%Overall, these findings confirm that our method is effective for clinical data and outperforms the comparison methods.

\begin{table}[htbp!]
\caption{PSNR and SSIM of the reconstructed clinical data. The bold numbers mark the best performances.}
\centering
\begin{tabular}{cc|cccccc}
\hline
\multicolumn{2}{c|}{}& EM     & EM-NMF & MAP-TV     & DIP-B  & INR-B  & NINRF \\ \hline
\multicolumn{1}{c|}{\multirow{2}{*}{\begin{tabular}[c]{@{}c@{}}without\\ random events\end{tabular}}} & PSNR & 19.7  & 24.81 & 28.33   & 28.87  & 26.88  & \textbf{29.60} \\
\multicolumn{1}{c|}{} & SSIM & 0.5173 & 0.7632 & 0.8451 & 0.8461 & 0.7904 & \textbf{0.8585} \\ \hline
\multicolumn{1}{c|}{\multirow{2}{*}{\begin{tabular}[c]{@{}c@{}}with\\ random events\end{tabular}}}   & PSNR & 19.46  & 24.18 & 25.69   & 27.37  & 26.08  & \textbf{28.02} \\
\multicolumn{1}{c|}{} & SSIM & 0.4352 & 0.6807 & 0.7361 & 0.7613 & 0.7154 & \textbf{0.7985} \\ \hline
\end{tabular}
\label{rec3psnr}
\end{table}

\section{Conclusion}\label{conclude}
In this paper, we proposed a novel approach that combines INRs and NMF for modeling dynamic PET images. Fully connected neural networks are employed to represent the non-negative matrices in the NMF model. By treating the dynamic PET image as tensor data, we demonstrated that the proposed model can effectively represent it using INR, which captures the low-rank structure inherent in the data. We applied this framework to reconstruct dynamic PET images affected by Poisson noise. 

Numerical experiments on both simulated data and clinical data demonstrated that our proposed method outperformed other comparison methods. Our model achieved higher PSNR and SSIM values, preserving fine details in reconstructed images across different datasets. Additionally, in the calculation of the net influx rate constant for the brain data, our method also outperformed others across all regions. The experimental results confirm that our proposed method is capable of achieving high-quality image reconstruction across different types of data, making it a robust and effective solution for dynamic imaging applications.

\appendix
\section{Proof of \cref{thm0}}\label{proof_thm0}
\begin{proof}
    Since $\text{F-rank}[G]_3=K$, there exists a tensor $\mathcal{T}$ such that $\text{rank}(\mathbf{T}^{(3)})=K$, where $\mathbf{T}^{(3)}$ is the unfolding matrix of $\mathcal{T}$ along mode $3$. The coordinate vectors of sampling $\mathcal{T}$ from $G(\cdot)$ is $\mathbf{v}_1\in\mathbb{R}^{n_1}$, $\mathbf{v}_2\in\mathbb{R}^{n_2}$ and $\mathbf{v}_3\in\mathbb{R}^{n_3}$. For any $v_1\in D_h$ and $v_2\in D_w$, we introduce new coordinate vectors $\Tilde{\mathbf{v}}_1=[\mathbf{v}_1, v_1]$ and $\Tilde{\mathbf{v}}_2=[\mathbf{v}_2, v_2]$. A new tensor $\Tilde{\mathcal{T}}\in\mathbb{R}^{(n_1+1)\times(n_2+1)\times n_3}$ is defined as $\Tilde{\mathcal{T}}_{(i,j,k)}=G(\Tilde{\mathbf{v}_1}_{(i)}, \Tilde{\mathbf{v}_2}_{(j)}, {\mathbf{v}_3}_{(k)})$ and $\text{rank}(\Tilde{\mathbf{T}}^{(3)})=K$. There exist $\mathcal{T}_{(i_1,j_i,:)}, \mathcal{T}_{(i_2,j_2,:)}, \dots, \mathcal{T}_{(i_K,j_K,:)}$ which are $K$ basis vectors of the column space of $\Tilde{\mathbf{T}}^{(3)}$. Thus, $G(v_1,v_2,\mathbf{v}_3)=\Tilde{\mathcal{T}}_{(n_1+1,n_2+1,:)}$ can be linearly represented as
    \begin{equation*}
        G(v_1,v_2,\mathbf{v}_3) = \sum_{l=1}^{K}{\mathbf{c}_l\mathcal{T}_{(i_l,j_l,:)}} = \sum_{l=1}^{K}{\mathbf{c}_lG({\mathbf{v}_1}_{(i_l)}, {\mathbf{v}_2}_{(j_l)}, \mathbf{v}_3)}
    \end{equation*}
    where $\mathbf{c}$ is a unique coefficient vector. Then we introduce a new vector $\Tilde{\mathbf{v}_3}=[\mathbf{v}_3, v_3]$, which leads to a new tensor $\Hat{\mathcal{T}}\in\mathbb{R}^{(n_1+1)\times(n_2+1)\times(n_3+1)}$ defined as $\Hat{\mathcal{T}}_{(i,j,k)}=G(\Tilde{\mathbf{v}_1}_{(i)}, \Tilde{\mathbf{v}_2}_{(j)}, \Tilde{\mathbf{v}_3}_{(k)})$ and $\text{rank}(\Hat{\mathbf{T}}^{(3)})=K$. Thus $\hat{\mathcal{T}}_{(i_1,j_i,:)}, \hat{\mathcal{T}}_{(i_2,j_2,:)}, \dots, \hat{\mathcal{T}}_{(i_K,j_K,:)}$ are $K$ basis vectors of the column space of $\Hat{\mathbf{T}}^{(3)}$. Similarly, $\Hat{\mathcal{T}}_{(n_1+1,n_2+1,:)}$ can be linearly represented by these basis vectors. Since the first $n_1$ elements are just the vector of $\Tilde{\mathcal{T}}_{(n_1+1,n_2+1,:)}$ and due to the uniqueness of the coefficient vector, the linear representation is supposed to be
    \begin{equation*}
        G(v_1,v_2,\Tilde{\mathbf{v}_3}) = \hat{\mathcal{T}}_{(n_1+1,n_2+1,:)} = \sum_{l=1}^{K}{\mathbf{c}_l\Hat{\mathcal{T}}_{(i_l,j_l,:)}} = \sum_{l=1}^{K}{\mathbf{c}_lG({\mathbf{v}_1}_{(i_l)}, {\mathbf{v}_2}_{(j_l)}, \Tilde{\mathbf{v}_3})}.
    \end{equation*}
    Thus the last element is $G(v_1,v_2,v_3)=\sum_{l=1}^{K}{\mathbf{c}_lG({\mathbf{v}_1}_{(i_l)}, {\mathbf{v}_2}_{(j_l)}, v_3)}$. In this equation, $\mathbf{c}_l$ is related to the choice of $v_1$ and $v_2$. We set $H_f(v_1,v_2)=\mathbf{c}$ and  $H_g(v_3)=[G({\mathbf{v}_1}_{(i_1)},{\mathbf{v}_2}_{(j_1)}, v_3),$ $G({\mathbf{v}_1}_{(i_2)}, {\mathbf{v}_2}_{(j_2)}, v_3), \dots, G({\mathbf{v}_1}_{(i_K)}, {\mathbf{v}_2}_{(j_K)}, v_3)]^{\top}$, which results in the conclusion.
\end{proof}

\section{Proof of \cref{thm1}}\label{proof_thm1}
\begin{proof}
    We denote all the elements of $\mathcal{S}_1$ as $\{\mathcal{T}_{(i_l,j_l,:)}\}_{l=1}^{K}$. Since $\text{Cone}(\mathcal{S}_0)\subset\text{Cone}(\mathcal{S}_1)$,  for any $v_1\in D_h$ and $v_2\in D_w$, the tensor $G(v_1,v_2,\mathbf{v}_3)$ can be represented by the non-negative linear combination of $\mathcal{S}_1$, which is 
    \begin{equation*}
        G(v_1,v_2,\mathbf{v}_3) = \sum_{l=1}^{K}\mathbf{c}_l\mathcal{T}_{(i_l,j_l,:)} = \sum_{l=1}^{K}\mathbf{c}_l G({\mathbf{v}_1}_{(i_l)},{\mathbf{v}_2}_{(i_l)},\mathbf{v}_3).
    \end{equation*}
    Since $\text{rank}(\mathbf{T}^{(3)})=\text{rank}_{+}(\mathbf{T}^{(3)})=K$, all the vectors in $\mathcal{S}_1$ are linearly independent and $G(v_1,v_2,\mathbf{v}_3)$ is in this linear space. $\text{F-rank}[G]_3=\text{F-rank}_{+}[G]_3=K$ indicates that $\mathbf{c}$ is a unique coefficient vector.
    
    For any $v_3\in D_T$, we introduce a new vector $\Tilde{\mathbf{v}_3}=[\mathbf{v}_3, v_3]$. We also denote $\Tilde{\mathbf{v}}_1=[\mathbf{v}_1, v_1]$ and $\Tilde{\mathbf{v}}_2=[\mathbf{v}_2, v_2]$ and define $\Hat{\mathcal{T}}\in\mathbb{R}^{(n_1+1)\times(n_2+1)\times(n_3+1)}$ whose element is $\Hat{\mathcal{T}}_{(i,j,k)}=G(\Tilde{\mathbf{v}_1}_{(i)}, \Tilde{\mathbf{v}_2}_{(j)}, \Tilde{\mathbf{v}_3}_{(k)})$ and $\text{rank}(\Hat{\mathbf{T}}^{(3)})=K$. Thus $\hat{\mathcal{T}}_{(i_1,j_i,:)}, \hat{\mathcal{T}}_{(i_2,j_2,:)}, \dots, \hat{\mathcal{T}}_{(i_K,j_K,:)}$ are $K$ basis vectors of the column space of $\Hat{\mathbf{T}}^{(3)}$. Similarly, $\Hat{\mathcal{T}}_{(n_1+1,n_2+1,:)}$ can be linearly represented by these basis vectors. Then the following proof is exactly same as in \cref{proof_thm0}.
\end{proof}

\section{Numerical Results on Clinical Data without Random Events}\label{rec3res}
\begin{figure}[htbp!]
    \centering
    \resizebox{\textwidth}{!}{
    \begin{tabular}{c@{\hspace{2pt}}c@{\hspace{1pt}}c@{\hspace{1pt}}c@{\hspace{1pt}}c@{\hspace{1pt}}c@{\hspace{1pt}}c@{\hspace{1pt}}c}
		\put(-15,10){\rotatebox{90}{Frame 6}}&
        \includegraphics[width=2cm]{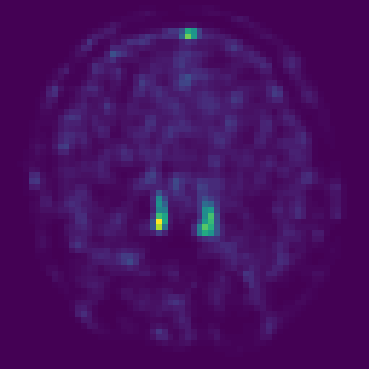}&
		\includegraphics[width=2cm]{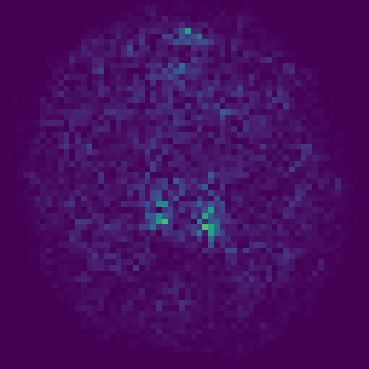}&
		\includegraphics[width=2cm]{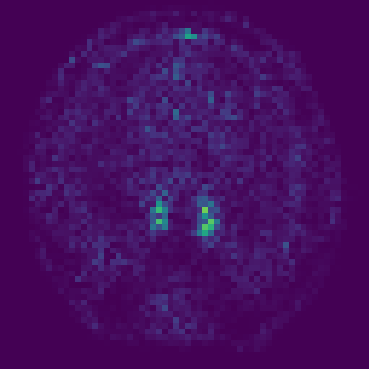}&
		\includegraphics[width=2cm]{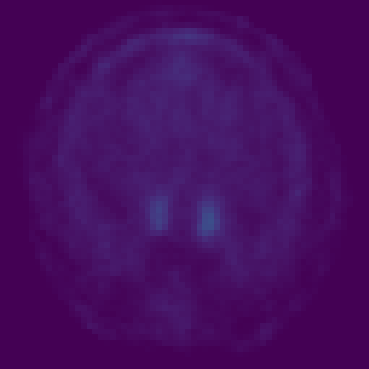}&
		\includegraphics[width=2cm]{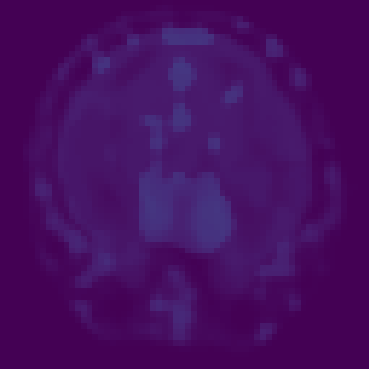}&
		\includegraphics[width=2cm]{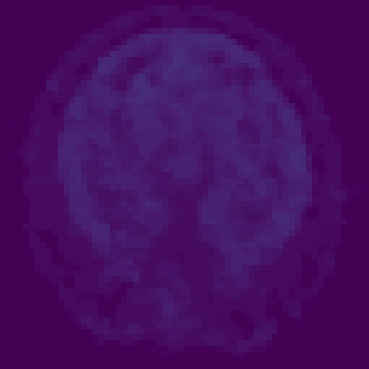}&
		\includegraphics[width=2cm]{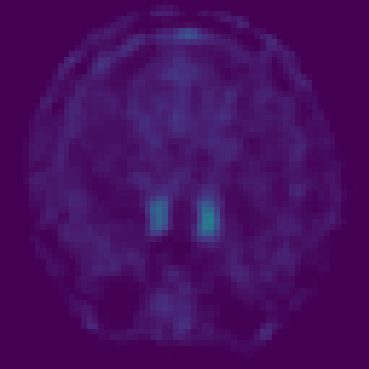}\\
        \put(-15,10){\rotatebox{90}{error map}}& &
        \includegraphics[width=2cm]{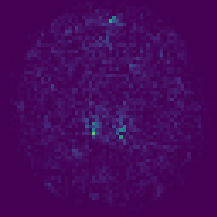}&
	\includegraphics[width=2cm]{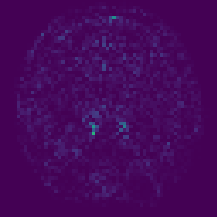}&
	\includegraphics[width=2cm]{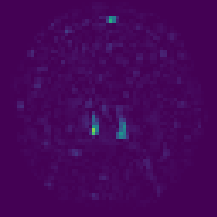}&
	\includegraphics[width=2cm]{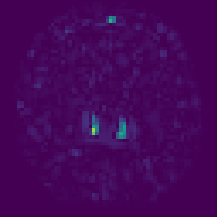}&
	\includegraphics[width=2cm]{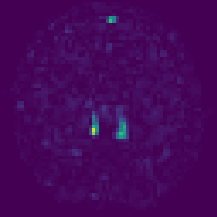}&
        \includegraphics[width=2cm]{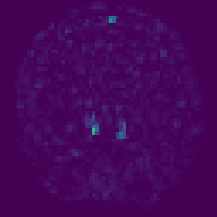}\\
        \put(-15,10){\rotatebox{90}{Frame 16}}&
	\includegraphics[width=2cm]{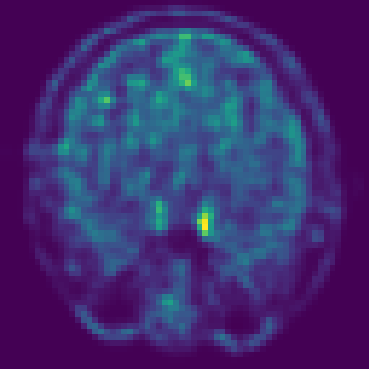}&
		\includegraphics[width=2cm]{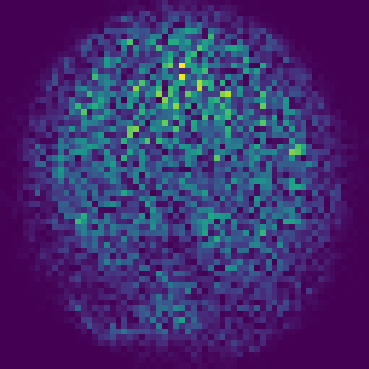}&
		\includegraphics[width=2cm]{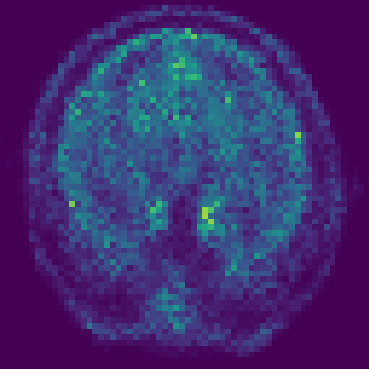}&
		\includegraphics[width=2cm]{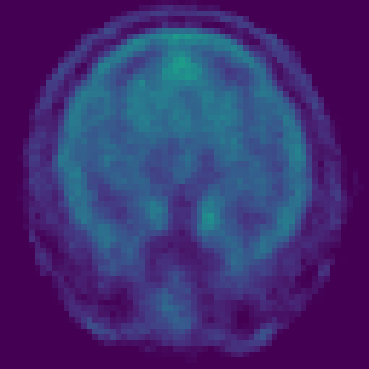}&
		\includegraphics[width=2cm]{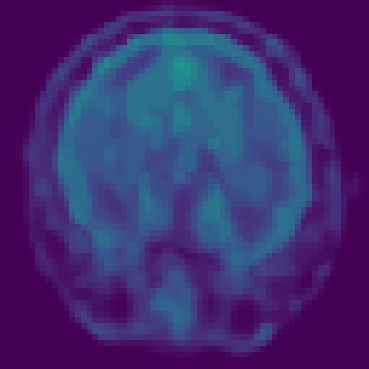}&
		\includegraphics[width=2cm]{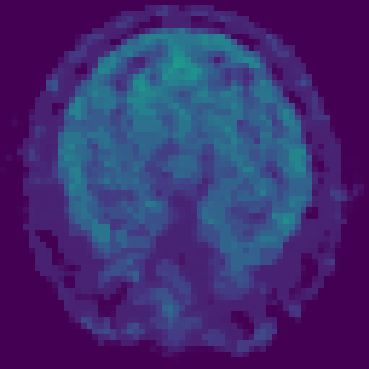}&
		\includegraphics[width=2cm]{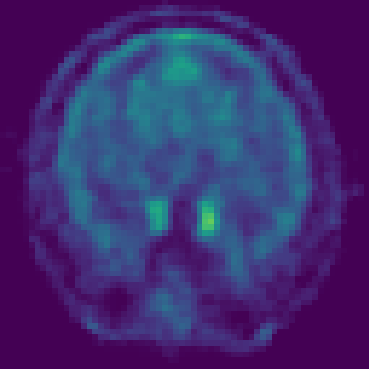}\\
  \put(-15,10){\rotatebox{90}{error map}}& &
        \includegraphics[width=2cm]{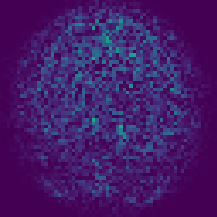}&
	\includegraphics[width=2cm]{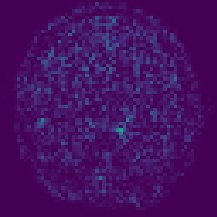}&
	\includegraphics[width=2cm]{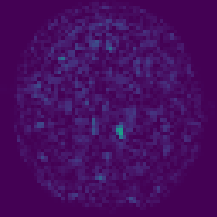}&
	\includegraphics[width=2cm]{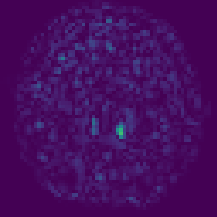}&
	\includegraphics[width=2cm]{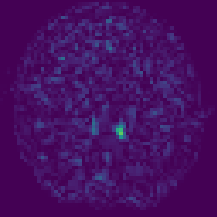}&
        \includegraphics[width=2cm]{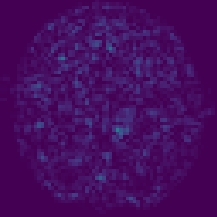}\\
 %  \put(-15,10){\rotatebox{90}{Frame 26}}&
 %      \includegraphics[width=2cm]{img/rec3/num16/utruth_t=26.png}&
	% 	\includegraphics[width=2cm]{img/rec3/num16/uEMU_t=26.png}&
	% 	\includegraphics[width=2cm]{img/rec3/num16/uEMNMF_t=26.png}&
	% 	\includegraphics[width=2cm]{img/rec3/num16/uADMMU_t=26.png}&
	% 	\includegraphics[width=2cm]{img/rec3/num16/uDIPEM_t=26.png}&
	% 	\includegraphics[width=2cm]{img/rec3/num16/uINREM_t=26.png}&
	% 	\includegraphics[width=2cm]{img/rec3/num16/uINRINR_t=26.png}\\
 %  \put(-15,10){\rotatebox{90}{error map}}& &
 %        \includegraphics[width=2cm]{img/rec3/showres2/reuEMU_t=26.png}&
	% \includegraphics[width=2cm]{img/rec3/showres2/reuEMNMF_t=26.png}&
	% \includegraphics[width=2cm]{img/rec3/showres2/reuADMMU_t=26.png}&
	% \includegraphics[width=2cm]{img/rec3/showres2/reuDIPEM_t=26.png}&
	% \includegraphics[width=2cm]{img/rec3/showres2/reuINREM_t=26.png}&
 %        \includegraphics[width=2cm]{img/rec3/showres2/reuINRINR_t=26.png}\\
    \put(-15,10){\rotatebox{90}{Frame 36}}&
      \includegraphics[width=2cm]{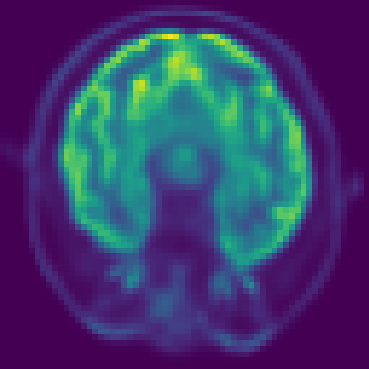}&
		\includegraphics[width=2cm]{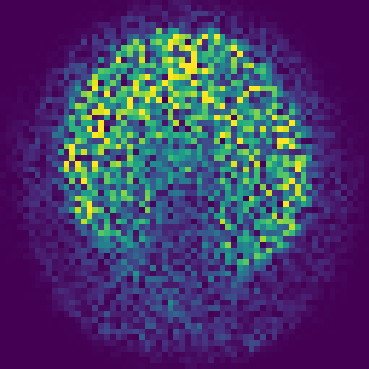}&
		\includegraphics[width=2cm]{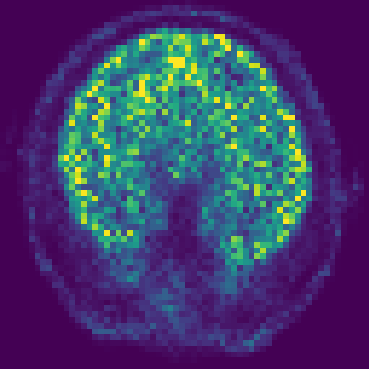}&
		\includegraphics[width=2cm]{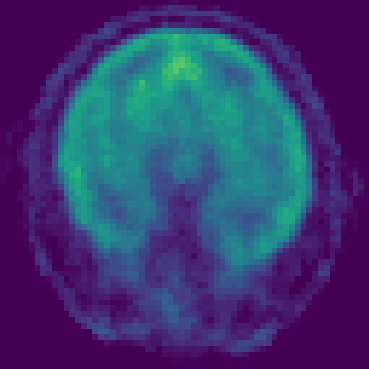}&
		\includegraphics[width=2cm]{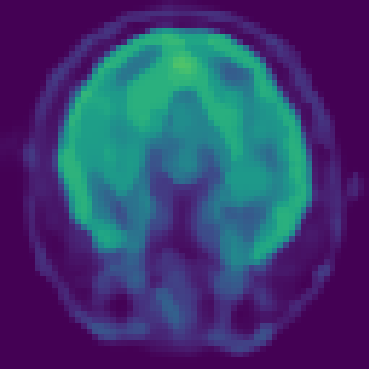}&
		\includegraphics[width=2cm]{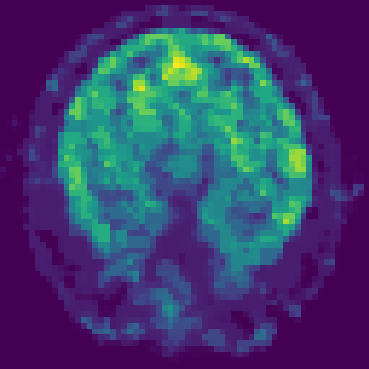}&
		\includegraphics[width=2cm]{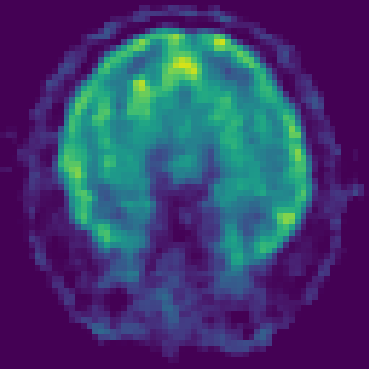}\\
  \put(-15,10){\rotatebox{90}{error map}}& &
        \includegraphics[width=2cm]{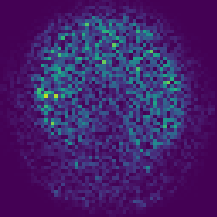}&
	\includegraphics[width=2cm]{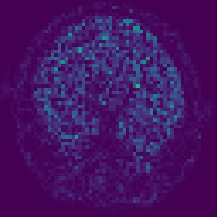}&
	\includegraphics[width=2cm]{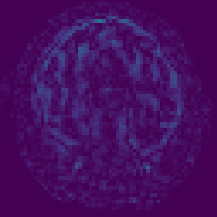}&
	\includegraphics[width=2cm]{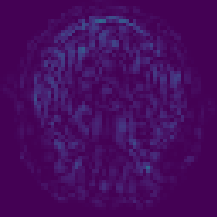}&
	\includegraphics[width=2cm]{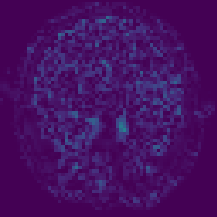}&
        \includegraphics[width=2cm]{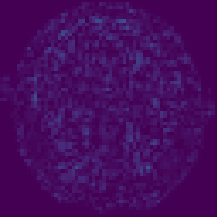}\\
   &Truth &\texttt{EM}& \texttt{EM-NMF}&\texttt{MAP-TV}& \texttt{DIP-B}& \texttt{INR-B}& \texttt{NINRF}
   
    \end{tabular}
    }  
\caption{The reconstructed images and corresponding error maps at Frame 6, 16, and 36 (without random events). }
\label{rec3_show}
\end{figure}

\begin{figure}[htbp!]
    \centering
    \subfigure[PSNR]{
    \includegraphics[width=0.4\textwidth]{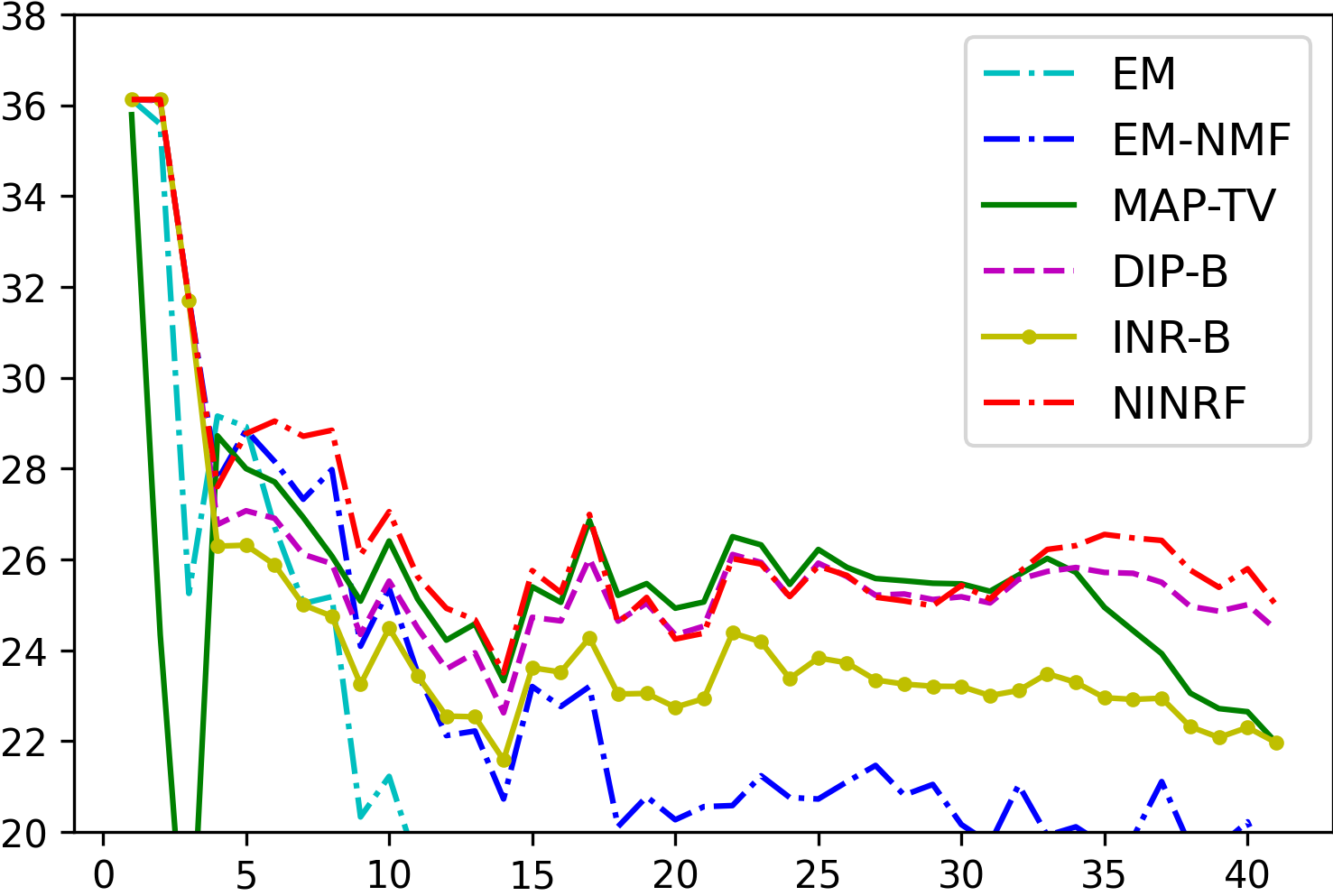}
    }
    \subfigure[SSIM]{
    \includegraphics[width=0.4\textwidth]{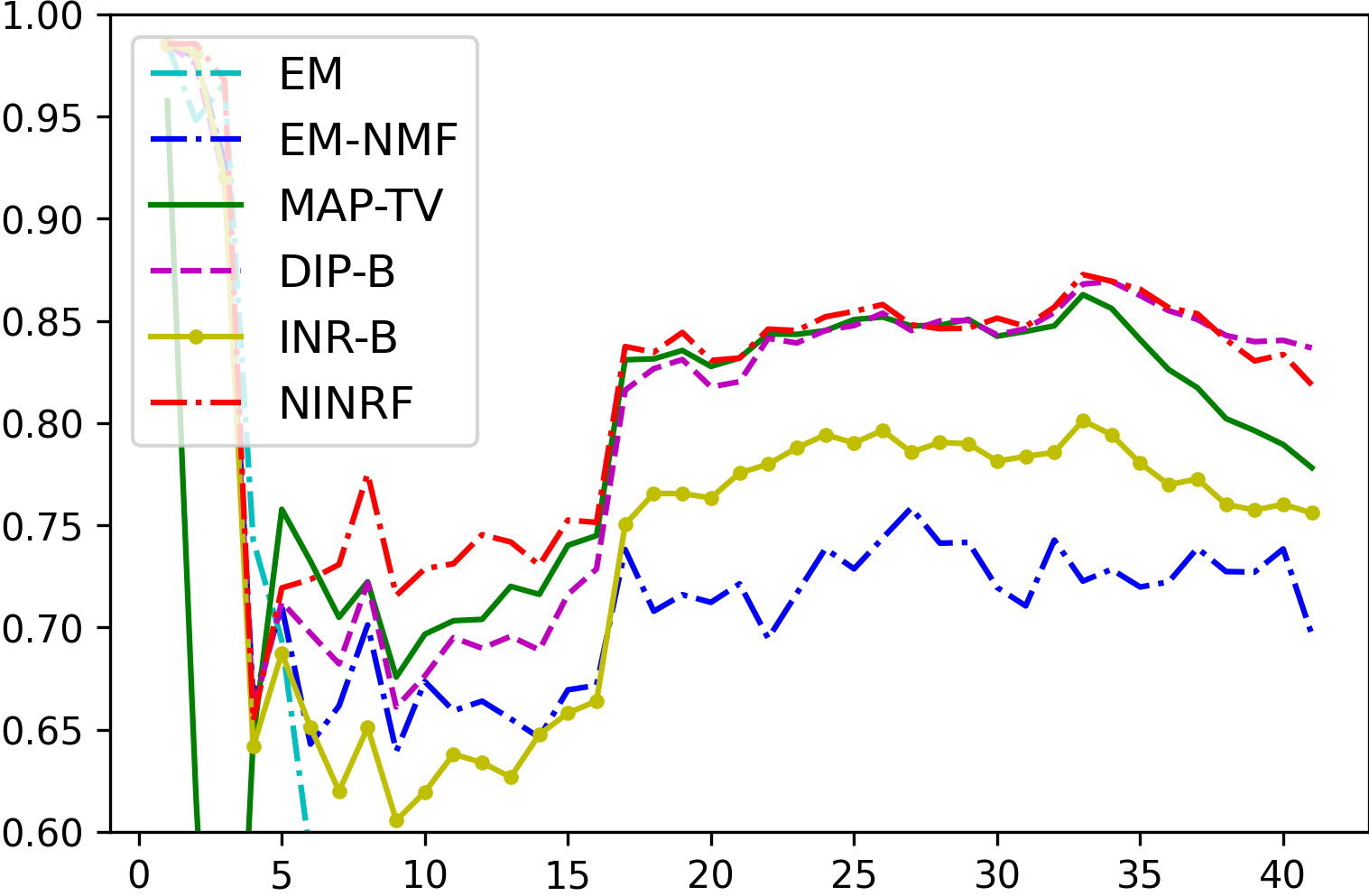}
    }
    \caption{PSNR and SSIM of every frame of the reconstructed clinical brain image. The horizontal axis represents time, while the vertical axis denotes the corresponding values.}
    \label{rec3_PS}
\end{figure}

Reconstructed images and corresponding error maps without random events are shown in \cref{rec3_show}. Frame-wise PSNR and SSIM values without random events are presented in \cref{rec3_PS}.

% \section*{Acknowledgments}
% We would like to acknowledge the assistance of volunteers in putting
% together this example manuscript and supplement.

\bibliographystyle{plainnat}
\bibliography{references.bib}

\end{document}